%% file: main.tex
\documentclass[pdflatex,sn-mathphys-num]{sn-jnl}
\usepackage{subcaption}%
\usepackage{graphicx}%
\usepackage{multirow}%
\usepackage{amsmath,amssymb,amsfonts}%
\usepackage{amsthm}%
\usepackage{mathrsfs}%
\usepackage{tikz}%
\usepackage[title]{appendix}%
\usepackage{textcomp}%
\usepackage{manyfoot}%
\usepackage{booktabs}%
\usepackage{algorithm}%
\usepackage{algorithmicx}%
\usepackage{algpseudocode}%
\usepackage{listings}%
\usepackage[table]{xcolor}
\usepackage{makecell}
\usepackage{anyfontsize}
\usepackage{tabularx}%
\usepackage{rotating}
\usepackage{pdflscape}
\usepackage[shortcuts,acronym,nohypertypes={acronym}]{glossaries}
\usepackage{pgfplots}
\pgfplotsset{compat=1.18}
\usepackage[table,xcdraw]{xcolor}
\usepackage{booktabs}
\usepackage{color}
\usepackage{rotating}
\usepackage{placeins}
\usepackage{tabularray}
\usepackage{multicol}
\usepackage{pdfpages}
\definecolor{DustyGray}{rgb}{0.607,0.607,0.607}

\theoremstyle{thmstyleone}%
\theoremstyle{thmstyletwo}%
\theoremstyle{thmstylethree}%
\newacronym{eu}{EU}{European Union}
\newacronym{ml}{ML}{machine learning}
\newacronym{ai}{AI}{artificial intelligence}
\newacronym{metric}{METRIC}{}
\newacronym{aimq}{AIMQ}{AIM quality}

\usepackage{pgf}

\begin{document}


\title[Article Title]{\vspace{-3cm}Metric Hub: A metric library and practical selection workflow for use-case-driven data quality assessment in medical AI}


\author*[1]{%
  \fnm{Katinka} \sur{Becker}}
\email{katinka.becker@ptb.de}\equalcont{These authors contributed equally to this work.}

\author[2,3]{%
  \fnm{Maximilian P.} \sur{Oppelt}}
\email{maximilian.oppelt@iis.fraunhofer.de}\equalcont{These authors contributed equally to this work.}

\author[2]{%
  \fnm{Tobias S.} \sur{Zech}}
\email{tobias.sebastian.zech@iis.fraunhofer.de}

\author[1]{%
  \fnm{Martin} \sur{Seyferth}}
\email{martin.seyferth@ptb.de}

\author[4]{%
  \fnm{Sandie} \sur{Cabon}}
\email{sandie.cabon@univ-rennes.fr}

\author[5,6]{%
  \fnm{Vanja} \sur{Miskovic}}
\email{vanja.miskovic@polimi.it}

\author[7]{%
  \fnm{Ivan} \sur{Cimrak}}
\email{ivan.cimrak@fri.uniza.sk}

\author[8]{%
  \fnm{Michal} \sur{Kozubek}}
\email{kozubek@fi.muni.cz}

\author[9]{%
  \fnm{Giuseppe} \sur{D'Avenio}}
\email{giuseppe.davenio@iss.it}

\author[9]{%
  \fnm{Ilaria} \sur{Campioni}}
\email{ilaria.campioni@iss.it}

\author[10]{%
  \fnm{Jana} \sur{Fehr}}
\email{jana.fehr@bih-charite.de}

\author[11]{%
  \fnm{Kanjar} \sur{De}}
\email{kanjar.de@ri.se}

\author[12]{%
  \fnm{Ismail} \sur{Mahmoudi}}
\email{ismail.mahmoudi@cetic.be}

\author[13]{%
  \fnm{Emilio} \sur{Dolgener Cantu}}
\email{emilio.dolgener.cantu@hhi.fraunhofer.de}

\author[2]{%
  \fnm{Laurenz} \sur{Ottmann}}
\email{laurenz.ottmann@iis.fraunhofer.de}

\author[1]{%
  \fnm{Andreas} \sur{Klaß}}
\email{andreas.klass@ptb.de}

\author[14]{%
  \fnm{Galaad} \sur{Altares}}
\email{altares@multitel.be}

\author[13]{%
  \fnm{Jackie} \sur{Ma}}
\email{jackie.ma@hhi.fraunhofer.de}

\author[11]{%
  \fnm{Alireza} \sur{Salehi M.}}
\email{alireza.salehi@ri.se}

\author[2]{%
  \fnm{Nadine R.} \sur{Lang-Richter}}
\email{nadine.lang-richter@iis.fraunhofer.de}

\author[1, 15, 16]{%
  \fnm{Tobias} \sur{Schaeffter}}
\email{tobias.schaeffter@ptb.de}

\author[1]{%
  \fnm{Daniel} \sur{Schwabe}}
\email{daniel.schwabe@ptb.de}

\affil[1]{Division Medical Physics and Metrological Information Technology, Physikalisch-Technische Bundesanstalt, Berlin, Germany}
\affil[2]{Department Digital Health and Analytics, Fraunhofer IIS, Fraunhofer Institute for Integrated Circuits IIS, Erlangen, Germany}
\affil[3]{Department Artificial Intelligence in Biomedical Engineering, Friedrich-Alexander-University Erlangen Nuremberg, 91052 Erlangen}
\affil[4]{Univ Rennes, Inserm, LTSI - UMR 1099, F-35000 Rennes, France}
\affil[5]{Nearlab, Department of Electronics, Information, and Bioengineering, Politecnico
di Milano, Milano, Italy}
\affil[6]{AI-ON-Lab, Medical Oncology Department, Fondazione IRCCS Istituto Nazionale dei Tumori di Milano, Milan, Italy}
\affil[7]{Department of Software Technologies, Faculty of Management Science and Informatics, University of Žilina,
010 26 Žilina, Slovakia}
\affil[8]{Faculty of Informatics, Masaryk University, Brno, Czech Republic}
\affil[9]{National Centre Artificial Intelligence and Innovative Technologies for Health, Istituto Superiore di
Sanità, Rome, Italy}
\affil[10]{QUEST Center for Responsible Research, Berlin Institute of Health (BIH), Charité Universitätsmedizin Berlin, Berlin, Germany}
\affil[11]{RISE Research Institutes of Sweden, Sweden}
\affil[12]{CETIC Centre d’excellence en technologies de l’information et de la communication, Belgium}
\affil[13]{Fraunhofer Institute for Telecommunications Heinrich-Hertz-Institute HHI, Berlin, Germany}
\affil[14]{Multitel Research Center, Mons, Belgium}
\affil[15]{Department of Medical Engineering, Technical University Berlin, Berlin, Germany}
\affil[16]{Einstein Centre for Digital Future, Berlin, Germany}

\abstract{Machine learning (ML) in medicine has transitioned from research to concrete applications aimed at supporting several medical purposes like therapy selection, monitoring and treatment. Acceptance and effective adoption by clinicians and patients, as well as regulatory approval, require evidence of trustworthiness. A major factor for the development of trustworthy AI is the quantification of data quality for AI model training and testing. 
We have recently proposed the METRIC-framework for systematically evaluating the suitability (fit-for-purpose) of data for medical ML for a given task. 
Here, we operationalize this theoretical framework by introducing a collection of data quality metrics -- the metric library -- for practically measuring data quality dimensions. 
For each metric, we provide a metric card with the most important information, including definition, applicability, examples, pitfalls and recommendations, to support the understanding and implementation of these metrics.
Furthermore, we discuss strategies and provide decision trees for choosing an appropriate set of data quality metrics from the metric library given specific use cases.
We demonstrate the impact of our approach exemplarily on the PTB-XL ECG-dataset.
This is a first step to enable fit-for-purpose evaluation of training and test data in practice as the base for establishing trustworthy AI in medicine.}


\maketitle
\section{Introduction}
Machine learning-based systems in medicine have moved from research prototypes to clinical applications that support diagnosis, monitoring, and treatment~\cite{chenAlgorithmicFairnessArtificial2023,zhangEthicsGovernanceTrustworthy2023}. Their acceptance by clinicians and patients, requires evidence that these systems are trustworthy~\cite{bikkasaniNavigatingArtificialGeneral2025,hannaEthicalBiasConsiderations2025}. Recent initiatives by international organizations and regulators emphasize that trustworthiness depends on the quality and governance of the underlying data~\cite{WHO-DHI2023,Griesinger2025,EC-2019,EC-2020}. For high-risk systems, the EU AI Act explicitly requires that datasets are \emph{relevant}, sufficiently \emph{representative}, and, to the best extent possible, \emph{free of errors and complete} for their intended purpose, and that data quality be documented~\cite{HarmonisedRulesArtificial2024}. This places quantitative data quality assessment at the core of development, evaluation, and regulation of medical \gls{ai}.

By contrast, quantitative methods for assessing data quality remain underdeveloped compared to the rich ecosystem of model-centric performance metrics of AI models for classification, regression, detection, and segmentation~\cite{SOKOLOVA2009427,taha_metrics_2015,miller_review_2024,reinkeUnderstandingMetricrelatedPitfalls2024}. Data issues are a major source of failures in \gls{ml}~\cite{sculleyHiddenTechnicalDebt2015,rohSurveyDataCollection2021}, yet formal, systematic evaluation of dataset quality is rarely reported. Existing approaches often focus on isolated aspects (e.g., completeness, timeliness, inter-rater agreement) or on specific domains, and do not provide a harmonized, use case–specific methodology for evaluating the data used to train and test medical \gls{ml} models~\cite{Maier-Hein2024}. In practice, it remains common to infer data quality indirectly from model behaviour, treating high model performance as evidence of ``good'' data and poor performance as evidence of ``bad'' data.

This model-oriented view of data quality is problematic. Model performance jointly reflects data properties, model architecture, hyperparameters, optimization procedures, and regularization techniques~\cite{el-sayedDataCentricHitL2025}. Attributing performance differences solely to the data can obscure fundamental issues such as label noise, coverage gaps, or imbalances~\cite{gongSurveyDatasetQuality2023}. If test data do not reflect deployment conditions, even high reported performance does not guarantee clinical utility, and conversely poor performance may stem from suboptimal model design rather than data defects~\cite{gongSurveyDatasetQuality2023,el-sayedDataCentricHitL2025}. Improvements in data realism, such as more heterogeneous populations or inclusion of rare cases, can reduce aggregate performance while improving robustness and clinical usefulness~\cite{zhouEnhancingCornYield2024}. Artifacts in datasets can further lead models to exploit spurious correlations rather than clinically meaningful signals~\cite{lapuschkinUnmaskingCleverHans2019}. These observations motivate a distinction between \emph{intrinsic} data quality and \emph{extrinsic} fitness for a specific task and deployment context~\cite{gongSurveyDatasetQuality2023}, and show that explicit, model-independent data quality assessment is required.

To provide a conceptual basis for such assessment in medicine, we recently introduced the \gls{metric}-framework for data quality in trustworthy medical \gls{ai}~\cite{schwabeMETRICframeworkAssessingData2024}. It compromises approximately 30~years of data quality research and adopts it to the context of medical \gls{ml} data, defining 26 data quality dimensions grouped into five clusters: measurement process, timeliness, representativeness, informativeness, and consistency. The dimensions cover both intrinsic and extrinsic aspects of data quality and map to the terminology of the EU AI Act~\cite{HarmonisedRulesArtificial2024}, positioning the \gls{metric}-framework as a bridge between technical evaluation and regulatory requirements. However, the \gls{metric}-framework is intentionally conceptual. It only specifies which characteristics to investigate but not how to quantitatively assess them, with quantitative metrics, in a systematic and harmonized way for reproducible and defensible data quality evaluations~\cite{fengClinicalArtificialIntelligence2022,ngPerceptionsDataSet2023,dehondGuidelinesQualityCriteria2022,vallevikCanTrustMy2024,elmoreDataQualityData2021,deandradeDiscoveryDataQuality2026}. This often leads to ad hoc judgments about dataset suitability, limited comparability across datasets and studies, and difficulties in prioritizing remediation and monitoring data quality over time.

\begin{table}[htb!]
\centering
\begin{tabular}{p{0.35\linewidth}p{0.35\linewidth}}
\toprule Measurement Process & Accuracy \\
& Noisy labels \\
& Completeness \\
\midrule Timeliness & Currency \\
\midrule Representativeness & Target class balance\\
& Granularity\\
& Dataset size\\
& Variety\\
\midrule Informativeness & Feature importance \\
& Informative missingness\\
& Uniqueness\\
\midrule Consistency & Distribution drift \\
& Homogeneity\\
& Syntactic consistency\\
\bottomrule
\end{tabular}
\caption{Data quality dimensions of the METRIC-framework per cluster that can be assessed utilizing mostly quantitative methods, called quantitative dimensions.}
\label{tab:quantitativedim}
\end{table}

\begin{figure}
\centering
\includegraphics[scale=0.42]{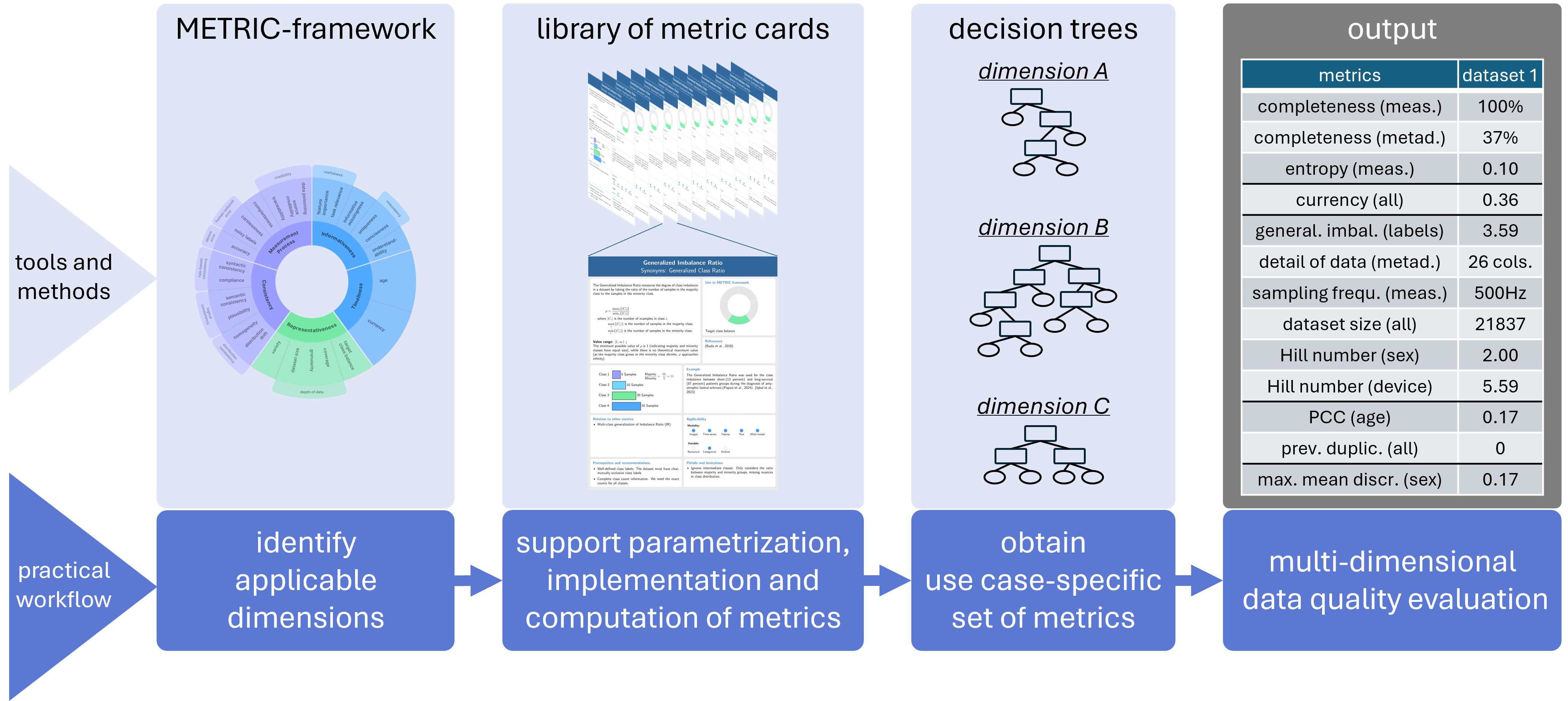}
\caption{Recommended workflow for multi-dimensional data quality evaluation: Identify the data quality dimensions most relevant for risk and intended use; implement and parametrize metrics using the metric cards; follow dimension-specific decision trees to select applicable metrics.}
\label{fig:graphicalabstract}
\end{figure}

In this work, we operationalize the dimensions of the \gls{metric}-framework that can be assessed using mostly quantitative methods (see Table~\ref{tab:quantitativedim}). Using an iterative, focus-group-based process with domain experts from multiple institutions, we collected, curated, and categorized data quality metrics relevant to medical \gls{ml}. The resulting metric library comprises 60 metrics mapped to the 14 quantitative data quality dimensions of the \gls{metric}-framework and to cross-cutting distributional and correlation properties. For each metric, we provide a concise ``metric card'' that summarizes its definition, value range, applicability, prerequisites, pitfalls, and recommendations, inspired by the cheat sheets for metrics on ML model quality provided by Metrics Reloaded~\cite{Maier-Hein2024,reinkeUnderstandingMetricrelatedPitfalls2024}. To facilitate access to and future extension of our library with its metric cards, we have created an online platform called Metric Hub~\cite{Metrichub}. In addition to hosting the metric library, this platform summarizes the results on the METRIC-framework and thus provides a convenient entry point to the topic of fit-for-purpose evaluation of medical ML data. To support defensible, fit-for-purpose metric choice, we design dimension-specific decision trees that translate key use case properties (e.g., data modality, variable type, \gls{ml} task, availability of reference information) into a recommended set of metrics. In contrast to existing decision-oriented frameworks such as \gls{aimq}~\cite{leeAIMQMethodologyInformation2002} and the metric-quality criteria by Heinrich and colleagues~\cite{heinrichMetricsMeasureingData2007,heinrichRequirementsDataQuality2017}, and complementing formal, theoretical approaches to data quality~\cite{Bronselaer2018}, we adopt a pragmatic, applicability-driven approach that emphasizes practical metric selection in constrained medical settings rather than detailed analysis of intrinsic metric properties. We illustrate the utility and scope of the resulting decision trees and metric library using the PTB-XL ECG dataset~\cite{wagner2020ptbxl}, where we derive a use case–specific metric set for a multiclass ECG classification task and compute the corresponding metrics for the original dataset and stratified subsets with synthetically induced perturbations of sex balance, device distribution, and target class balance. This example demonstrates how our approach enables structured, quantitative characterization of multiple data quality dimensions, 
and provides a first step towards systematic, fit-for-purpose data quality evaluation for trustworthy \gls{ai} in medicine.

In summary, our contributions are: (1) an extensive, harmonized library of 60 quantitative data quality metrics aligned with the METRIC-framework, documented via standardized metric cards, available on the website Metric Hub~\cite{Metrichub} (2) a decision-tree–based workflow that supports fit-for-purpose, model-independent, and task-specific metric selection, and (3) a demonstration on a real-world clinical dataset that showcases the feasibility and limitations of quantitative data quality assessment in practice. The presented tools and methods suggest a practically usable workflow to support fit-for-purpose data quality evaluation for medical \gls{ml} data (see Figure~\ref{fig:graphicalabstract}).

\section{Results}
\subsection*{Metric library}
\label{sec:metric-collection}
In order to build a collection of actionable metrics to quantify data quality, we reviewed the state-of-the-art of data quality metrics. 
We employed a focus-group-based process~\cite{Krueger1995FocusGroups} (see Methods), gathering expert opinions in iterative steps to collect metrics and associate them to the dimensions of the METRIC-framework.

\begin{table}[ht!]
    \input{tables/metrics-table}
    \caption{Metrics in metric library. The gray boxes indicate that a metric can be utilized to evaluate aspects of a dimension.}
    \label{tab:metricsinlib}
\end{table}

\begin{figure}[htb!]
\begin{multicols*}{2}
    \centering
    \begin{subfigure}{0.48\textwidth}
        \centering
        \includegraphics[width=\linewidth]{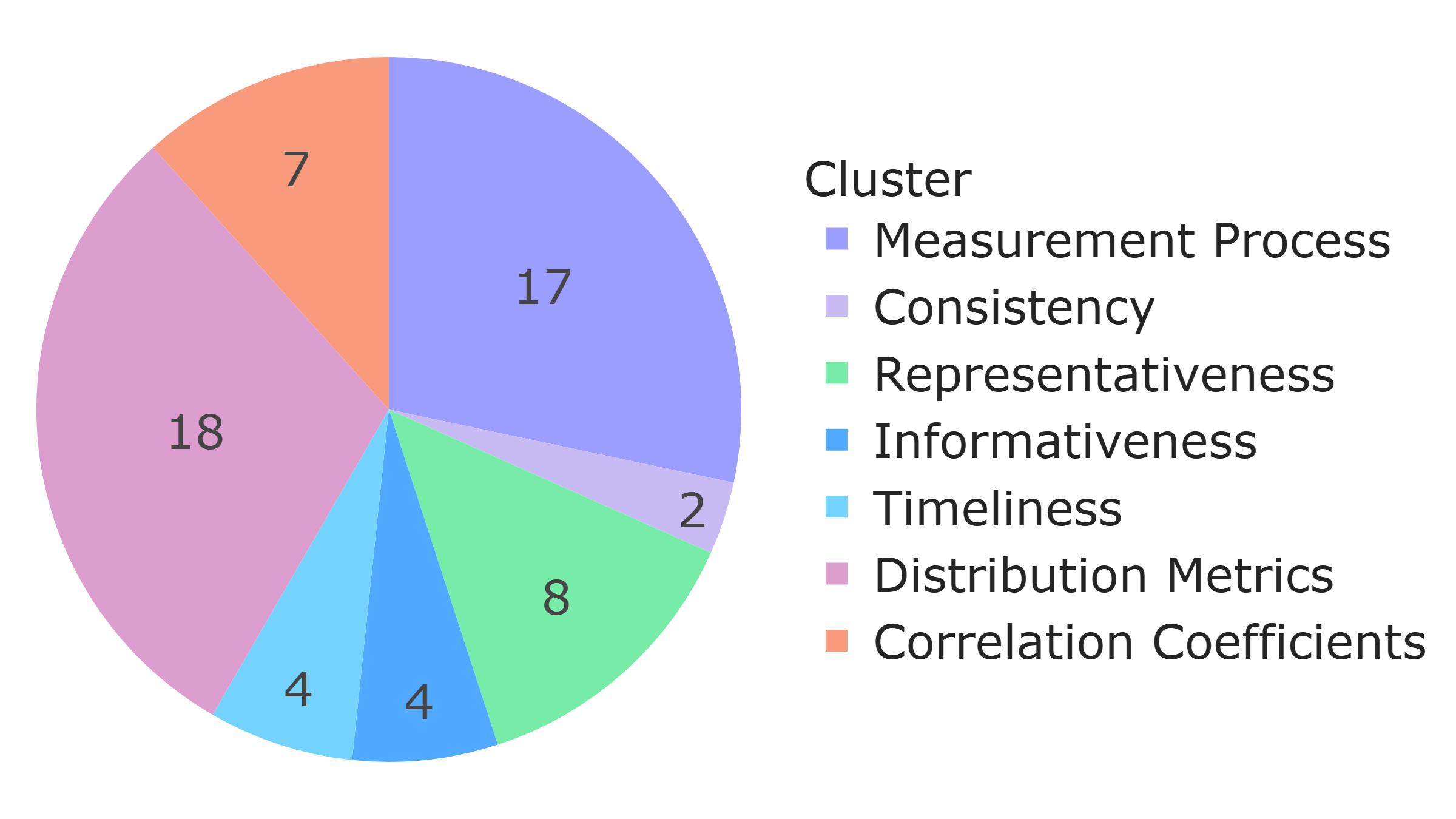}
        \caption{}
        \label{fig:metriccollection:a}
    \end{subfigure}
    \begin{subfigure}{0.48\textwidth}
        \centering
        \includegraphics[width=\linewidth]{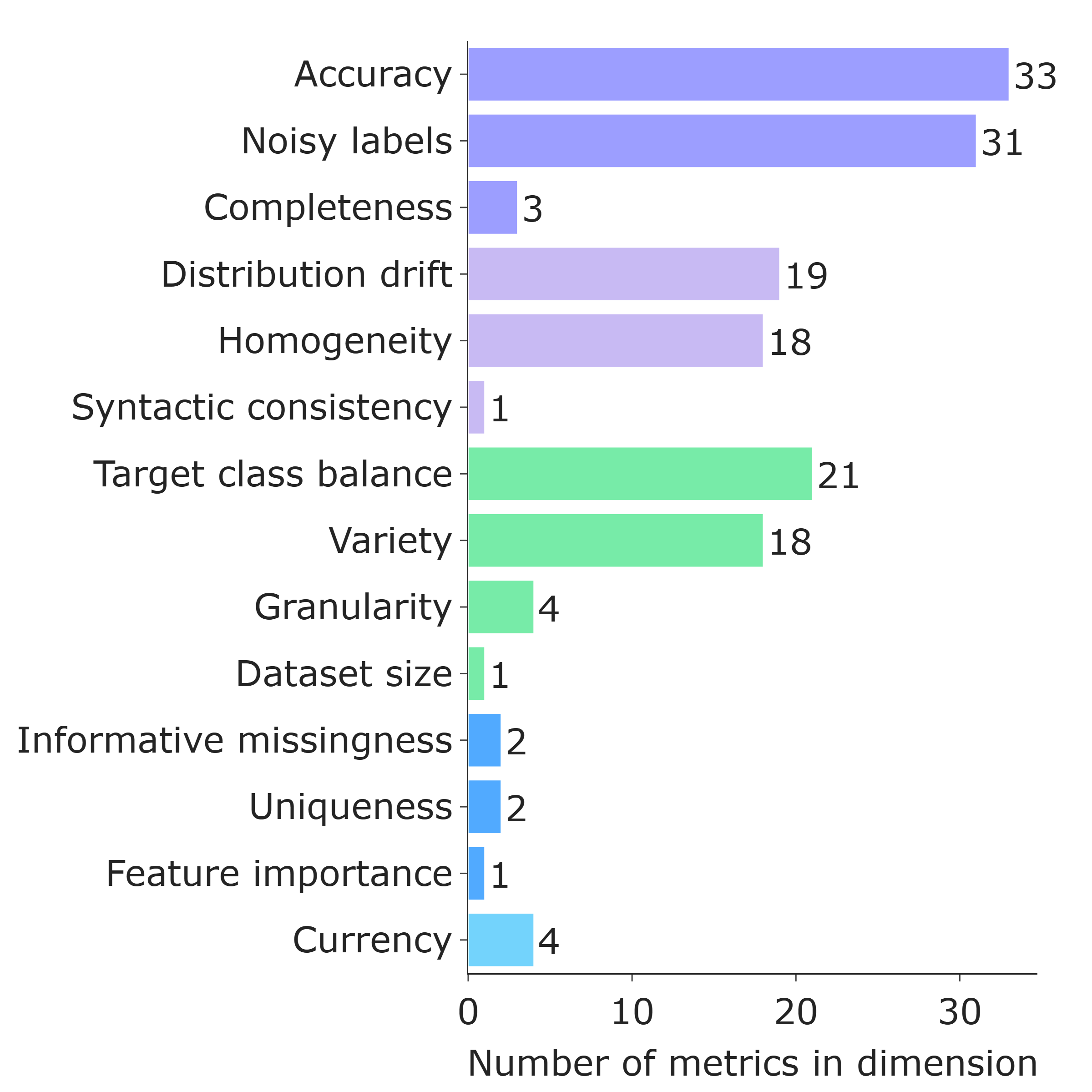}
        \caption{}
        \label{fig:metriccollection:b}
    \end{subfigure}
    
    \begin{subfigure}{0.41\textwidth}
        \centering
        \includegraphics[width=\linewidth]{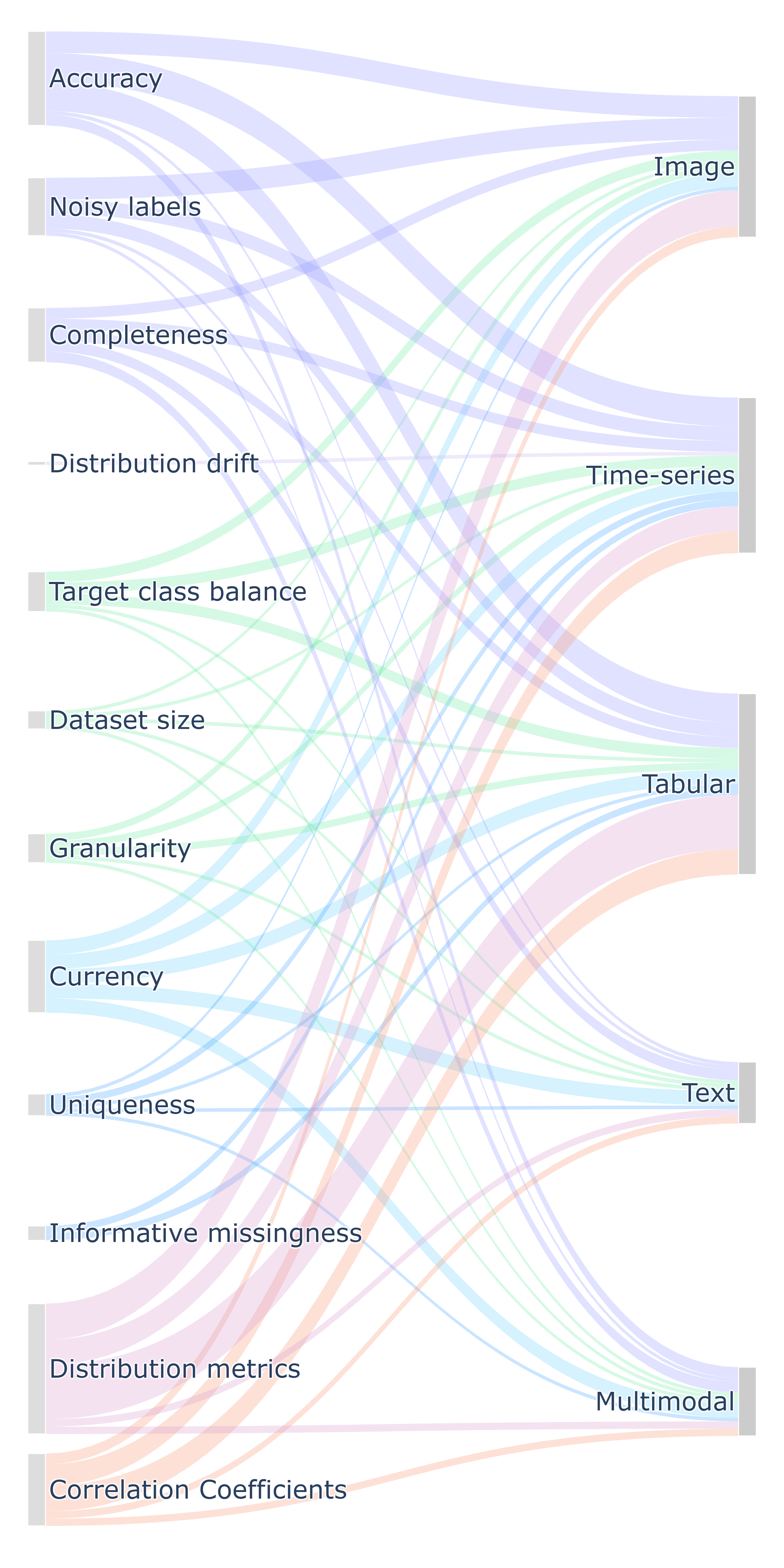}
        \caption{}
        \label{fig:metriccollection:c}
    \end{subfigure}

\end{multicols*}

    \caption{Overview of metrics in metric library. a: The number of metrics associated to the seven disjoint groups (Measurement process, consistency, representativeness, informativeness, timeliness, distribution metrics, correlation coefficients); b: The number of metrics per dimension in the METRIC-framework (with repeated counts); c: The distribution of metrics across different data modalities (tabular, image, time series, multimodal, text).}
    \label{fig:metriccollection}
\end{figure}

As a result, we obtain a comprehensive library of 60 metrics which is accessible on the associated Metric Hub website~\cite{Metrichub}.

While about half of the metrics are associated uniquely with one dimension of the METRIC-framework, other metrics can be applied to several dimensions. We summarize these more universal metrics under the groups \emph{distribution metrics} and \emph{correlation coefficients}:
\begin{itemize}
\item The six data quality dimensions accuracy, noisy labels, homogeneity, distribution drift, target class balance and variety are all entirely or in parts concerned with the comparison of \emph{distributions} of different populations. Therefore, distribution metrics can be utilized to assess these dimensions with different input populations: statistical evaluation of sample distributions (accuracy and noisy labels), comparison of sample distributions (homogeneity and distribution drift), comparison of a sample distribution to expected target distribution (target class balance and variety).
\item The three dimensions accuracy, noisy labels and feature importance are partly or entirely concerned with \emph{correlations}: correlations between repeated measurements/to gold standard (accuracy), inter-rater correlations (noisy labels), correlation between feature and target classes (feature importance). 
\end{itemize}

Thus, the metric library can be organized into seven disjoint groups of metrics (five groups relating to the clusters of the METRIC-framework and the two additional groups \emph{distribution metrics} and \emph{correlation coefficients}). Table~\ref{tab:metricsinlib} lists all individual metrics with their association to the dimensions of the METRIC-framework and the seven metric groups.

Concretely, we obtain (see Figure~\ref{fig:metriccollection:a}):
\begin{itemize}
    \item 18 metrics related to distributions~\cite{Freedman2007Statistics, Ricotta2021, Gretton2012KernelTwoSample, Panaretos2019Wasserstein, Szekely2013, Heusel2017, Binkowski2018DemystifyingMMDGANs, Cohen1988PowerAnalysis, Kullback1951Information, Nielsen2019, Kurian2024KLDivergence, Epps01121986, Scholz01091987, Mann1947},
    \item 7 metrics related to correlation coefficients~\cite{Akoglu2018,Altman1990PracticalStatistics,Bland1986Agreement, Kendall1938, Spearman1904Association, Kruskal1954},
    \item 17 metrics related to measurement process~\cite{Christodoulides2017Analysis, Bland1986Agreement, Lee2021OCTARepeatability, Shannon1948, Armbruster1994, Dice1945EcologicAssociation, Jaccard1912AlpineFlora, Gisev2013, Krippendorff2006, Blake2011, Liu2017DataCompleteness, Weiskopf2013EHRCompleteness},
    \item 2 metrics related to consistency~\cite{Batini2016,Hinkley1970},
    \item 8 metrics related to representativeness~\cite{Han2012DataMining, Kimball2016Reader, Oppenheim1978, Cole2022, Buda2018, Ortigosa2017, Zhu2018},
    \item 4 metrics related to timeliness~\cite{Ballou1998InformationQuality, Li2012, Hinrichs2002Datenqualitaet, Heinrich2007HowToMeasureDQ},
    \item 4 metrics related to informativeness~\cite{Qi2013FindDuplicates, Thompson2012Sampling, Little1988, Diggle1994}.
\end{itemize}
From the perspective of the data quality dimensions, Figure~\ref{fig:metriccollection:b} shows that the amount of available metrics per dimension varies. This is mainly due to whether distribution metrics or correlation coefficients are applicable to them. Concerning data modality, we observe that most metrics are applicable to tabular, imaging and time-series data (Figure~\ref{fig:metriccollection:c}).

\subsection*{Decision trees for metric selection}

The next objective is the practical application of use case specific metrics from the metric library for comparable data quality characterization.
Computing all metrics in the library for a given use case is neither efficient nor informative. Many metrics are context-dependent and require prerequisites, such as repeated measures, multi-rater labels, or known update frequency, that may not be available. 
We therefore propose employing a decision-oriented approach in the form of decision trees to objectively identify applicable metrics, to make reproducible, defensible choices per dimension and to complement established metric selection approaches~\cite{heinrichRequirementsDataQuality2017}. Decision trees represent a selection and their possible outcomes using a tree-like structure of questions and answers.
Use-case-specific properties, such as data modality or ML task, are encoded as nodes in the decision tree. Branches represent the possible outcomes of these properties, while individual metrics form the leaves.

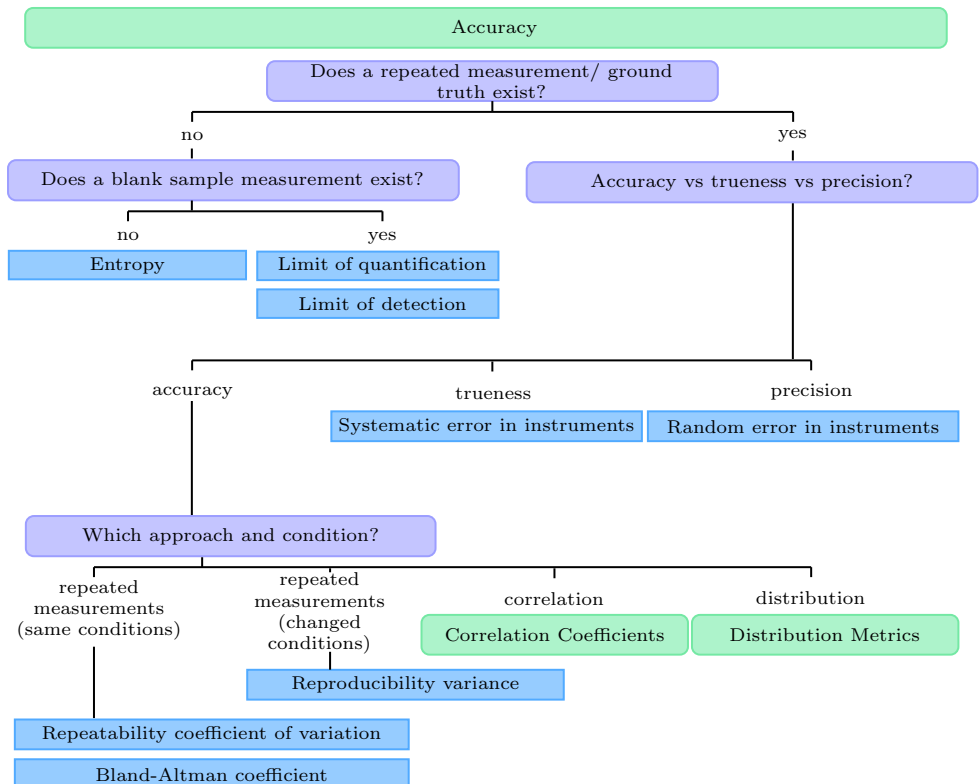
\begin{figure}[htb!]
    \centering
    \input{decision-trees/accuracy}
    \caption{Decision tree for the dimension ``Accuracy" -- A practical path to fit-for-purpose metrics per data quality dimension. The decision tree guides the user from data quality dimension to a suitable metric/ set of metrics based on the use case. The green boxes "Correlation Coefficients" and "Distribution Metrics" guide to general subtrees (see Figure~\ref{fig:decisiontree-correlation} and Figure~\ref{fig:decisiontree-distribution}).}
    \label{fig:decision_tree_accuracy}
\end{figure}

For each of the 14 quantitative dimensions of the \gls{metric}-framework we suggest one decision tree. All decision trees are presented and described in detail in the Methods section. As an example, we portray and describe the decision tree for the dimension accuracy here (see Figure~\ref{fig:decision_tree_accuracy}).
Accuracy should provide an estimate to noise in the measurement data.
Most metrics associated with accuracy require a ground truth, which can be based on repeated measurements or gold standard reference data, as is in line with international standards~\cite{ISO5725-1}. If such data is available, accuracy can be estimated using a range of metrics, such as the \emph{Bland-Altman coefficient of repeatability}~\cite{Bland1986Agreement}. However, real-world applications often lack access to ground truth data or repeated measurements. Therefore, the decision tree is organized along the central question ``Does a ground truth exist?". In the case of missing ground truth, the metric \emph{entropy}~\cite{Shannon1948} provides a quantification of signal noise, while \emph{limit of quantification (LoQ)}~\cite{Armbruster1994} and \emph{limit of detection (LoD)}~\cite{Armbruster1994} give an indication of the detection threshold of a target component for the employed method if a blank sample measurement is recorded. Besides that, all metrics associated with \emph{distribution metrics} and \emph{correlation coefficients} can be utilized to assess accuracy, e.g. by comparing distributions of repeated measurements or investigate their correlations, respectively.

The decision trees serve as a roadmap through the metric library, guiding users towards suitable metrics for a given context. 
They enhance reproducibility and auditability by providing a clear, documented path from context to chosen metrics, and they naturally scale as the library grows.
We organized the decision trees along key applicability criteria, including data modality, variable type, ML task and intrinsic metric properties. 
While some situations lead to a clear and singular metric choice, like entropy in the case where no ground truth or repeated measurement is available, others may justify the selection of multiple complementary metrics. In such cases, we recommend starting with metrics located on the left-hand side of the tree, which are typically less computationally demanding, and complementing them with more advanced metrics from the right-hand side.
Importantly, we strongly urge to keep a record of the selection rationale, configuration parameters, thresholds (if available), and monitoring cadence.

\subsection*{Metric cards}
Instead of merely providing a list of data quality metrics, we aim to enhance practical usability of the metric library. Therefore, we collected additional information for each metric in a structured process (see Methods) with the goal to provide the most essential information to understand, choose and utilize these metrics. We present them in form of cheat-sheet-style \emph{metric cards} inspired by the work of Reinke et al.~\cite{reinkeUnderstandingMetricrelatedPitfalls2024} to display the information as concise and digestible 
as possible. Similar to Metrics Reloaded~\cite{Metricsreloaded}, we have built a website (Metric Hub) containing all metric cards as well as supporting information~\cite{Metrichub}. This enables convenient access to the metric library and dynamic extension in the future. In this way, the website serves as an entry point to medical ML data quality evaluation.

\begin{figure}
\centering
\includegraphics[width=\linewidth]{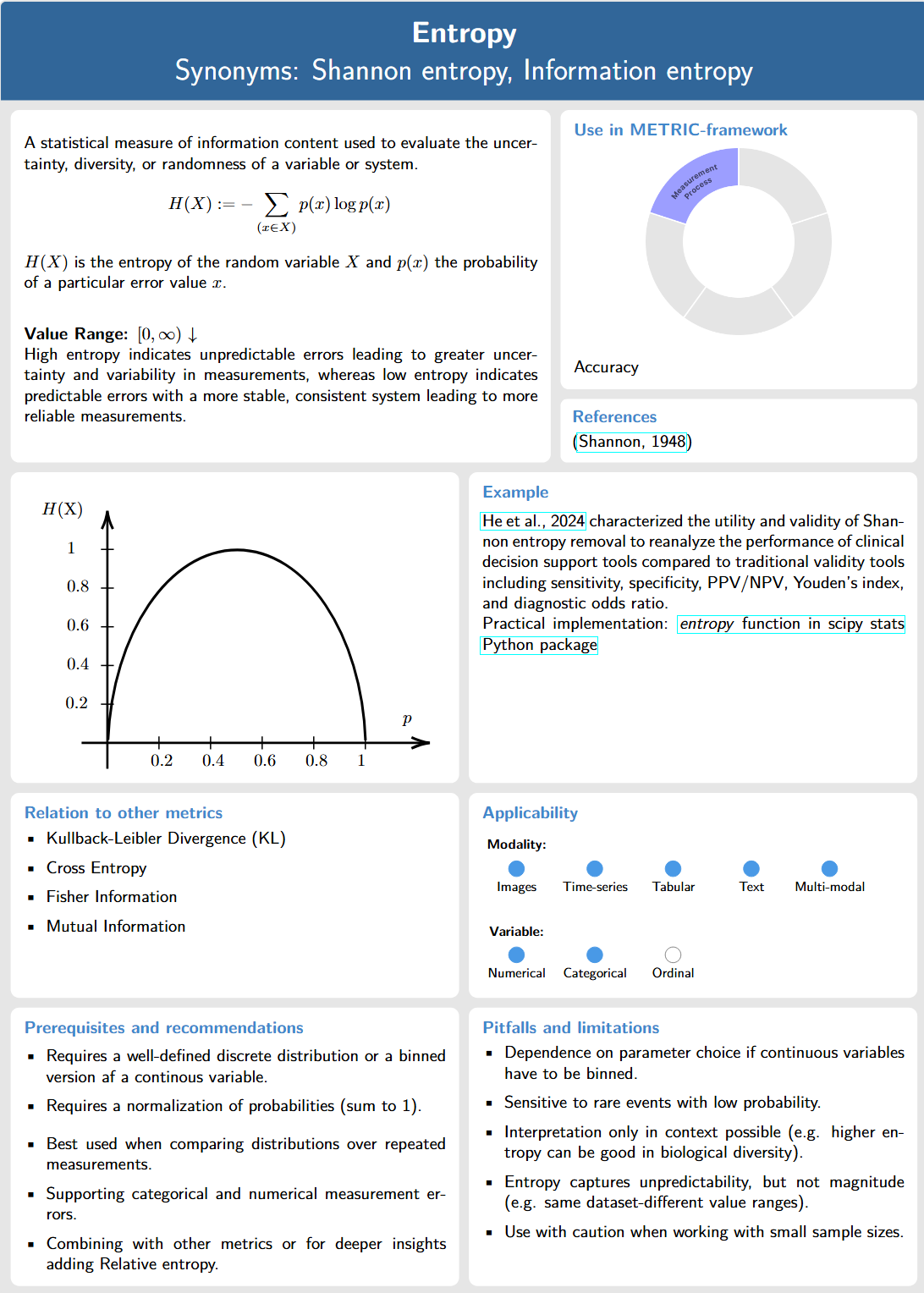}
\caption{Exemplary metric card for \emph{Entropy}. This is one of 60 cards available on our related website~\cite{Metrichub} The cards summarize important information on each metric including definition, visualization and value range, example, applicability, pitfalls and recommendations.}\label{fig:metriccard}
\end{figure}

A metric card for \emph{Entropy} is shown in Figure~\ref{fig:metriccard}, as an example and all other cards are presented in the Supplement.
Each metric card includes up to nine categories with succinct information that facilitates the metric's use and interpretation.
The card displays the name of the metric as its title together with common synonyms, if available.
The upper left-hand section provides an introduction, including a high-level summary, the mathematical definition and its value range accompanied by the meaning of high and low values to support initial understanding and interpretation.
Whenever possible, the intuitive idea of the metric is shown in a summarizing visualization below to support understandability. 

The upper right-hand section lists the data quality dimensions for which the metric can be used (\emph{Use in METRIC-framework}), followed by key references that provide technical details or implementation (\emph{References}).
If the metric has been used in literature for quantitative data quality assessment, we include an illustrative example (\emph{Example}). Since systematic quantitative data quality evaluation is an open challenge for many data quality dimensions, this category remains empty for some metrics.

Many metrics are related mathematically or represent a specialized variant. We provide this relation in the category \emph{Relation to other metrics}.
Furthermore, we indicate the applicability of the metric across different data modalities -- images, time-series, tabular, text and multimodal data -- and supported variable types -- numerical, categorical, ordinal (\emph{Applicability}). To employ certain metrics in a meaningful way, specific parameters or data characteristics are needed. This information is provided under \emph{Prerequisites and recommendations}.

Despite the seemingly unambiguous nature of quantitative results, many metrics have limitations and can yield unreliable or even meaningless outcomes when applied in unsuitable conditions. Therefore, the category \emph{Pitfalls and limitations} highlights typical sources of misinterpretation or inappropriate use, helping practitioners avoid common misconceptions.
We identified five common pitfalls across all metrics and dimensions.
A major challenge is the \emph{dependence on parameter choice}. While selecting parameters allows a metric to be tailored to the application, it might introduce additional errors and complicate the comparison of metric results between datasets.
Secondly, many metrics are \emph{sensitive to outliers}. Outliers can strongly distort the evaluation of distributions if left undetected.
Quantitative dimensions - such as variety, homogeneity, distribution drift, target class balance, noisy labels, and accuracy - that compare distributions of groups are particularly affected. 
Therefore, we strongly encourage users to screen the dataset for outliers using appropriate outlier detection methods alongside the chosen distribution metric.
Three additional common pitfalls have been identified: \emph{sensitivity to missing values}, \emph{instability for small sample sizes} and \emph{instability for imbalanced data}. These are dataset characteristics and should be considered when using affected metrics.

In summary, we recommend a lightweight workflow for selecting a use case-specific set of metrics in practice: identify the METRIC dimensions most salient for risk and intended use, follow dimension-specific decision trees to shortlist applicable metrics, implement and parametrize them using the metric cards, monitor over time for drift and quality gates, and document the selection rationale, parameters, and cadence to ensure traceability (see Figure~\ref{fig:graphicalabstract}).

\subsection*{Practical example: PTB-XL}
\label{sec:practicalexample}
To demonstrate the practical applicability of our approach, we present a case study illustrating how the decision-tree-based selection and evaluation of metrics from our library constitutes a first step towards fit-for-purpose quantification of data quality. 
For the practical evaluation, we utilize the PTB-XL dataset~\cite{wagner2020ptbxl}, a large, publicly available clinical 12‑lead ECG corpus comprising 21,837 ten‑second recordings from 18,885 patients. It is annotated with structured diagnostic statements and rich metadata, including demographic and device information. This dataset is representative of real‑world \gls{ml} applications in cardiology and exhibits diverse data quality characteristics (e.g., varying signal quality, device heterogeneity, demographic variability), making it well suited to illustrate our decision‑tree–based metric selection and evaluation workflow.
The ML task is a multiclass classification of patient diagnoses from multi-lead ECGs (e.g., myocardial infarction, hypertrophy, or normal).

To illustrate how metric values respond to synthetically induced changes in the underlying data, we built three stratified subsets from the original PTB-XL data (Table~\ref{tab:example}, column 3). 
The first subset introduces a deliberate sex imbalance with 4000 male and 1000 female patients based on a random seed (\emph{disturbed sex balance}, Table~\ref{tab:example}, column 4). 
The second subset (\emph{disturbed device balance}, Table~\ref{tab:example}, column 5) was synthetically constructed to illustrate how data quality evaluation changes across measurement devices. For this, we examined the entropy for all measurement devices with more than 1000 records. Subset 2 includes only samples measured by the same device (CS-12) which had the highest mean entropy among all examined devices.
The third subset (\emph{disturbed target class balance}, Table~\ref{tab:example}, column 6) incorporates a modified target class distribution achieved by choosing 250 samples of healthy patient data (``norm" class) and 4750 samples from different classes.

First, we systematically derive an appropriate set of data quality metrics by traversing the decision tree associated with each quantitative data quality dimension in the METRIC-framework. This metric selection step is the same for the original and all subsets.
Some dimensions -- such as completeness and granularity -- contain only a small number of metrics that should be evaluated whenever possible and applicable to the data. Others, however, include metrics suitable for different situations, as is the case for accuracy. 

To determine the appropriate metric for accuracy exemplarily, we progress through the associated decision tree (Figure~\ref{fig:decision_tree_accuracy}).
The tree starts with the question “Does the ground truth exists?”. Since no ground truth exists, the user is next asked whether a blank-sample measurement is available, if so, metrics such as limit of detection (LOD) and limit of quantification (LOQ) apply. Otherwise, entropy is suggested. If a ground truth would exist (right branch in Figure~\ref{fig:decision_tree_accuracy}), the path proceeds to agreement assessments appropriate to the study design in a hierarchical manner.
For the PTB-XL dataset, the tree selects entropy as the only computable proxy for measurement noise. 
We expect this to be a common outcome of accuracy metric selection when studying medical datasets, where reference measurements and blank‑sample controls are unavailable, rendering classical accuracy metrics based on ground truth or repeatability inapplicable. 
Concretely, we use the sample entropy~\cite{Richman2000} which has proven useful for physiological time-series. 

In total, this procedure yields 16 metrics (across all quantitative dimensions) which are listed in the ``Metric" column, column 2, of Table~\ref{tab:example}. 
Some metrics are computed on the measurement data, here ECG time series, and some rely purely on metadata, such as patient information. Additionally, some metrics can be applied to individual features of the data and thus be computed many times (e.g., variety metrics such as Hill numbers). Therefore, we indicate with respect to which feature or property the metric was calculated in brackets behind the metric name.
Once the metrics are selected, their implementation is supported by the metric cards, which provide details such as value range and guidance for interpretation.

Next, we calculate the selected data quality metrics for the datasets, yielding quantitative indicators for each data quality dimension (Table~\ref{tab:example}, columns 3-6). We employed the tool MetricLib~\cite{seyferth_metriclib_2025} to compute the results,
by rounding values to two decimal places.
We emphasize that, at this stage, the evaluation of metrics does not yield a universally interpretable score that would allow a definitive rating of a data quality dimension or dataset, as no thresholds are established. However, metric values close to their theoretical minimum or maximum can still provide an indication of the cause and potential severity of a problem, and help prioritize related issues for further investigation or remediation. Furthermore, our approach can be applied to compare different datasets, even when the values of the metrics are not close to their theoretical minimum or maximum.

The results for the dataset \emph{PTB-XL original} are presented in Table~\ref{tab:example}, column 3. Below, we highlight the metric outcomes that yield particularly relevant insights; the remaining metrics either reflect expected results for this dataset or do not permit meaningful interpretation when considered in isolation:
\begin{itemize} 
\item \emph{Completeness:} Measurement data is complete (100$\%$); metadata completeness appears relatively low (37$\%$). However, the observed low metadata completeness does not necessarily indicate poor quality, as it reflects the extensive and diverse list of 26 features considered.
\item \emph{Noisy labels:} Intended to capture the noise in the labels provided by annotators and crucial as ML models will learn and propagate this noise through their predictions. 
Although PTB-XL provides up to two annotations per record, the prevalence of single-annotator labels limits the applicability of metrics for this dimension.
Addressing this gap is an important direction for future work. 
\item \emph{Accuracy:} Assessed via entropy because no reference measurements exist. While entropy is not a classical metric for accuracy, it can provide an indication of the noise distribution in the data. 
Since this scenario is expected to be common in real-world applications, we believe that methods, like entropy evaluation, are needed to explore noise without ground truth or repeated measurements and identify systematic issues with the data recording procedure or instrument.
A low mean entropy across all patients ($0.10$), as observed, may suggest that the measurement data are relatively clean. While this is beneficial in reducing noise introduced to the model, overly clean data can be suboptimal, since machine-learning models require sufficiently challenging examples for robust learning~\cite{ZhengLLP23}.
\item \emph{Currency:} Currency by Heinrich (value range: $[0,1]$) appears relatively low with $0.36$; note the dependence on parameter choice ($decline(A)=10^{-9}$).
\item \emph{Variety:} 
The range for the continuous feature ``age" is $298$ years. All patients with age greater than $99$ years were set to $300$ in the original dataset due to data protection. This preprocessing influences the metric value and highlights the importance of outlier detection.
The feature ``sex" has a perfect distribution of male and female types (Hill number is $2.00$).
\end{itemize}

Investigating the stratified subsets (Table~\ref{tab:example}, column 4-6), we observe that, as expected, metrics associated with the deliberately disturbed attributes -- sex, device, and target-class distributions -- show clear changes across the subsets, while metrics unrelated to these attributes remain largely unchanged. 
The metric used to assess the sex and device balance is the Hill number. For the feature ``sex", the Hill number is close to $2$ for all sets except for the \emph{disturbed sex balance} subset, indicating an almost perfect sex distribution in the other three datasets.
In the \emph{disturbed sex balance} dataset, the metric decreases to $1.47$.
Similarly, the Hill number for ``device" remains relatively stable across the original, first and third subsets ($5.59$, $5.13$, $5.04$) but is notably lower for the second one ($1.00$). 
The generalized imbalance ratio ranges from $3.24$ to $4.41$ for the original and the first two subsets, increasing to $8.77$ for the \emph{disturbed target class balance} dataset. This ratio has a value range from $1$ to infinity, with $1$ representing perfect class balance, indicating that the class balance is worse in the last subset. 
These changes in the metric values confirm that our approach is sensitive to meaningful variations in the data.

Examining the metric evaluation for subset 2 (synthetically disturbed \emph{disturbed device balance}) more closely, we note that, while the mean entropy increased as expected, the metric results for target class balance and variety of ``age" change simultaneously. 
This might be an indication that the relative increase in entropy is not due to the device itself but to specific classes and/or patient characteristics. 

The example illustrates the practical impact of our approach on systematic data quality evaluation. By quantifying all relevant dimensions, we obtain an initial, comprehensive view of the dataset’s quality from multiple perspectives. 
At the same time, the analysis highlights key challenges, including the absence of well-defined thresholds for standardized assessment, gaps in the metric library (e.g., metrics for noisy labels with single labeler), and the need to understand interactions between dimensions. 
Despite these limitations, the metric library already enables meaningful comparisons between datasets under specific assumptions, such as subsets derived from the same source. 
Such analysis could support designing and composing high quality test data for ML model assessment which is a conceivable implementation of our approach.

\begin{landscape}
\begin{table}[ht]
  \begin{tabular}{| c | c || c | c || c || c || c ||}
    \hline
    &&\textbf{ PTB-XL original} & \textbf{PTB-XL subset 1} & \textbf{PTB-XL subset 2} & \textbf{PTB-XL subset 3}\\
    \textbf{Cluster and Dimension} & \textbf{Metric} &  & (disturbed sex balance) & (disturbed device balance) & (disturbed target class balance) \\ 
    \hline
    \hline
    \textbf{Measurement process}  & &  &  &&\\ 
     Completeness & Completeness (measurements) & 100\% & 100\% & 100\% & 100\%\\ 
      &Completeness (metadata) & 37\% & 37\% & 37\% & 36\%\\ 
      &Patient-level completeness (measurements) & 100\% & 100\% & 100\% & 100\%\\ 
     Noisy labels & - &-&-&-&-\\ 
     Accuracy & Entropy (measurements) &  0.10 & 0.10 & 0.13 & 0.11 \\ 
    \hline
    \hline
    \textbf{Timeliness}  & &  &  &&\\ 
     Currency & Currency by Heinrich (all) & 0.36 & 0.37 & 0.36 & 0.37 \\ 
    \hline
    \hline
    \textbf{Representativeness}  & &  &  &&\\ 
    Target class balance & Generalized imbalance ratio (labels) & 3.59 & 3.24& 4.41 & 8.77 \\
      Granularity  & Detail of data (metadata) & 26 columns & 26 columns & 26 columns & 26 columns\\
      & Sampling frequency (measurements) & 500Hz & 500Hz & 500Hz & 500Hz \\
      Dataset size & Dataset size (all) & 21837 records & 5000 records & 4048 records & 5000 records \\
    Variety  & Range (age) & 298 & 295 & 278 & 290 \\
      & Mean and standard deviation (age) &(62.77, 32.31)& (60.84, 27.90)& (70.88, 37.70)& (69.71, 36.06)\\
      & Hill number (sex) & 2.00 & 1.47 & 1.98 & 1.97 \\
      & Hill number (device) & 5.59 & 5.13 & 1.00 & 5.04 \\
    \hline
    \hline
    \textbf{Informativeness}  & &  &  & &\\ 
     Feature importance & PCC (age) & 0.17 & 0.15 & 0.14 & 0.08\\
     Uniqueness & Prevalence of duplicates (all) &0 duplicates &0 duplicates & 0 duplicates & 0 duplicates \\
    \hline
    \hline
    \textbf{Consistency}  & &  &  &&\\ 
     Homogeneity & Maximum mean discrepancy (sex) & 0.17 & 0.22 & 0.16 & 0.33 \\     
    \hline
\end{tabular}
\caption{Evaluation of selected set of metrics for PTB-XL on the original dataset and three synthetically disturbed subsets (disturbed sex balance, disturbed device balance,  disturbed target class balance).}\label{tab:example}
\end{table}
\end{landscape}

\section{Discussion}
This work provides a first point of reference to conduct a practical, adequate and use-case-specific data quality assessment. We operationalized the theoretical METRIC-framework~\cite{schwabeMETRICframeworkAssessingData2024} by providing a structured metric library and a decision-tree–based approach for data quality metric selection.

To facilitate adoption in practice, we made the metric library easily accessible via the Metric Hub platform~\cite{Metrichub}. We complemented it with concise, cheat-sheet–style metric cards that support the practical implementation of individual metrics, thereby improving accessibility, particularly for users without extensive prior expertise in data quality research.
A central goal of this work was to translate data quality assessment from theory into practice. 
The Metric Hub is intentionally designed to be extendible, allowing the library to evolve alongside emerging methodological developments in medical \gls{ml} and to foster collaborative research in this dynamic field.

While the metric library and cards support systematic and harmonized quality evaluations, quantifying all available metrics for a given use case is neither efficient nor meaningful. The suitability of individual metrics depends strongly on the characteristics of the use case, making use-case–specific metric selection essential for valid, interpretable, and actionable assessments.
Several approaches to metric selection have been discussed in literature. For instance, Heinrich’s framework~\cite{heinrichRequirementsDataQuality2017}, which is one of the most prominent in the field of data quality metrics, defines requirements R1–R5 that metrics should satisfy to ensure theoretical soundness. 
Applying these requirements to \emph{entropy} illustrates the limitations of a strictly rule-based selection process. While entropy satisfies some requirements (e.g., bounded extrema, interval scale, computational efficiency), other criteria, such as sound aggregation (R4) and quality of parametrization and value determination (R3), are only partially or conditionally met. From a strict Heinrich perspective, this could lead to discarding entropy as a metric. Yet, in practice, entropy can provide meaningful insights into measurement noise and signal complexity, particularly when no ground-truth or blank-sample measurements are available.
Motivated by this gap between theoretical rigor and practical applicability, we developed structured decision trees to guide metric selection. Decision trees allow users to incorporate contextual knowledge about their use case, emphasizing actionable relevance over strict adherence to formal rules. By providing decision trees for each quantitative dimension, we ensure that metrics like entropy can be retained as pragmatic proxies where appropriate, enhancing usability and interpretability of the library without compromising systematic evaluation. 
While we acknowledge that this approach may include metrics that do not evaluate dimensions in the classical sense, we argue that adopting a broader perspective can encourage discussion and facilitate the exploration of data quality from angles that strictly rule-based approaches may overlook due to a lack of appropriate tools.

We do not claim that the current metric library or decision trees provide a complete or exhaustive collection.
Instead, they are intended as living artifacts that evolve alongside emerging metrics and evidence, supporting the assessment of datasets for specific \gls{ml} tasks in medicine.
They serve as a pragmatic starting point and reference for developers, assessors, and auditors to systematically evaluate the trustworthiness of their \gls{ml} systems, while remaining adaptable to new insights and evolving standards.
A major focus for future work is the application of this methodology to diverse use cases, which will help identify gaps and refine both the library and the decision trees.
Extending the library should primarily focus on the implementation or development of metrics that address gaps in scenarios we expect to be common in medical \gls{ml} development. These typically involve situations where reference information is limited or unavailable -- for example, the absence of a second annotator for assessing noisy labels or a ground truth for evaluating accuracy. Addressing such cases will be crucial for supporting realistic use cases.

We note that multiple aspects for a practical data quality evaluation remain unaddressed.
One such aspect is the determination of thresholds.
To move from metric calculation and data assessment to actionable evaluation, it is necessary to answer the question: when is a metric result ``good” or ``bad”?
Identifying potential tipping points in data quality beyond which the behaviour of the resulting system deteriorates is essential. To achieve this, it is necessary to systematically investigate how specific data quality dimensions affect the overall performance of a \gls{ml} system. We note that since most data quality metrics provide relative measures -- dependent on dataset-specific properties, their primary value lies in within-dataset comparisons. Investigating absolute quality measures for cross-dataset and cross-use case comparability is an open challenge. However, since the impact of each dimension is likely to vary across use cases, overarching fixed thresholds are unlikely to exist. Nonetheless, establishing general ranges or best-practice guidelines may provide valuable guidance for practical applications and should be addressed in future work.
In addition to understanding individual metric results and the effects of data quality on \gls{ml} models, it is crucial to develop methods for aggregating these metrics into higher-level quality scores that characterize the overall quality of a dimension, cluster, or even an entire dataset. Such composite scores could provide a more intuitive and practical basis for data quality evaluation and significantly simplify communication with end users. 
Multiple pathways are conceivable, ranging from a weighted average with use case-specific weights, to a traffic light system based on minimum thresholds per metric, to a Nutri-Score (Regulation (EU) No 1169/2011) style evaluation classifying the data quality into multiple ranked categories.
A key challenge in the direction of data quality threshold and scores is to understand interactions and dependencies between data quality dimensions. To analyse the impact of a specific dimension on the \gls{ml} system, which is essential for determining an appropriate threshold, it is necessary to ensure that its effect can be isolated from the effects of other dimensions.

The presented tools translate regulatory requirements into specific data quality dimensions, provide actionable metrics to measure these dimensions, and propose methods for use case–specific evaluation of medical training and test data. By operationalizing these concepts, they offer practical instruments for quantitative evaluation of medical \gls{ml} datasets.
Such evaluations support multiple stages of the \gls{ml} lifecycle: they inform dataset design during product development, contribute to technical documentation for conformity assessment, harmonize regulatory audits, and enable post-market monitoring. In combination with rigorous \gls{ml} model testing, quantitative data quality assessments form the backbone of evidence-based quality assurance, ultimately enhancing the trustworthiness of medical \gls{ml} systems and facilitating compliance with regulatory requirements, including the AI Act.

\section{Methods}\label{sec:methods}
\subsection*{Theoretical foundations}
\begin{figure}
\centering
\includegraphics[scale=0.1]{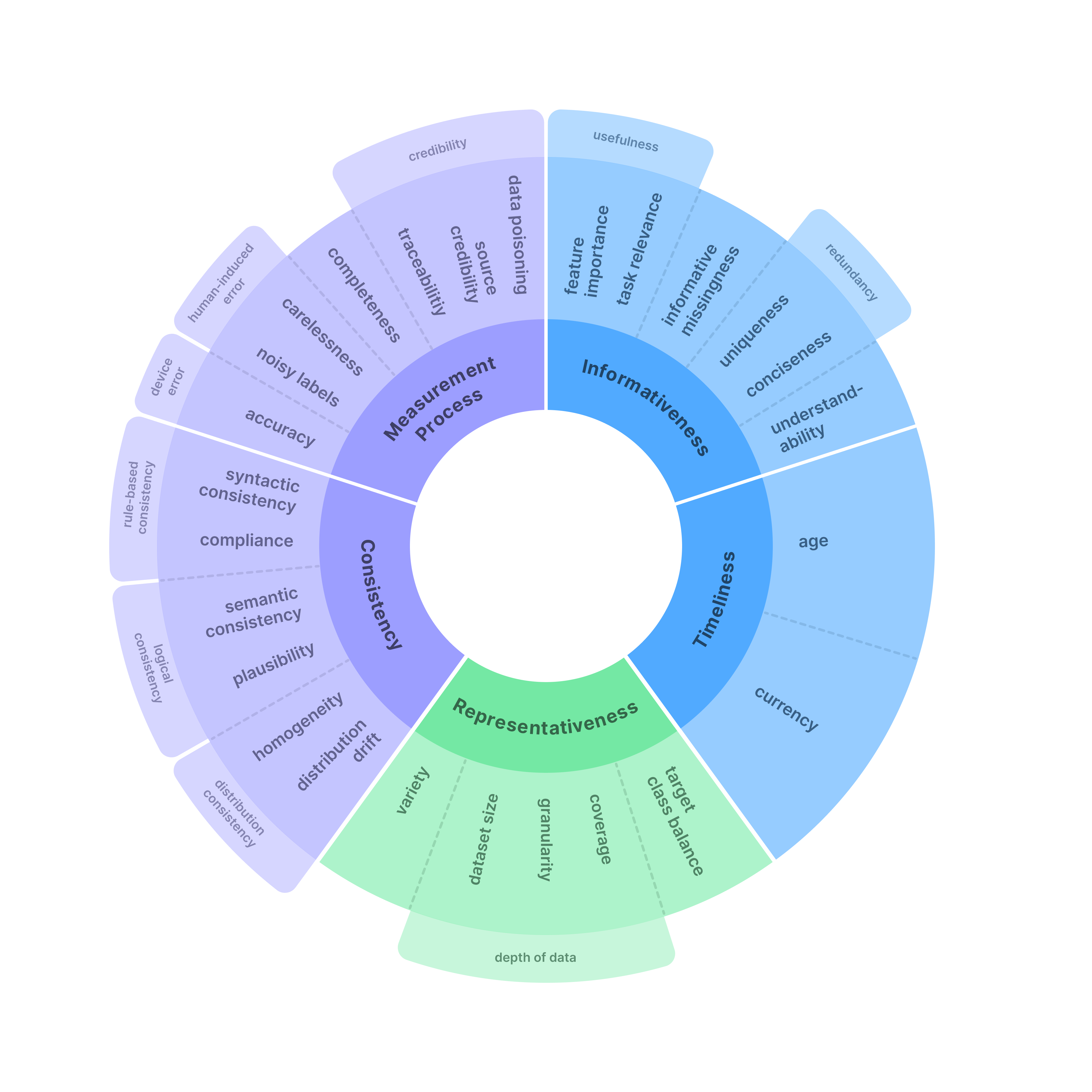}
\caption{Revised version of the METRIC-framework for systematic ML data quality evaluation in medicine. The inner circle divides data quality into five clusters. These clusters contain a total of 26 data quality dimensions, which are shown on the outer circle. Related dimensions are summarised into groups at the borders.}\label{fig:metricwheel}
\end{figure}

The theoretical basis for this work is the METRIC-framework~\cite{schwabeMETRICframeworkAssessingData2024}, which specifies a set of characteristics, referred to as awareness dimensions. In Figure~\ref{fig:metricwheel}, we present a revised version of this framework with an updated visual design and several structural refinements. In particular, dimensions and subdimensions have been consolidated into a single layer: dimensions that previously included multiple subdimensions now group conceptually related dimensions, and the former subdimensions have been elevated to the dimension level. These modifications are primarily organizational rather than conceptual, and are intended to simplify the framework and improve the clarity with which it can be communicated:
\begin{itemize}
    \item \emph{accuracy} and \emph{precision} is collapsed to accuracy to remove redundancy with respect to accuracy, precision and trueness,
    \item \emph{outliers} as a dimension is removed since outliers is an aspect of multiple dimensions rather than a dimension itself,
    \item \emph{expertise} is renamed to source credibility for clarity,
    \item \emph{credibility} replaces source credibility as a group name,
    \item \emph{variety} in demographics and variety in data sources are combined to variety,
    \item \emph{task relevance} is added as a dimension so that \emph{usefulness} is now represented by a qualitative dimension, namely task relevance, and a quantitative dimension, namely feature importance.
\end{itemize}

For operationalizing the theoretical METRIC-framework, we for now focus on those dimensions that can be evaluated using \emph{mostly quantitative} methods and call them quantitative dimensions (see Table~\ref{tab:quantitativedim}). We evaluate these quantitative dimensions with the help of data quality metrics: quantifiable indicators that measure a specific property, characteristic, or dimension of a dataset as a basis to assess its suitability for intended use (fit-for-purpose)~\cite{ehrlingerSurveyDataQuality2022, orrDataQualitySystems1988}. A metric should be quantifiable (numeric, categorical, or ordinal; computed systematically and reproducibly), property-focused ~\cite{heinrichProcedureDevelopMetrics2009}, purpose-driven (relevant, interpretable, and actionable for the application context and stakeholders)~\cite{parssianAssessingDataQuality2004}, and context-dependent (meaning and importance vary by domain and use; proper interpretation requires domain knowledge)~\cite{leeAIMQMethodologyInformation2002, heinrichRequirementsDataQuality2017}.

\subsection*{Metric collection and creation of metric cards}
For the collection of metrics and creation of the metric cards, we adopted a focus-group process in two consecutive steps, following established qualitative research practices~\cite{Krueger1995FocusGroups}.
Quantitative evaluation of data quality for ML is an emerging field with limited methodological foundations making expert opinion the most reliable source of validation. We therefore employed this approach to enable participants to independently reflect on the topic while gradually building a shared understanding.

Our focus group consisted of 20 international researchers contributing to the ``Standards and Quality" work package within the TEF-Health consortium. The composition of the focus group was interdisciplinary, with experts from different backgrounds, such as computer science, ML engineering, digital health, and AI Ethics. This diversity facilitated the integration of multiple perspectives and domain understandings, helping ensure that the resulting set of metrics reflects a broad and well-informed conceptualization of each data quality dimension. 
Metric collection was conducted in four iterative rounds. For this, the focus group was divided into five subgroups, with each subgroup assigned to one METRIC-framework cluster. 
In the first round, each subgroup independently identified candidate metrics for its assigned cluster, drawing on expert experience and targeted literature searches. In the second and third round, subgroups rotated between clusters to review, refine, and expand the collected metrics of the other subgroups. 
Disagreements were resolved through structured discussion or, when needed, escalated to the core group. All iterations were documented in a shared tracking system to ensure traceability.
In a final synthesis step, the core group reviewed the full set of collected metrics and associated them across dimensions where appropriate. This consolidated version was subsequently reviewed again by the full focus group.

Following metric collection, we developed metric cards to enhance the usability of the resulting metric library. An open workshop following the principles as presented by Krueger, was held with the entire focus group to brainstorm potential card categories and assess their anticipated value for end-users~\cite{Krueger1995FocusGroups}.

Each expert subsequently provided an independent vote on whether to include each category, producing a quantitative rating. 
In a second workshop, we qualitatively reviewed all categories, considered the voting results, and agreed on a cut-off (minimum 7 votes) in the ratings to determine the final set of categories to include in the metric cards.
After establishing the categories, each subgroup populated the metric cards for their assigned quality dimension based on their expertise and relevant literature. This was followed by two iterative review rounds with rotated subgroup assignments to ensure thorough cross-evaluation.
To achieve a harmonized and coherent library, we then transitioned from a vertical (cluster-based) to a horizontal (category-based) review approach. Subgroups were reorganized by category and reviewed that category across all metrics and clusters. This horizontal review was repeated in two additional rounds to refine content, resolve discrepancies, and ensure consistency across the full collection.

\subsection*{Metric library and decision tree description}
In the following, we give an overview of the metrics included in the metric library (sorted by dimensions of the METRIC-framework) and how they are organized within the decision trees to facilitate use-case-specific selection metric.

\subsubsection*{Measurement process cluster}
The \textbf{measurement process cluster} includes three quantitative dimensions (see Table~\ref{tab:quantitativedim}), which are accuracy, noisy labels and completeness.

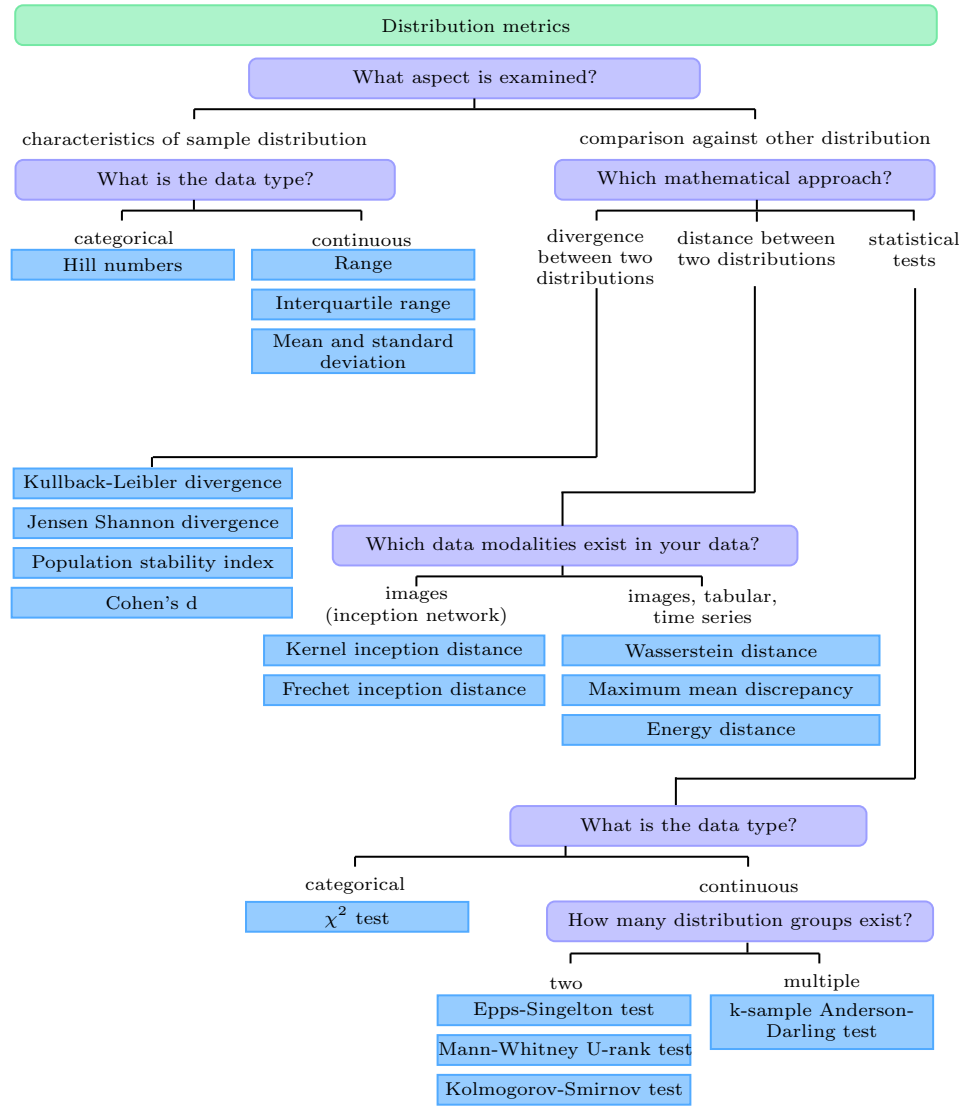
\begin{figure}[htb!]
  \centering
    \input{decision-trees/distribution-metrics}
  \caption{Decision tree for ``Distribution Metrics". This tree is a subtree which can be used for various dimensions.}\label{fig:decisiontree-distribution}
\end{figure}

\paragraph*{Accuracy}
Metrics solely associated with \emph{accuracy} are presented in blue in the corresponding decision tree in Figure~\ref{fig:decision_tree_accuracy}. Accuracy should provide an estimate to noise in the measurement data. Most metrics associated with accuracy require a ground truth, which can be based on repeated measurements or gold standard reference data, as is in line with the International standards~\cite{ISO5725-1}. If such data is available, accuracy can be estimated using a range of metrics, such as the \emph{Bland-Altman coefficient of repeatability}~\cite{Bland1986Agreement}. However, real-world applications often lack access to ground truth data or repeated measurements. Therefore, the decision tree is organized along the central question ``Does a ground truth exist?". In the case of missing ground truth, the metric \emph{entropy}~\cite{Shannon1948} provides a quantification of signal noise, while \emph{limit of quantification (LoQ)}~\cite{Armbruster1994} and \emph{limit of detection (LoD)}~\cite{Armbruster1994} give an indication of the detection threshold of a specific analyte for the employed method if a blank sample measurement is recorded.
Besides that, all metrics associated with \emph{distribution metrics} (Figure~\ref{fig:decisiontree-distribution}) and \emph{correlation coefficients} (Figure~\ref{fig:decisiontree-correlation}) can be utilized to assess accuracy as indicated in the decision tree.
    
\paragraph*{Noisy labels}
Metrics for the dimension \emph{noisy labels} are applicable only when annotations from multiple raters are available, which we made explicit with the first question ``How many annotators have labeled your dataset?". If multiple raters exist, noisy labels can be quantified using classical metrics, such as \emph{Cohen’s kappa}\cite{Gisev2013}) (for classification with two raters) or \emph{Fleiss’ kappa}~\cite{Gisev2013} (for classification involving more than two raters). Besides that, noisy labels can be assessed using \emph{distribution metrics} (Figure~\ref{fig:decisiontree-distribution}) to compare the distributions of ratings and \emph{correlation coefficients} (Figure~\ref{fig:decisiontree-correlation}) to quantify their correlation.
    
\begin{figure}[htb!]
  \centering
    \input{decision-trees/completeness}
  \caption{Decision tree for the dimension ``Completeness".}
  \label{fig:decision-tree_completeness}
\end{figure}
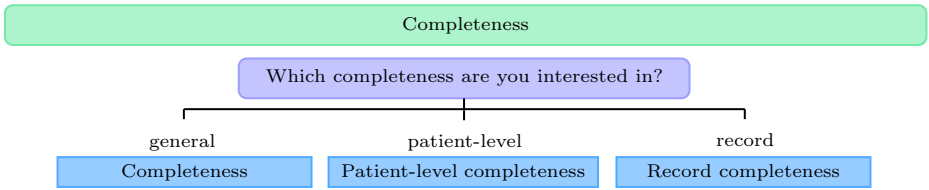

\paragraph*{Completeness}
The dimension \emph{completeness} is quantified by three metrics, depending on the quantity of interest (Figure~\ref{fig:decision-tree_completeness}). The metric \emph{dataset completeness}~\cite{Blake2011} calculates the ratio of missing data values to total data values. The metric \emph{patient-level completeness}~\cite{Liu2017DataCompleteness} quantifies how many patients exist with at least one complete entry for a given variable while \emph{record completeness}~\cite{Weiskopf2013EHRCompleteness} quantifies the number of complete records. We recommend to calculate all of these metrics, if computationally feasible, to get a broader view of the dimension.

\subsubsection*{Consistency cluster}
For the three quantitative dimensions in the \textbf{consistency cluster} -- syntactic consistency, homogeneity, distribution drift -- the amount of cluster-specific metrics is low, since most of its quantitative dimensions are concerned with distribution consistency.

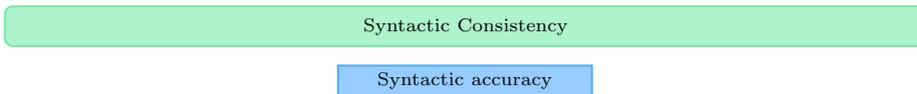
\begin{figure}[htb!]
  \centering
    \input{decision-trees/syntatic-consistency}
  \caption{Decision tree for the dimension ``Syntactic Consistency".}
  \label{fig:decisiontree-syncons}
\end{figure}

\paragraph*{Syntactic accuracy}
The metric \emph{syntactic accuracy}~\cite{Batini2016} estimates \emph{syntactic consistency} by counting how many entries are part of an internal dictionary (Figure~\ref{fig:decisiontree-syncons}).

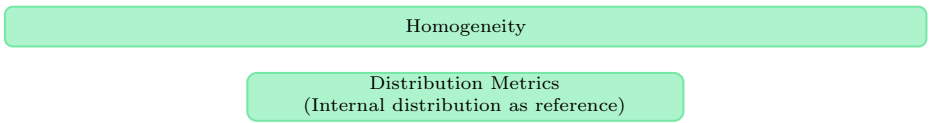
\begin{figure}[htb!]
  \centering
    \input{decision-trees/homogenity}
  \caption{Decision tree for the dimension ``Homogeneity".}
  \label{fig:decisiontree-homogeneity}
\end{figure}

\paragraph*{Homogeneity}
The dimension \emph{homogeneity} is concerned with evaluating whether the distributions of multiple subclusters (e.g., subsets with respect to demographic information) differ. All distribution metrics can be utilized for that purpose (Figure~\ref{fig:decisiontree-homogeneity}). 

\begin{figure}[htb!]
  \centering
    \input{decision-trees/distribution-drift}
  \caption{Decision tree for the dimension ``Distribution Drift".}
  \label{fig:decisiontree-distributiondrift}
\end{figure}
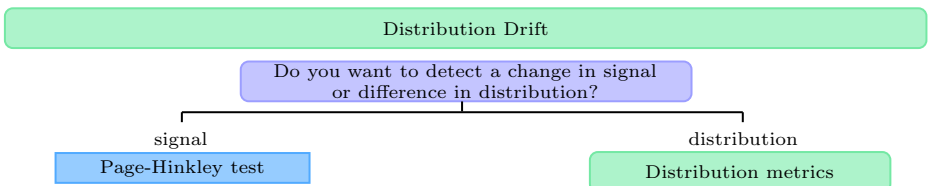

\paragraph*{Distribution drift}
\emph{Distribution drift} measures how the data distribution changes over time. Accordingly, instead of comparing internal subsets, the dataset is split into time intervals and distributions can be compared using the \emph{distribution metrics} (Figure~\ref{fig:decisiontree-distributiondrift}). A specific method for this purpose is the \emph{Page–Hinkley test}~\cite{Hinkley1970}, which detects changes in sequential data.

\subsubsection*{Representativeness cluster}
Representativeness assesses the extent to which a dataset reflects the diversity, distributional characteristics, and structural properties of the target population. Within the METRIC-framework, the \textbf{representativeness cluster} contains four quantitative dimensions -- variety, dataset size, granularity and target class balance.

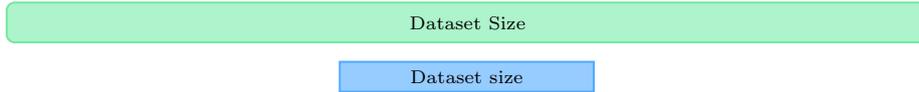
\begin{figure}[htb!]
  \centering
    \input{decision-trees/dataset-size}
  \caption{Decision tree for the dimension ``Dataset Size".}
  \label{fig:decisiontree-datasetsize}
\end{figure}

\paragraph*{Dataset size}
Since \emph{dataset size} is a straightforward dimension, it is equipped with a single metric providing the total number of records~\cite{Han2012DataMining} (Figure~\ref{fig:decisiontree-datasetsize}).

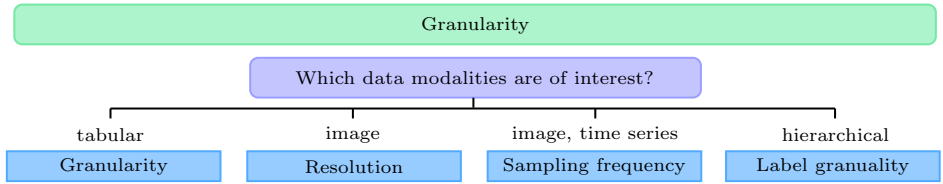
\begin{figure}[htb!]
  \centering
    \input{decision-trees/granularity}
  \caption{Decision tree for the dimension ``Granularity".}
  \label{fig:decisiontree-granularity}
\end{figure}

\paragraph*{Granularity}
The dimension \emph{granularity} should capture the level of detail encoded in the data. The collected metrics can be divided along the type of data available (see Figure~\ref{fig:decisiontree-granularity}): \emph{granularity}~\cite{Kimball2016Reader} for tabular data, simply counting the number of features; \emph{sampling frequency}~\cite{Oppenheim1978} capturing the temporal or spatial resolution; \emph{resolution} for imaging data~\cite{Cole2022}. The fourth metric, \emph{label granularity}~\cite{Cole2022}, captures the depth of a hierarchical structure, e.g. the label space.

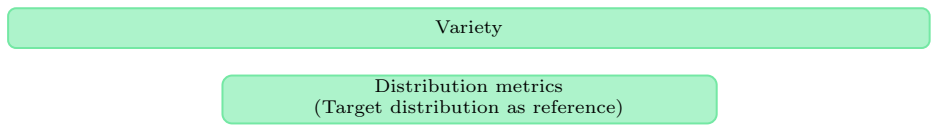
\begin{figure}[htb!]
  \centering
    \input{decision-trees/variety}
  \caption{Decision tree for the dimension ``Variety".}
  \label{fig:decisiontree-variety}
\end{figure}

\paragraph*{Variety}
\emph{Variety} characterizes the heterogeneity of features in the data. 
All distribution metrics can be used to evaluate variety. The target population should be employed as reference (Figure~\ref{fig:decisiontree-variety}).

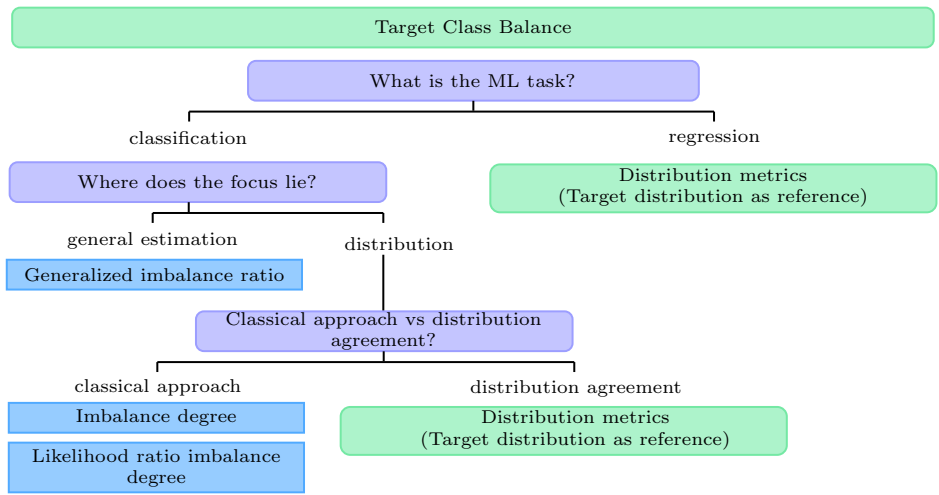
\begin{figure}[htb!]
  \centering
    \input{decision-trees/target-class-balance}
  \caption{Decision tree for the dimension ``Target Class Balance". }\label{fig:decisiontree-targetclassbalance}
\end{figure}

\paragraph*{Target class balance}
The dimension \emph{target class balance} should evaluate the distribution of outcome classes (Figure~\ref{fig:decisiontree-targetclassbalance}). The first question in the decision tree separates along the ML task (classification vs regression).
In the case of classification, classical metrics can be applied like the \emph{generalized imbalance ratio}~\cite{Buda2018}, a simple indicator, expressing the deviation from a uniform distribution as the ratio between the size of the majority and minority classes. 
More complex metrics, like \emph{imbalance degree}~\cite{Ortigosa2017} and \emph{likelihood ratio imbalance degree (LRID)}~\cite{Zhu2018} take the whole target class distributions in account and compare to uniform distribution. These can be utilized besides all other \emph{distribution metrics} (Figure~\ref{fig:decisiontree-distribution}). For regression tasks, the \emph{distribution metrics} are the only option. The expected target distribution should then be applied as reference.

\subsubsection*{Timeliness cluster}
The \textbf{timeliness cluster} in the METRIC-framework, which evaluates changes in time and whether they are appropriately reflected in the dataset, contains one quantitative dimension -- currency.

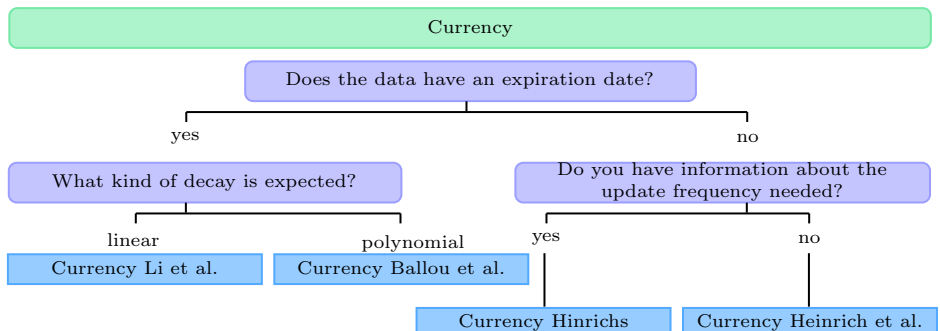
\begin{figure}[htb!]
  \centering
    \input{decision-trees/currency}
  \caption{Decision tree for the dimension ``Currency".}
  \label{fig:decisiontree-currency}
\end{figure}

\paragraph*{Currency}
We collected four metrics from literature (see Figure~\ref{fig:decisiontree-currency}), all called \emph{currency} by different authors~\cite{Ballou1998InformationQuality, Li2012, Hinrichs2002Datenqualitaet, Heinrich2007HowToMeasureDQ}, whose use depends on whether or not the data has an expiration date and on the shape of the expected decay.

\subsubsection*{Informativeness cluster}
The \textbf{informativeness cluster} is concerned with how well the data conveys the information it describes. It contains three quantitative dimensions -- uniqueness, informative missingness and feature importance.

\paragraph*{Uniqueness}
For the \emph{uniqueness} dimension, the metric \emph{prevalence of duplicates}~\cite{Qi2013FindDuplicates} provides a direct estimate of the proportion of duplicate data records in the dataset.
The metric \emph{effective sample size}~\cite{Thompson2012Sampling} on the other hand provides a more elaborate quantification of unique information but requires additional parameters that might not be available for most use cases (Figure~\ref{fig:decision-tree_uniqueness}). We recommend to use both for a comprehensive overview. 

\paragraph*{Informative missingness}
The dimension \emph{informative missingness} should quantify how much information missing values carry. In the decision tree, the metrics are separated along the question ``Which missingness mechanism shall be determined?" (Figure~\ref{fig:decision-tree_infomissingness}). The metric \emph{Little's test}~\cite{Little1988} is designed to determine whether missing values are ``missing completely at random'' (MCAR), i.e., independently from both observed and unobserved data, while the metric \emph{Joint likelihood model for informative dropout}~\cite{Diggle1994} quantifies the information in case missing values are not MCAR but rather ``missing at random" (MAR) or ``missing not at random" (MNAR). In the former case, the probability of missingness depends on the observed data but not on the missing data, whereas for MNAR it depends on the missing data themselves. This reflects the three aspects of \emph{informative missingness}~\cite{Rubin1976}. We recommend to use both in the general case.

\begin{figure}[htb!]
  \centering
    \input{decision-trees/feature-importance}
  \caption{Decision tree for the dimension ``Feature Importance". }
  \label{fig:decision-tree_featureimp}
\end{figure}
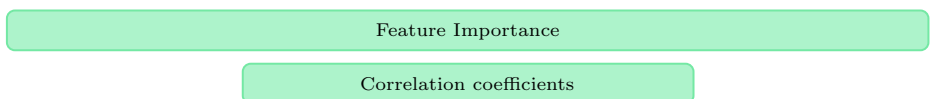

\paragraph*{Feature importance}
\emph{Feature importance} should quantify whether features of the data are relevant for the desired ML task. This is intended as an a priori evaluation in contrast to approaches from the field of explainability where the contribution of each feature to a specific prediction is quantified. For now, the \emph{correlation coefficients} are utilized to measure this dimension and show relations between features and target (Figure~\ref{fig:decision-tree_featureimp}).

\subsubsection*{Distribution metrics}
While the previously described metrics and decision trees are dimension specific, the following two groups of metrics and associated decision trees are utilized across multiple dimensions. 
Metrics within the group of \textbf{distribution metrics} (see decision tree in Supplement, Figure~\ref{fig:decisiontree-distribution}) are concerned with quantifying distributional aspects of the data.

We identified two main aspects of metrics within this group. The first one are metrics that evaluate characteristics of a single distribution, such as~\emph{interquartile range}~\cite{Freedman2007Statistics} for continuous variables and \emph{Hill numbers}~\cite{Ricotta2021} for categorical variables.
The second one are metrics that compare one distribution to another. This second group is divided again into three mathematical approaches: distance based metrics (quantifying how much two distributions differ), such as \emph{Wasserstein Distance}~\cite{Panaretos2019Wasserstein}; divergences (also quantifying differences of distribution but without satisfying mathematical properties of a metric), such as \emph{Kullback-Leibler Divergence (KLD)}~\cite{Kullback1951Information}; and statistical tests (hypothesis testing to estimate the probability whether apparent distributions or distinct samples originate from the same fundamental distribution), e.g., \emph{Kolmogorov-Smirnov test}~\cite{Freedman2007Statistics} and \emph{Chi-squared test}~\cite{Freedman2007Statistics}. 
For practical application, we recommend to use multiple metrics from different categories to provide a broad overview.

\subsubsection*{Correlation coefficients}
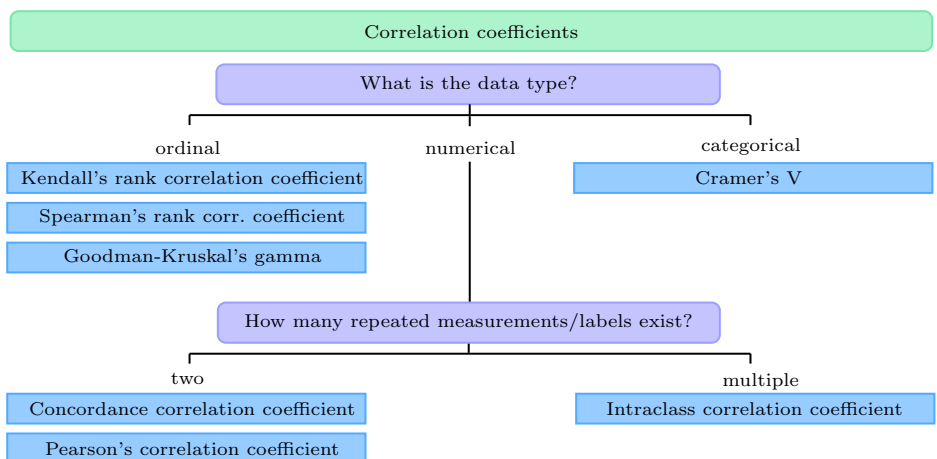
\begin{figure}[htb!]
  \centering
    \input{decision-trees/correlation-coefficients}
  \caption{Decision tree for ``Correlation Coefficients".  This tree is a subtree which can be used for various dimensions.}\label{fig:decisiontree-correlation}
\end{figure}

The group of \textbf{correlation coefficients} (Supplement, Figure~\ref{fig:decisiontree-correlation}) contains metrics that assess the correlation of two or more variables with each other.
The selection of metrics depends on the data type: \emph{Cramer's V}~\cite{Akoglu2018} for categorical variables, \emph{Spearman rank correlation}~\cite{Spearman1904Association}, amongst others, for ordinal variables and, e.g., \emph{Concordance correlation coefficient}~\cite{Altman1990PracticalStatistics} for numerical variables.

\FloatBarrier
\begin{figure}[htb!]
  \centering
    \input{decision-trees/noisy-labels}
  \caption{Decision tree for the dimension ``Noisy Labels".}
  \label{fig:decision-tree_noisylabels}
\end{figure}
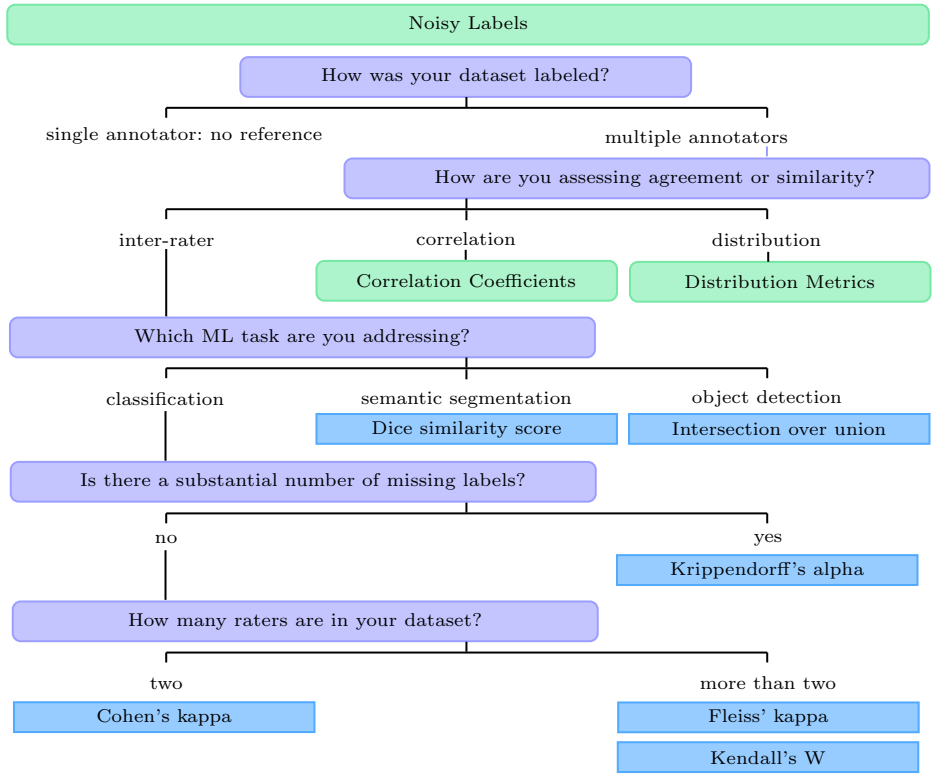

\begin{figure}[htb!]
  \centering
  \input{decision-trees/uniqueness}
  \caption{Decision tree for the dimension ``Uniqueness".}
  \label{fig:decision-tree_uniqueness}
\end{figure}
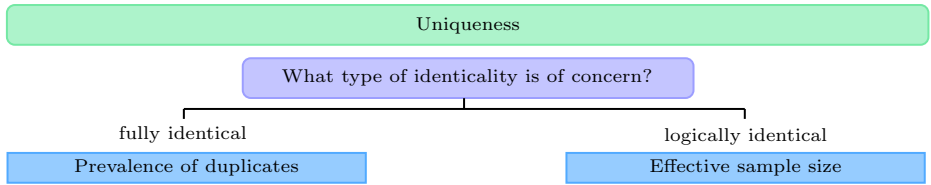

\begin{figure}[htb!]
  \centering
    \input{decision-trees/informative-missingness}
  \caption{Decision tree for the dimension ``Informative Missingness". }
  \label{fig:decision-tree_infomissingness}
\end{figure}
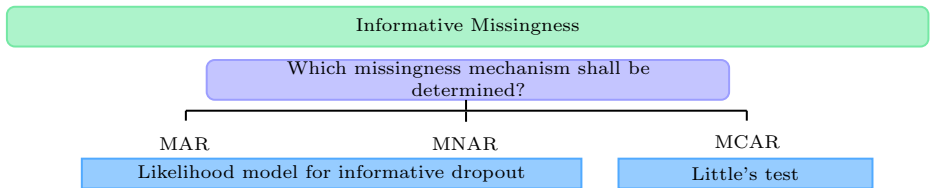

\FloatBarrier

\subsection*{Online platform: Metric Hub}
\begin{figure}[htb!]
    \centering
    \begin{subfigure}{0.48\textwidth}
        \centering
        \includegraphics[width=\linewidth]{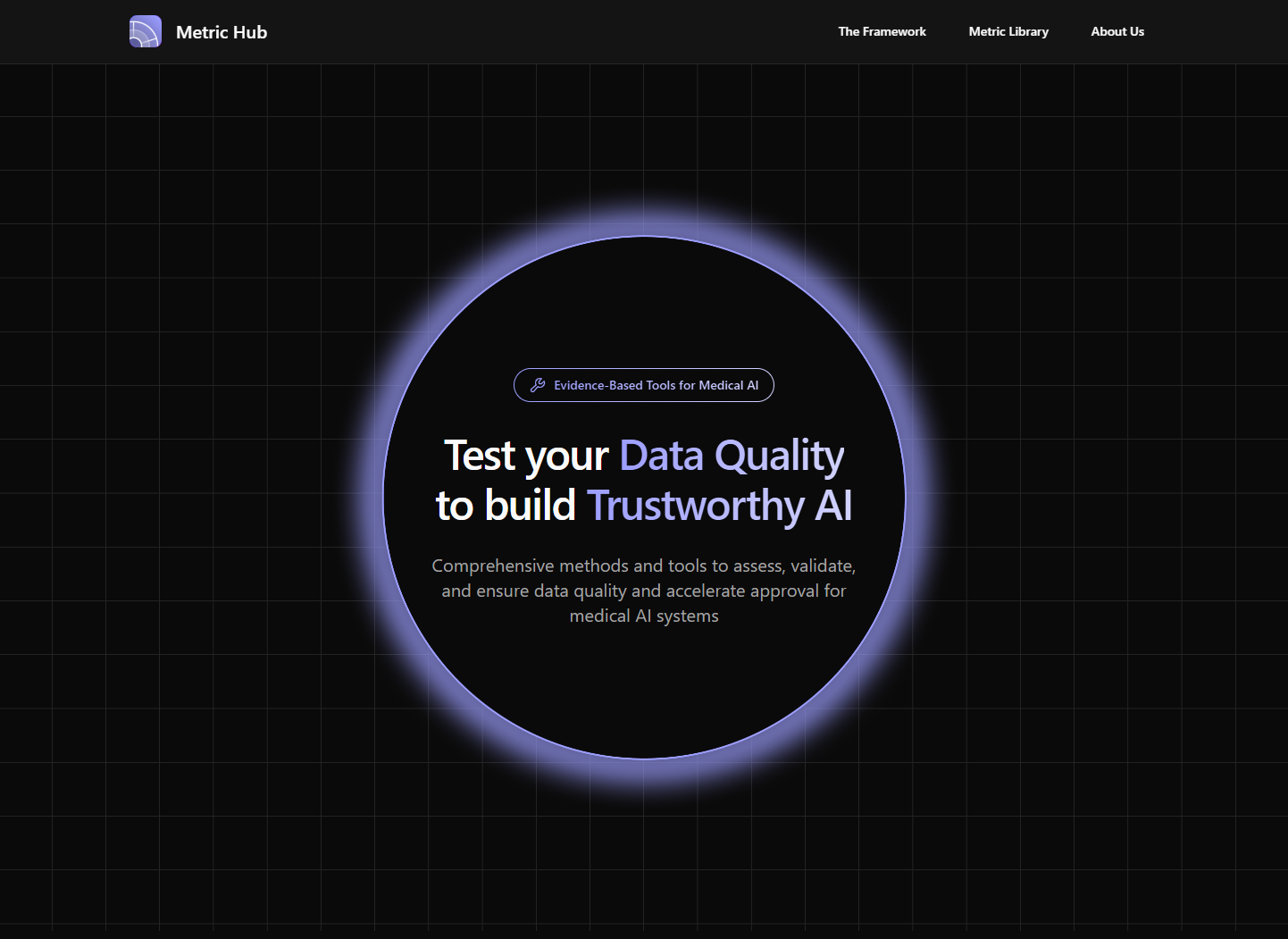}
        \caption{}
        \label{fig:metrichub:a}
    \end{subfigure}
    \hfill
    \begin{subfigure}{0.48\textwidth}
        \centering
        \includegraphics[width=\linewidth]{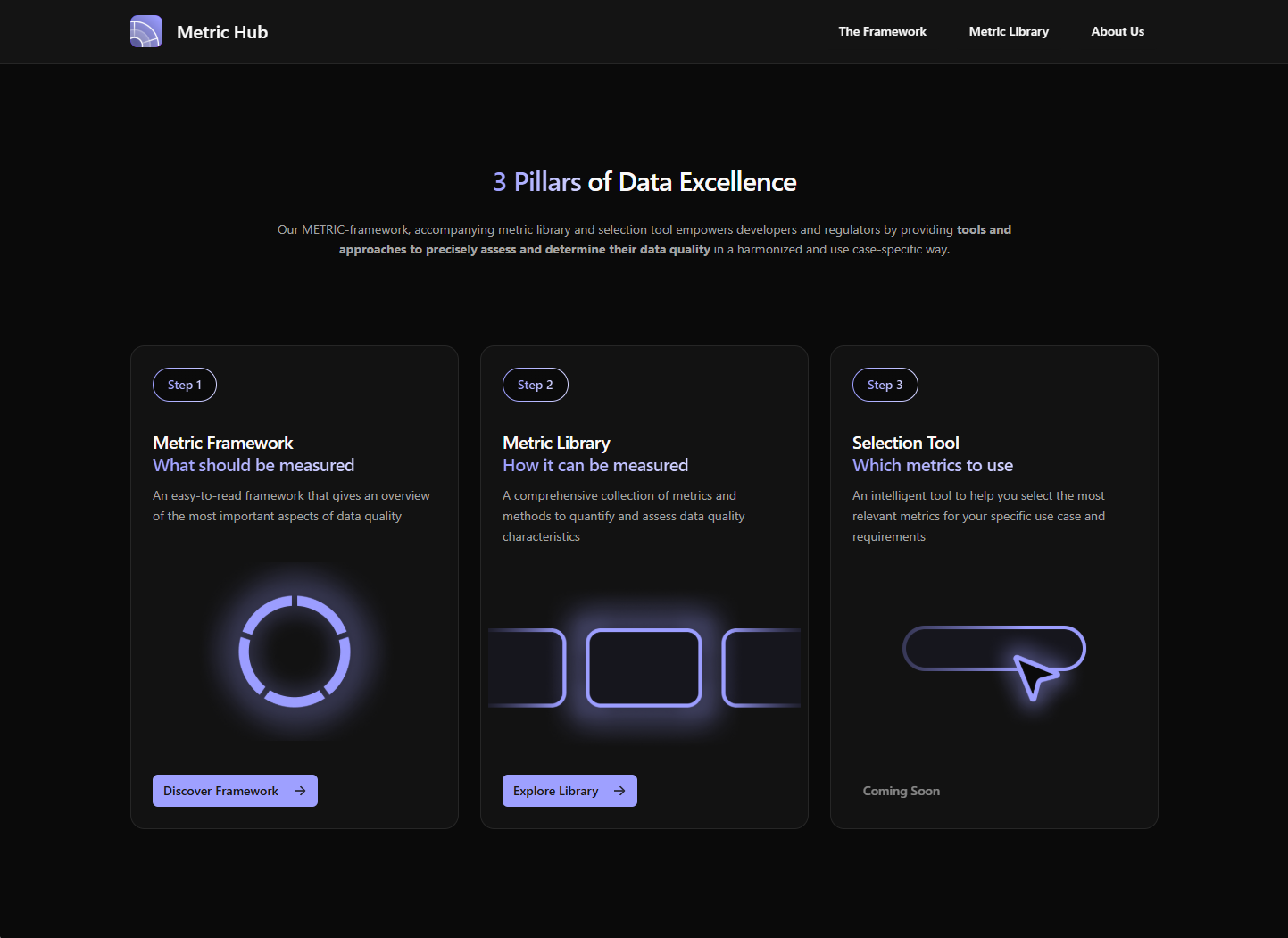}
        \caption{}
        \label{fig:metrichub:b}
    \end{subfigure}

    \vspace{0.5em}

    \begin{subfigure}{0.48\textwidth}
        \centering
        \includegraphics[width=\linewidth]{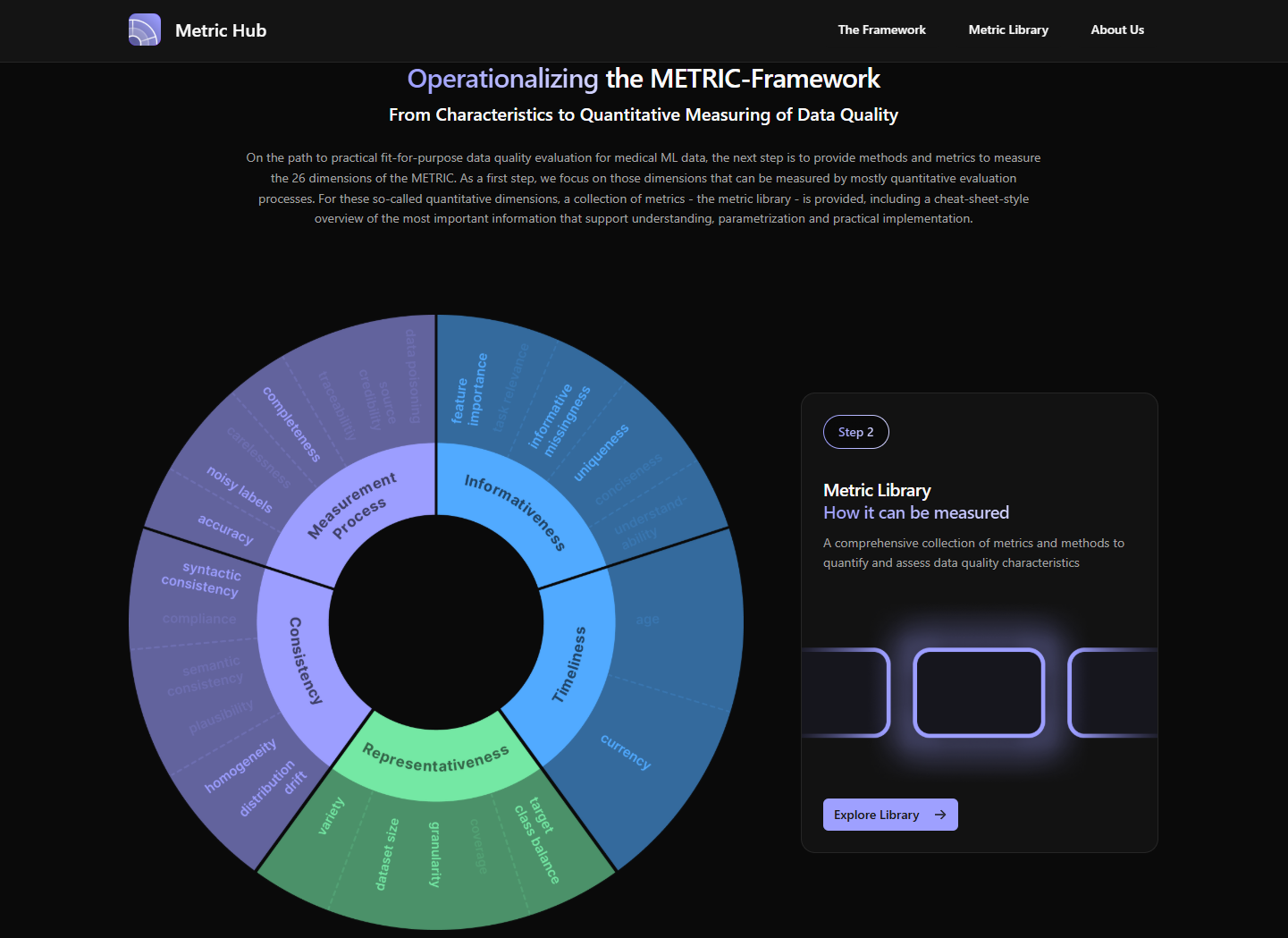}
        \caption{}
        \label{fig:metrichub:c}
    \end{subfigure}
    \hfill
    \begin{subfigure}{0.48\textwidth}
        \centering
        \includegraphics[width=\linewidth]{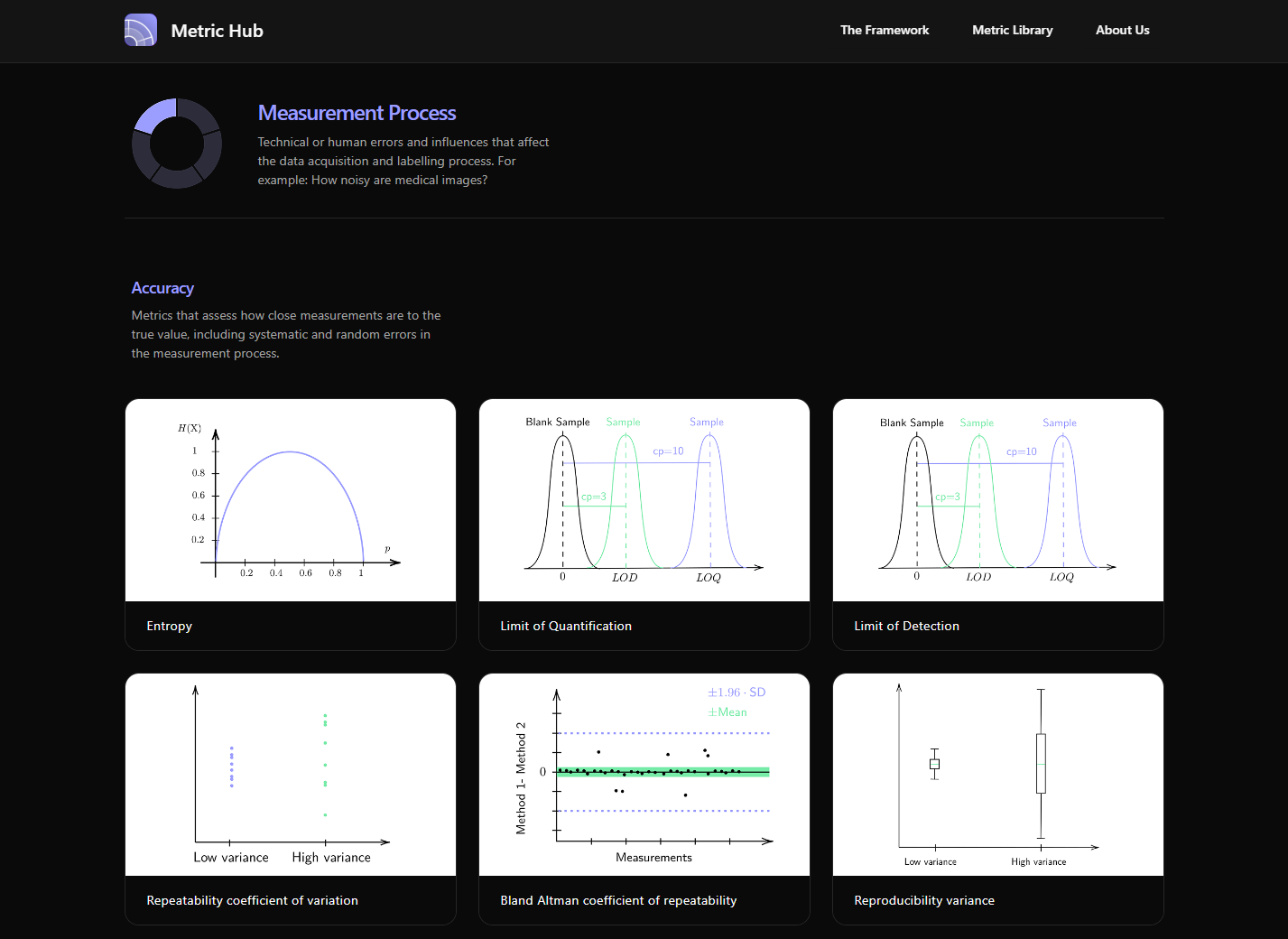}
        \caption{}
        \label{fig:metrichub:d}
    \end{subfigure}

    \caption{Online platform Metric Hub~\cite{Metrichub}.}
    \label{fig:metrichub}
\end{figure}

In order to facilitate access to the accumulated information and enable extension of the library in the future, we have created an online platform called Metric Hub (Figure~\ref{fig:metrichub})~\cite{Metrichub}. This platform is intended to be a first entry point for quantitative, systematic fit-for-purpose data quality evaluation for AI in medicine. It presents the work at hand (metric library) as well as the METRIC-framework~\cite{schwabeMETRICframeworkAssessingData2024}. The information from both publications is condensed into short text blocks supported by illustrations (Figure~\ref{fig:metrichub:c}). The metric library is organized by metric groups and associated data quality dimensions so that the metric cards can be easily viewed and downloaded (Figure~\ref{fig:metrichub:d}).

\section{Data availability}
All data generated or analysed during this study is available. 
The PTB-XL original dataset is included in~\cite{wagner2020ptbxl} and its supplementary information files.
The datasets analysed in the example and the code used to generate Figure~\ref{fig:metriccollection} are available at \url{https://github.com/katinkab/MetricLibrary}.

\section{Author contributions}
KB, DS designed and supervised the study.
KB, MO, TZ, MS, SC, VM, ICI, MK, GDA, ICA, JF, KD, IM, EDC, LO, AK, GA, AS, DS participated in the focus group for metric collection and metric card creation.
KB, MO, TZ, DS were members of the core group.
KB, MO, TZ, DS developed the decision trees.
KB, MO, TZ, SC, VM, JF, MS, DS wrote the manuscript.
All authors reviewed and approved the manuscript.

\section{Acknowledgments}
The authors acknowledge partial funding by the EU project TEF-Health. The project TEF-Health has received funding from the European Union’s Digital Europe programme under grant agreement no. 101100700. 
M.K. acknowledges funding by the Czech Ministry of Education (project LM2023050 Czech-BioImaging).
I.CI. acknowledges funding by the EU NextGenerationEU through the Recovery and Resilience Plan for Slovakia under the project No. 09I03-03-V04-00705. 
We thank JERY for the design of the associated website and support in redesigning Figure~\ref{fig:metricwheel}.
We thank Svenja Frister and Anne-Marie Strauch for early contributions during the metric collection process.

\section{Competing interests}
The authors declare no competing interests.

\bibliography{sn-bibliography}

\newpage


\section*{Supplementary material}

# 9 Supplementary information

## 9.1 Measurement Process

### 9.1.1 Accuracy

# Entropy
## Synonyms: Shannon entropy, Information entropy

A statistical measure of information content used to evaluate the uncertainty, diversity, or randomness of a variable or system.

$$H(X) := -\sum_{x \in X} p(x) \log p(x)$$

$H(X)$ is the entropy of the random variable $X$ and $p(x)$ the probability of a particular error value $x$.

**Value Range:** $[0, \log |X|]$ ↓
High entropy indicates unpredictable errors leading to greater uncertainty and variability in measurements, whereas low entropy indicates predictable errors with a more stable, consistent system leading to more reliable measurements.

### Use in METRIC-framework

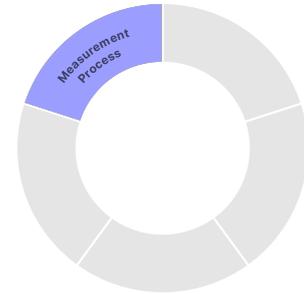

Accuracy

### References
(Shannon, 1948)

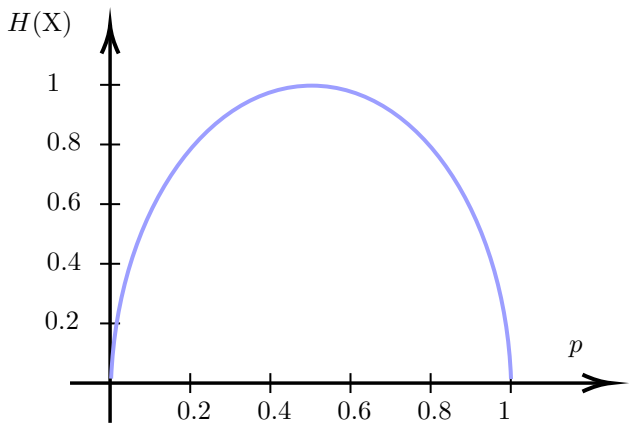

### Example
He et al., 2024 characterized the utility and validity of Shannon entropy removal to reanalyze the performance of clinical decision support tools compared to traditional validity tools including sensitivity, specificity, PPV/NPV, Youden's index, and diagnostic odds ratio.
Practical implementation: *entropy* function in scipy stats Python package

### Relation to other metrics
- Kullback-Leibler Divergence (KL)
- Cross Entropy
- Fisher Information
- Mutual Information

### Applicability
**Modality:**

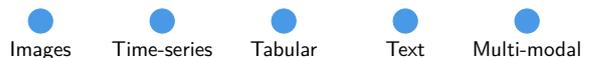

Images  Time-series  Tabular  Text  Multi-modal

**Variable:**

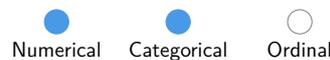

Numerical  Categorical  Ordinal

### Prerequisites and recommendations
- Requires a well-defined discrete distribution or a binned version af a continous variable.
- Requires a normalization of probabilities (sum to 1).
- Best used when comparing distributions over repeated measurements.
- Supporting categorical and numerical measurement errors.
- Combining with other metrics or for deeper insights adding Relative entropy.

### Pitfalls and limitations
- Dependence on parameter choice if continuous variables have to be binned.
- Sensitive to rare events with low probability.
- Interpretation only in context possible (e.g. higher entropy can be good in biological diversity).
- Entropy captures unpredictability, but not magnitude (e.g. same dataset-different value ranges).
- Use with caution when working with small sample sizes.

# Limit of Detection (LoD)
## Synonyms: Minimum detectable signal, LoD

The LoD represents the absolute lower limit at which a signal can be detected with a statistical significance. It is the point where the signal is distinguishable from the blank/background signal.

$$\text{LoD} = \text{LoB} + c_p \cdot \text{SD}_p$$

LoB is the mean background signal of the blank sample, $c_p$ is the coverage factor (typically 3.0 according to DIN 32645), and $\text{SD}_p$ is the standard deviation of the low concentration sample, pooled if more than one sample.

**Value Range:** $[0, +\infty)$ ↓

The range depends on the method type, the instrument technique characteristics and the possibility of blank samples.

A lower LoD means a method is more sensitive, able to detect minimal amounts.

### Use in METRIC-framework

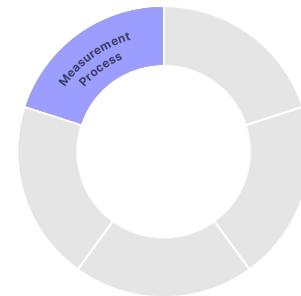

Accuracy

### References
(DIN 32645:2008-11, 2008; Armbruster, 1994)

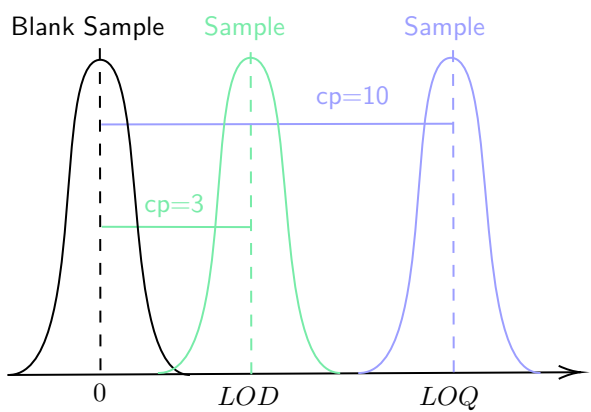

### Example

Validating a new immunoassay method for detecting the Prostate-Specific Antigen (PSA) in blood serum by a laboratory (DIN32645:2008-11).

- Blank samples (no PSA) are measured several times to calculate the mean of blank (LoB).
- Low concentration sample is measured several times to calculate the Standard deviation $\text{SD}_p$.
- DIN32645:2008-11 requires the corrected coverage factor $c_p = 3$.

**Result:** Any value greater or equal the result of LoD can be statistically distinguished from the background noise.

### Relation to other metrics
- Limit of Blank (LoB)
- Limit of Quantification (LoQ)
- Signal-to-noise ratio (SNR)

### Applicability
**Modality:**
● Images  ● Time-series  ● Tabular  ○ Text  ● Multi-modal

**Variable:**
● Numerical  ○ Categorical  ○ Ordinal

### Prerequisites and recommendations
- Blank samples are required to determine background signal (LoB).
- Requires repeatable measurement under defined conditions.
- Choosing consistent instrument settings and environment.
- Having standardized calibration of instruments.
- Preferring methods with low variation for a reliable LOD.
- Ideally reported with confidence interval (e.g., 95%)

### Pitfalls and limitations
- Mainly sensitive to noise level in blank sample.
- Varying across devices and laboratories.
- Over or underestimation if SD or LoB are poorly estimated.

# Limit of Quantification (LoQ)
Synonyms: Minimum quantifiable concentration, LoQ

LoQ is the lowest concentration or signal level at which a substance can be quantified with acceptable precision and trueness. It represents the point beyond the LoD at which the signal is not only detectable but also reliably measurable.

$$\text{LoQ} = \text{LoB} + c_p \cdot \text{SD}_p$$

LoB is the mean background signal of a blank sample, $c_p$ the corrected coverage factor (typically 10.0 according to DIN 32645) and $\text{SD}_p$ the standard deviation, pooled if more than one sample is used.

**Value Range:** $[0, +\infty) \downarrow$
The range depends on the method type, the instrument technique characteristics and the possibility of blank samples.
Lower values of LOQ indicate higher sensitivity.

## Use in METRIC-framework

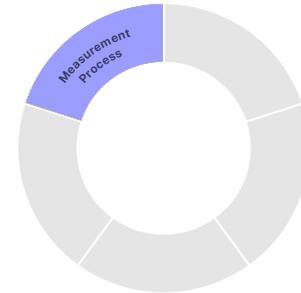

Accuracy

## References
(DIN32645:2008-11, 2008; Armbruster, 1994)

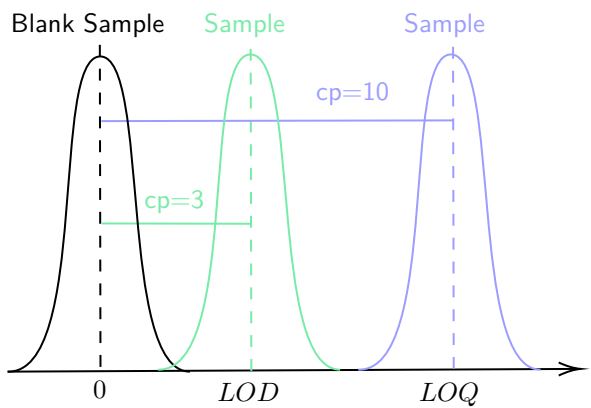

## Example
Validating a new immunoassay method for detecting the Prostate-Specific Antigen (PSA) in blood serum by a laboratory (DIN32645:2008-11, 2008).

## Relation to other metrics
- Limit of Blank (LoB)
- Limit of Detection (LOD)
- Signal-to-noise ratio (SNR)

## Applicability
**Modality:**

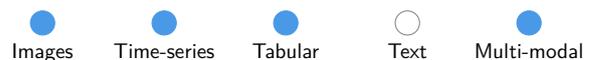

Images  Time-series  Tabular  Text  Multi-modal

**Variable:**

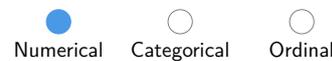

Numerical  Categorical  Ordinal

## Prerequisites and recommendations
- Blank samples are required to determine the background signal (LoB).
- Requires repeatable measurement under defined conditions.
- Choosing consistent instrument settings and environment.
- Having standardized calibration of instruments.
- Preferable when LoD alone is not sufficient for quantification.
- Ideally reported with confidence interval (e.g., 95%)

## Pitfalls and limitations
- Mainly sensitive to noise level.
- Varying across devices and laboratories.
- Over- or underestimation if SD or LoB are poorly estimated.
- Not for qualitative tests.

# Systematic error in instruments

A systematic error refers to the consistent deviation of measured values from the true value in a particular direction; it reflects a lack of accuracy rather than random variation.

$$\epsilon = v_m - v_t$$

where $v_m$ is the measured value, $v_t$ is the true value.

**Value Range:** $(-\infty, \infty)$
The range depends on the measuring instrument, method and calibration. Systematic errors can be positive (overestimation) or negative (underestimation). A high absolute value indicates a significant measurement bias, whereas a value close to 0 indicates more precise measurements closer to the true value.

## Use in METRIC-framework

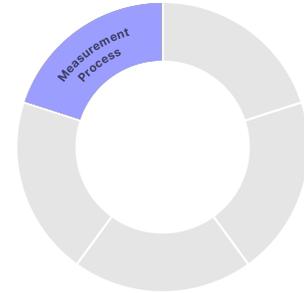

Accuracy

## References

(Christodoulides and Christodoulides, 2017)

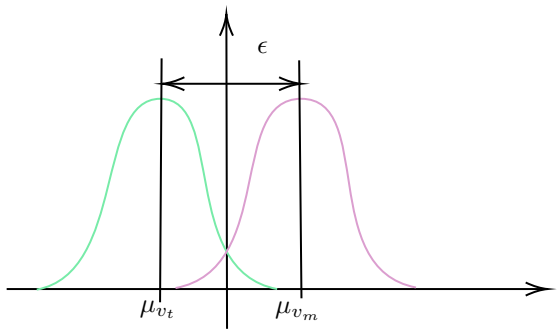

## Example

Systematic overestimation of blood glucose levels in a faulty glucometer (ISO 15197:2013/EN ISO 15197:2015).

## Relation to other metrics

- Random error in instruments
- Mean Absolute Error (MAE)

## Applicability

**Modality:**
● Images  ● Time-series  ● Tabular  ○ Text  ○ Multi-modal

**Variable:**
● Numerical  ○ Categorical  ○ Ordinal

## Prerequisites and recommendations

- A suitable instrument for the specific method should be selected.
- Calibration of devices is required for accurate measurement.
- Requires a known reference or gold standard for the true value.
- Regular validation with standard datasets.
- Always report with confidence intervals.

## Pitfalls and limitations

- Sensitive to incorrect calibration.
- Affected by environmental conditions, e.g. temperature, humidity.
- In case both systematic and random errors are mixed.
- If the true value is unknown or highly uncertain.
- Applying this metric to non-quantitative or categorical data where deviations are not numerical.

# Random error in instruments
## Synonyms: Instrument noise, Measurement noise

Random error refers to unpredictable fluctuations in measured values caused by inherent noise in the measurement process. Unlike systematic error, random error does not introduce a consistent bias but affects the precision of repeated measurements.

$$v_m = v_t + \epsilon_r, \qquad \mathbb{E}[\epsilon_r] = 0$$

where $v_m$ is the measured value, $v_t$ is the true value, and $\epsilon_r$ represents a random error term with zero mean. The magnitude of random error is commonly quantified using the variance or standard deviation:

$$\sigma^2 = \mathbb{E}[(v_m - \mathbb{E}[v_m])^2]$$

**Value Range:** $[0, \infty)$ ↓
Higher values indicate increased measurement variability and reduced precision, while values close to 0 indicate high repeatability.

### Use in METRIC-framework

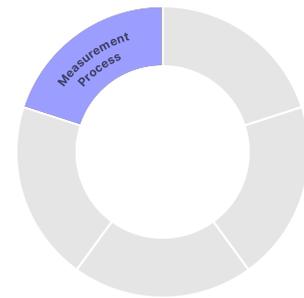

Accuracy

### References
(Christodoulides and Christodoulides, 2017)

### Relation to other metrics
- Systematic error in instruments
- Standard deviation
- Variance
- Signal-to-noise ratio (SNR)

### Applicability
**Modality:**

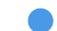 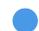 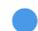 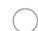 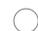
Images  Time-series  Tabular  Text  Multi-modal

**Variable:**

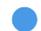 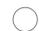 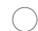
Numerical  Categorical  Ordinal

### Prerequisites and recommendations
- Repeated measurements under identical conditions are required.
- Measurement noise should be stationary over time.
- Use sufficiently large sample sizes to estimate variability reliably.
- Report dispersion measures such as standard deviation or variance.

### Pitfalls and limitations
- Cannot detect systematic bias in measurements.
- Sensitive to outliers, which may inflate variance estimates.
- May be confounded with systematic error if experimental conditions change.
- Interpretation depends on the assumed noise distribution.

# Bland-Altman Coefficient of Repeatability (CoR)

CoR quantifies the expected maximum difference between two repeated measurements or methods for the same subject under identical conditions. It provides a direct measure of agreement and repeatability.

$$CoR = 1.96 \cdot \sigma_{\text{difference}}$$

Where $\sigma_{\text{difference}}$ is the standard deviation of the differences between the two measurements and 1.96 is the 0,975-quantile of the standard normal distribution.

**Value range:** $[0, \infty)$ ↓
A smaller CoR indicates that the measurements are more consistent and agree closely within a pre-defined range, whereas a larger CoR indicates more variability.

## Use in METRIC-framework

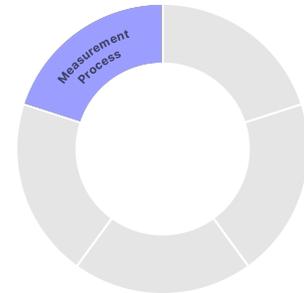

Accuracy

## References

(Barnhart and Barboriak, 2009; Bland JM, 1986)

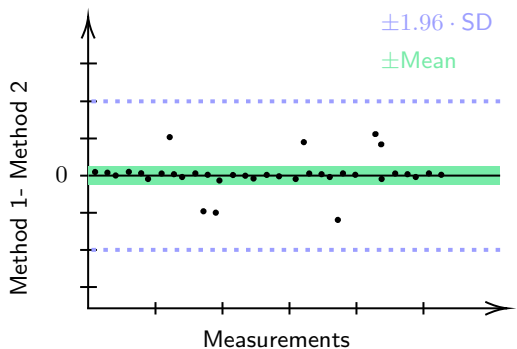

## Example

The Bland-Altman coefficient of repeatability is utilized in studies (see Barnhart& Barboriak, 2009) involving repeat imaging data sets of cancer patients to evaluate the variability in quantitative imaging parameters. For instance, it helps determine if a change in imaging measurements reflects a true change in a patient's condition or merely variability, thus aiding in the development and validation of imaging biomarkers. Additionally, it assesses whether different imaging devices can be interchangeably used for consistent measurements, ensuring reliable repeatability in studies aimed at improving cancer imaging techniques. Practical implementation: pyCompare Python package

## Applicability

**Modality:**

● Images   ● Time-series   ● Tabular   ○ Text   ○ Multi-modal

**Variable:**

● Numerical   ○ Categorical   ○ Ordinal

## Prerequisites and recommendations

- The data must be paired, such that each measurement has a corresponding repeat measurement.

## Pitfalls and limitations

- Unstable for small sample size
- CoR only considers repeatability, not systematic bias (e.g., if an instrument consistently overestimates or underestimates values).

# Repeatability Coefficient of Variation (CV)

CV is a normalized measure of measurement precision, used to assess how consistently a method or instrument performs across repeated trials.

$$CV = \frac{\sigma}{\overline{y_i}} \cdot 100$$

$\sigma$ is the standard deviation, $\overline{y_i}$ is the mean of the measurement results.

**Value range:** $[0, \infty)$ ↓
Values close to 0 indicate low variance. For values above 100 the standard deviation is bigger than the mean.

## Use in METRIC-framework

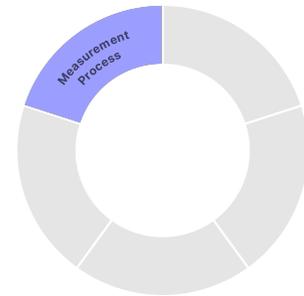

Accuracy

## References

( Lee, 2021, ISO 5725-4-2020-03)

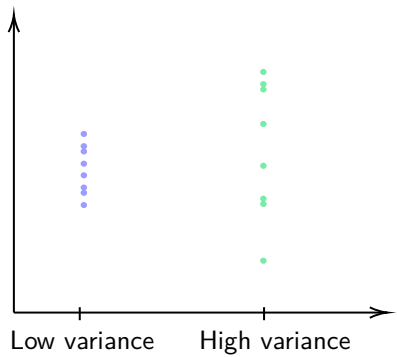

## Example

Lee, 2021 used CV to assess the short- and long-term repeatability of peripapillary optical coherence tomography angiography measurements in healthy eyes, Practical implementation: *variation* function in scipy stats Python package

## Relation to other metrics

- standard deviation
- mean

## Applicability

**Modality:**
○ Images  ● Time-series  ● Tabular  ○ Text  ○ Multi-modal

**Variable:**
● Numerical  ○ Categorical  ○ Ordinal

## Prerequisites and recommendations

- Mean should not be zero or close to zero.

## Pitfalls and limitations

- It should be used with caution, as comparing different units or scales can cause problems. Variance or standard deviation is preferred.
- The CV should be used for data scales with a meaningful zero, in order to ensure comparability between different scales.

# Reproducibility Variance

Reproducibility Variance captures the variability in measurements caused by changes in conditions such as operators, instruments, or locations, reflecting how consistently a method performs across different settings.

$$S_R^2 = \frac{1}{p-1}\sum_{i=1}^{p}(\overline{y_i} - \overline{y})^2 + (1 - \frac{1}{n})S_r^2$$

$p$ is the number of laboratories, $\overline{y_i} = \sum_{k=1}^{n} y_{ik}$ is the average of n measurement results $y_{ik}$ in laboratory i, $\overline{y} = \frac{1}{p}\sum_{i=1}^{p}\overline{y_i}$ and $S_r^2$ is the repeatability variance.

**Value range:** $[0, \infty)$ ↓
Reproducibility Variance should be low, meaning different operators or conditions do not introduce significant variation.

## Use in METRIC-framework

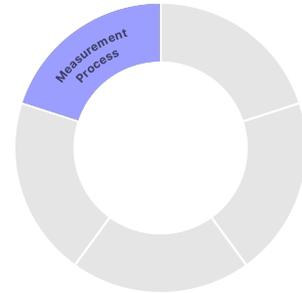

Accuracy

## References

( Lee, 2021, ISO 5725-4-2020-03)

## Relation to other metrics

- Alternatively Reproducibility Standard Deviation can be used.

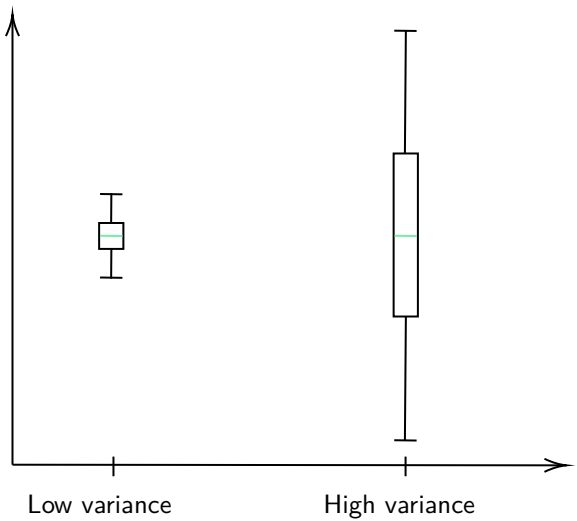

Low variance · High variance

## Applicability

**Modality:**
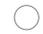 Images  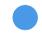 Time-series  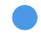 Tabular  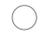 Text  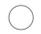 Multi-modal

**Variable:**
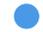 Numerical  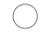 Categorical  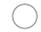 Ordinal

## Prerequisites and recommendations

- The measurements must be taken under different operators, instruments or conditions.
- A consistent measurement procedure must exist to minimize human-induces variability.
- Ensure that the used instruments are well calibrated.

## Pitfalls and limitations

- To ensure reproducibility the operator must be well trained to avoid errors in the measurement technique and to avoid personal biases.
- Make sure to use comparable measurement instruments.

### 9.1.2 Noisy labels

# Cohen's Kappa $\kappa$

Cohen's kappa $\kappa$ measures the agreement between two raters or classifiers assigning categorical labels, while adjusting for the agreement expected by chance. It is specifically applicable to categorical (nominal or ordinal) variables.

$$\kappa = \frac{p_o - p_e}{1 - p_e}$$

with $p_o$ being the mean observed agreement between all raters and $p_e$ the probability of expected agreement.

**Value Range:** $[-1, 1] \uparrow$

A value of $\kappa = 1$ indicates a perfect agreement, $\kappa = 0$ corresponds to chance-level agreement and a value of $\kappa < 0$ can be interpreted as an agreement worse than chance.

Interpretation depends on the context, this reference could be used Landis & Koch, 1977.

## Use in METRIC-framework

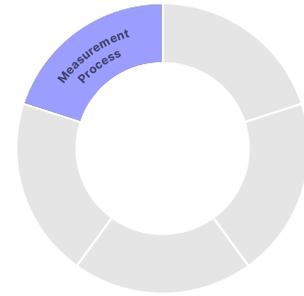

Noisy labels

## References

(Gisev et al., 2013; McHugh, 2012)

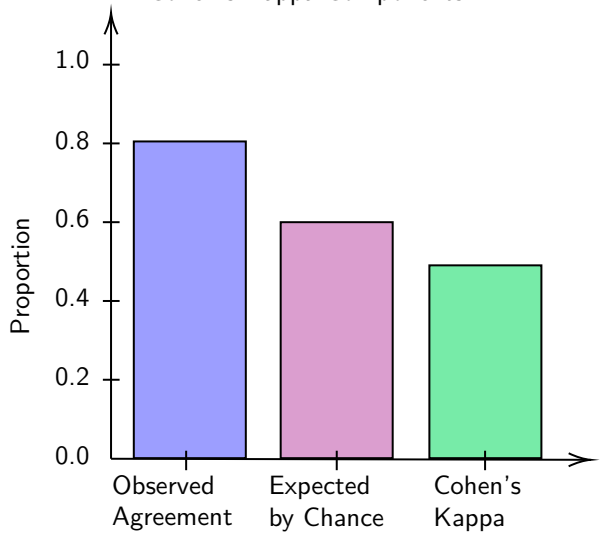

## Example

In Jue, 2021, Cohen's kappa index was used to evaluate the diagnostic accuracy and agreement between the AI-assisted microscope's HER2 scoring and the gold standard, as well as to compare pathologists' adjusted scores after reviewing AI outputs. Practical implementation: *cohen_kappa_score* function in scikit-learn Python package.

## Relation to other metrics

- Fleiss´ Kappa
- Kendall´s W
- Krippendorff´s Alpha

## Applicability

**Modality:**

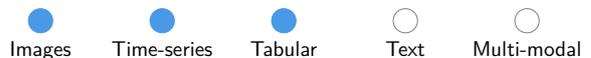

Images · Time-series · Tabular · Text · Multi-modal

**Variable:**

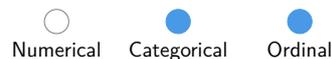

Numerical · Categorical · Ordinal

## Prerequisites and recommendations

- Categorical labelling by two raters or classification systems.
- Independent observations are required.
- Each rater assigns one category per item (mutually exclusive).
- Ratings should be comparable across the same scale or system.
- Having a sufficient sample size.
- Provide confidence intervals and percent agreement alongside $\kappa$.

## Pitfalls and limitations

- Sensitive to marginal distributions.
- Imbalanced data can have an influence on $\kappa$, even with high observed agreement.
- Only applicable for two raters or classifiers.
- Interpretation only possible in context.
- Unstable for imbalanced data - In case of imbalance, use additional metrics to $\kappa$.

# Fleiss' Kappa $\kappa$

Fleiss' Kappa $\kappa$ is a statistical measure of inter-rater agreement for categorical ratings when there are more than two raters or repeated ratings by fewer annotators. It extends Cohen's Kappa by accounting for chance agreement across multiple raters and is applicable only to categorical data.

$$\kappa = \frac{p_o - p_e}{1 - p_e}$$

with $p_o$ being the mean observed agreement between all raters and $p_e$ the probability of expected agreement by chance.

**Value Range:** $[-1, 1] \uparrow$
A value of $\kappa = +1$ indicates a perfect agreement, $\kappa = 0$ corresponds to chance-level agreement and a value of $\kappa < 0$ can be interpreted as an agreement worse than chance.

## Use in METRIC-framework

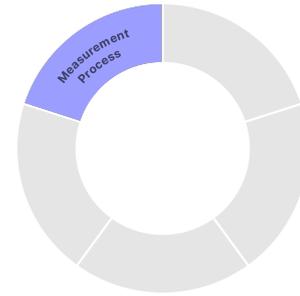

Noisy labels

## References

(Gisev et al., 2013; McHugh, 2012)

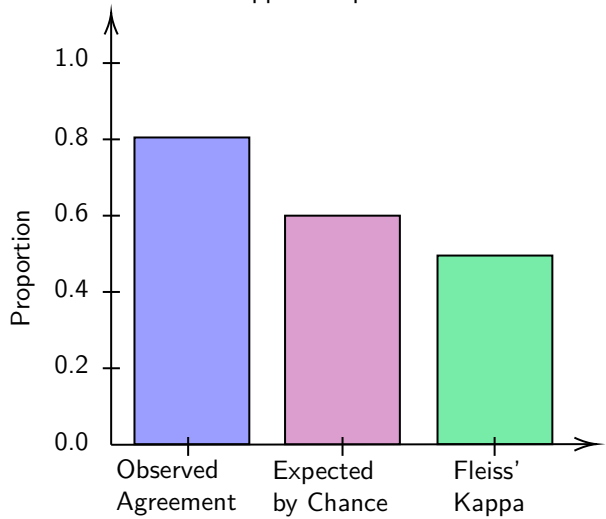

Fleiss' Kappa Components

## Example

Evaluating the agreement between more than two pathologists independently classifying tissue slides into cancer grades. Fleiss' kappa $\kappa$ evaluates the overall inter-rater consistency across all raters and cases. Each slide was rated by all experts without prior discussion, simulating real-world annotation variability often encountered in datasets used for machine learning. For the interpretation the scale by Landis and Koch (1977) can be used. Used in: Arimone Y., et al 2007 Practical implementation: *fleiss_kappa* function in statsmodels Python package

## Relation to other metrics

- generalisation of Cohen´s Kappa (from 2 to n)
- Krippendorff´s Alpha
- Kendall´s Coefficient of Concordance

## Applicability

**Modality:**

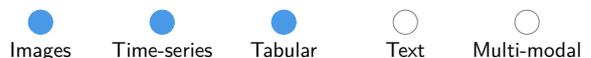

Images | Time-series | Tabular | Text | Multi-modal

**Variable:**

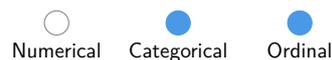

Numerical | Categorical | Ordinal

## Prerequisites and recommendations

- Only applicable for categorical ratings.
- Requires a fixed number and at least two raters per item.
- An independent rating is required.
- Standardized criteria for rating and raters to reduce subjective variability.
- Best used with balanced dataset.
- Stratifying results by subgroups (e.g. experience level of raters).

## Pitfalls and limitations

- Sensitive to marginal distributions.
- Unstable for imbalanced data, which can have an influence on $\kappa$, even with a high observed agreement.
- Interpretation only possible in context.
- Interpret using confidence intervals and category distributions

# Kendall's W
## Synonyms: Kendall´s coefficient of concordance

Kendall's W measures the degree of agreement among multiple raters ranking a set of items, indicating the strength and consistency of inter-rater rankings. It is especially useful for assessing ordinal consensus across several evaluators.

$$W = \frac{12 \cdot \sum_{i=1}^{n}(t_i - \bar{t})^2}{m^2(n^3 - n)}$$

where $t_i$ is the sum of ratings for object $i$, $\bar{t}$ the mean value of sum of ratings for all objects, $m$ the total number of raters and $n$ the total number of objects.

**Value Range:** $[0, 1]$ ↑
A value of 0 signifies no agreement and a value of $+1$ signifies perfect agreement.

### Use in METRIC-framework

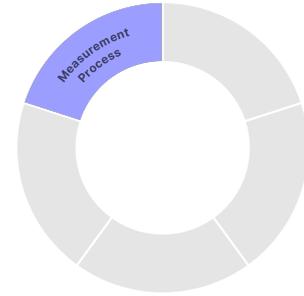

Noisy labels

### References
( Gisev et al., 2013)

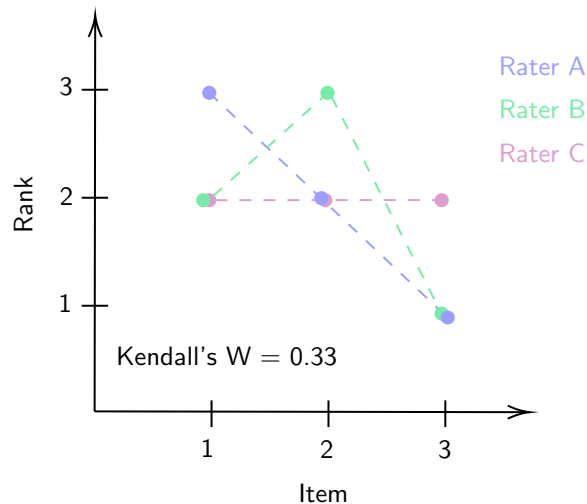

Kendall's W = 0.33

### Relation to other metrics
- Fleiss´ Kappa
- Krippendorff´s Alpha
- Cohen´s Kappa

### Applicability
**Modality:**

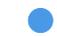 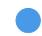 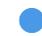 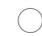 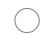
Images    Time-series    Tabular    Text    Multi-modal

**Variable:**

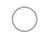 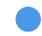 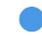
Numerical    Categorical    Ordinal

### Prerequisites and recommendations
- Data must be ranked or rankable.
- Same objects must be rated by all raters.
- Scale consistency across raters is required.
- Used for $m \geq 2$ raters and $n \geq 2$ objects.
- Independent ratings.
- Complete data sets.
- Report along with p-value to test for non-random concordance

### Pitfalls and limitations
- Sensitive to missing values.
- Ties can affect the results.
- No differentiation between types of disagreement.

# Krippendorff's Alpha
Synonyms: Krippendorff's coefficient

Krippendorff's Alpha is a flexible reliability metric that measures inter-rater agreement for any number of raters, while accounting for chance agreement and missing data.

$$\alpha = 1 - \frac{D_0}{D_e} = 1 - \frac{\sum_{x=1, x'=1}^{x} o_{xx'} \delta(x, x')}{\frac{1}{n-1} \sum_{x=1, x'=1}^{x} n_x n_{x'} \delta(x, x')}$$

$D_0$ is the disagreement observed and $D_e$ is the disagreement expected by chance.

**Value range:** $[-1, 1]$ ↑
Value $+1$ indicates perfect reliability, 0 indicates no reliability, alpha less than 0 implies systematic disagreements among raters.

### Use in METRIC-framework

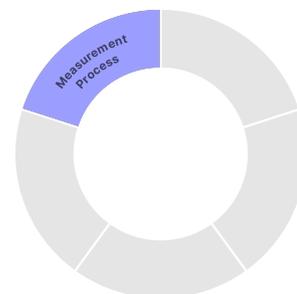

Noisy labels

### References
(Krippendorf, 2013; Nasser et al., 2019)

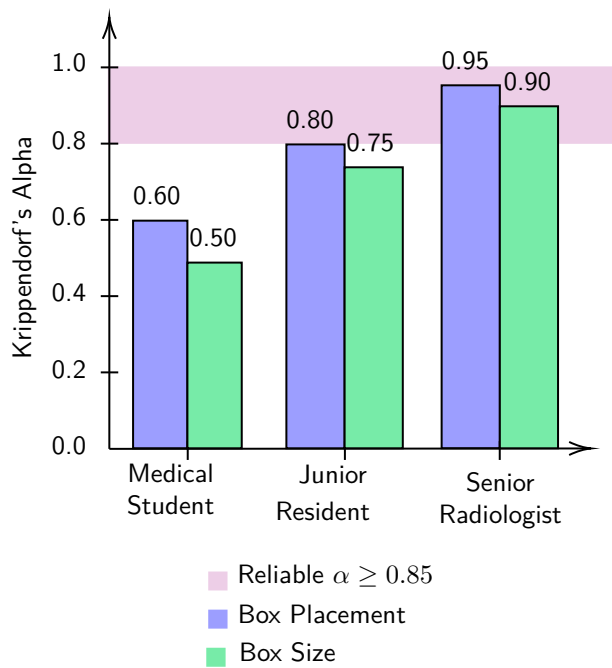

### Example
In a functional MRI data quality control study (Williams, 2023), Krippendorff's alpha was used to quantify the inter-rater agreement among multiple reviewers assessing the quality of functional MRI scans, providing a robust measure of consistency even in the presence of missing or variable-quality assessments.

Practical implementation: *krippendorff.alpha* function in krippendorff Python package.

### Applicability
**Modality:**
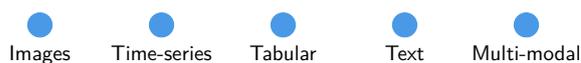
Images  Time-series  Tabular  Text  Multi-modal

**Variable:**
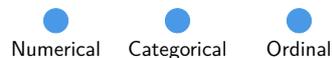
Numerical  Categorical  Ordinal

### Prerequisites and recommendations
- More annotators lead to more reliable results.
- Suitable for small and large sample sizes as it is adjusting itself.
- Can handle different scales and incomplete or missing data.

### Pitfalls and limitations
- The minimum acceptable $\alpha$ has to be chosen in its context.

# Dice Similarity Score (DSS)
## Synonyms: Dice-Sørensen coefficient, Dice's coefficient

DSS measures the overlap between two regions of interest by calculating the ratio of twice the intersection to the areas.

$$DSS = \frac{2|A \cap B|}{|A| + |B|} \qquad \text{for sets } A \text{ and } B.$$

**Value Range:** $[0, 1] \uparrow$
High DSS values close to 1 indicate strong overlap/agreement between annotators, whereas a value close to 0 indicates little overlap/agreement.

### Use in METRIC-framework

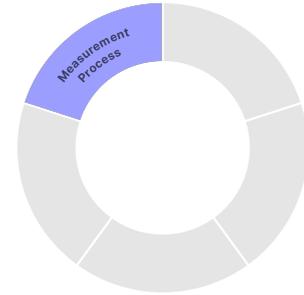

Noisy labels

### References
(Bilic et al., 2023; Reinke et al., 2024; Dice et al., 1945)

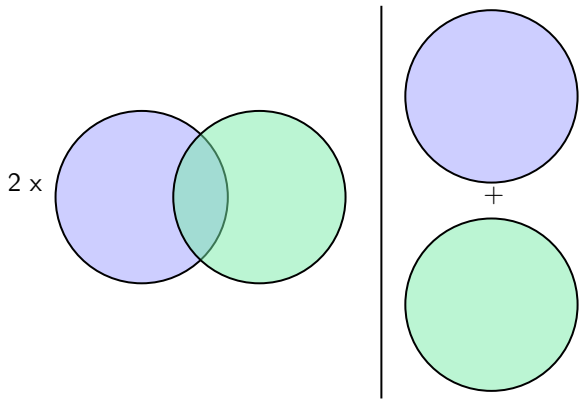

### Example
Comparing manual ground truth segmentations with automated AI-generated masks in the field of medical image processing.
DSS in Ma, 2024 was used as the primary metric to quantitatively evaluate and compare the segmentation of the foundation model against baseline and specialist models across 86 internal and 60 external medical image segmentation tasks. Practical implementation: *DiceMetric* class in monai Python package

### Relation to other metrics
- Intersection over Unit (IoU):
  $DSS = \frac{2IoU}{1+IoU}$
- DSS is equal to $F_1$ score at pixel level

### Applicability
**Modality:**
● Images  ○ Time-series  ○ Tabular  ○ Text  ○ Multi-modal

**Variable:**
● Numerical  ○ Categorical  ○ Ordinal

### Prerequisites and recommendations
- Data must be binary (or binarized) for proper comparison.
- Each pixel (or voxel) in the segmentation mask must correspond exactly to the same location in the reference (ground truth) mask, with no misalignment.
- Requires pre-processing (e.g. Resizing, Resampling, Padding).
- Best used for tasks involving overlapping regions (e.g. image segmentation).
- Avoid using for highly imbalanced classes.
- Suitable for noisy segmentation evaluation.

### Pitfalls and limitations
- Not reliable in very small or very sparse regions.
- Not differentiating between types of errors, i.e. does not distinguish false positives vs. false negatives; only the overlap matters.

# Intersection over union (IoU)
## Synonyms: Jaccard index, Tanimoto coefficient

The IoU measures the overlap between two sets or regions of interest by calculating the ratio of their intersection to their union. It is widely used in image analysis to assess inter-rater agreement.

$$IoU = \frac{|A \cap B|}{|A \cup B|} \quad \text{for sets } A \text{ and } B.$$

**Value range:** $[0,1]$ ↑
High values close to $+1$ indicate a strong overlap/agreement between regions, while values close to 0 mean little overlap/agreement.

### Use in METRIC-framework

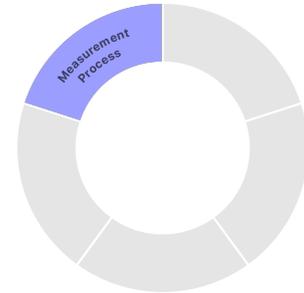

Noisy labels

### References
(Jaccard, 1912; Reinke et al., 2024)

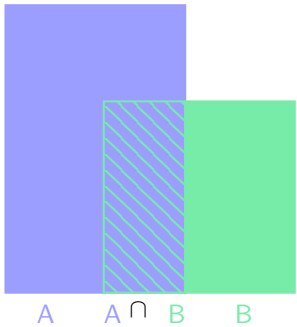

### Example
Practical implementation: *MeanIoU* class in monai Python package.

### Relation to other metrics

Dice Similarity Score (DSS):
$$IoU = \frac{DSS}{2-DSS}$$
$F_1$ Score:
$$IoU = \frac{F_1}{2-F_1}$$

### Applicability

**Modality:**

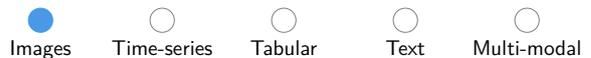

Images · Time-series · Tabular · Text · Multi-modal

**Variable:**

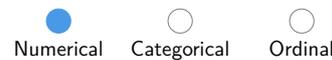

Numerical · Categorical · Ordinal

### Prerequisites and recommendations
- Most popular measure in object detection.
- $IoU$ should generally be used in combination with a boundary-based metric if boundary regions are of interest.
- The threshold for $IoU$ is largely dependent on the specific object detection task and dataset. However, a common threshold used in practice is 0.5.

### Pitfalls and limitations
- $IoU$ is unaware of shapes, boundaries, centers and distances.
- In small structures, small differences (even one pixel) have a high impact.
- $IoU$ only measures the region overlap, not in which case a complex structure shape is recognized better.

### 9.1.3 Completeness

# Completeness

This measure provides the proportion of values that are not missing in the dataset. A dataset is complete, if no values are missing.

$$C = 1 - \frac{M}{T}$$

where $M$ is the number of missing values and $T$ is the number of total values in the dataset.

**Value range:** $[0, 1]$ ↑
A value of 1 refers to a complete dataset, a value of 0 translates to a dataset where all values are missing.

## Use in METRIC-framework

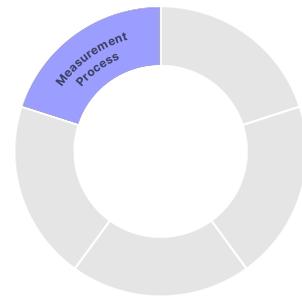

Completeness

## References
(Blake and Mangiameli, 2011; Emmanuel et al., 2021)

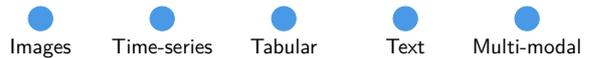

$$C = 1 - \frac{\ \rule{1cm}{0.3cm}\ }{\ \rule{1cm}{0.3cm} + \rule{1cm}{0.3cm}\ }$$

## Applicability

**Modality:**

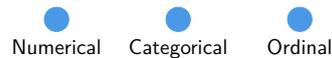

Images   Time-series   Tabular   Text   Multi-modal

**Variable:**

Numerical   Categorical   Ordinal

## Prerequisites and recommendations
- Knowledge about missing values.

## Pitfalls and limitations
- Does not indicate how the missing values are distributed over the data set.
- Does not indicate how missing values might depend on each other or observed values (informative missingness).

# Patient-level completeness

Proportion of patients with at least one recorded value for a specific attribute out of the total number of patients.

$$R = \frac{C}{P}$$

where $C$ is the number of patients having at least one record of the considered variable and $P$ is the total number of patients.

**Value range:** $[0, 1]$ ↑
Value 0 indicates that no patient has a record of the considered variable. Value 1 indicates that all patients have at least one record for the considered variable.

## Use in METRIC-framework

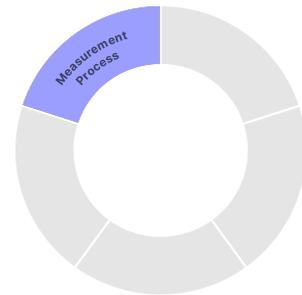

Completeness

## References
(Liu et al., 2017)

## Applicability

**Modality:**
● Images  ● Time-series  ● Tabular  ● Text  ● Multi-modal

**Variable:**
● Numerical  ● Categorical  ● Ordinal

## Prerequisites and recommendations

- Need to check if the considered variable can be applicable for all patients (e.g. checking for pregnancy, while also considering male patients in P might not be useful).

## Pitfalls and limitations

- Only having one record of the considered variable might not always be enough (e.g. when looking at the development during long period of time)
- Only checks for presence of data, not whether it's accurate, plausible, or up-to-date.

# Record completeness

Record completeness quantifies the proportion of records (rows) in a dataset that have no missing values across all considered variables. It measures the completeness of the dataset at the record level, rather than at the variable or patient level.

$$R = \frac{C_r}{N}$$

where $C_r$ is the number of complete records (no missing entries) and $N$ is the total number of records in the dataset.

**Value range:** $[0,1]$ ↑
A value of 0 indicates that no record is complete, while a value of 1 indicates that all records have no missing entries.

## Use in METRIC-framework

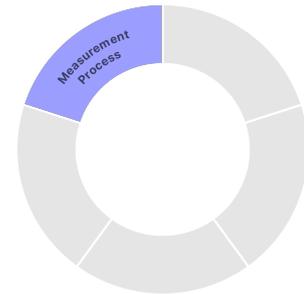

Completeness

## References
(Weiskopf et al.,2013)

## Applicability

**Modality:**

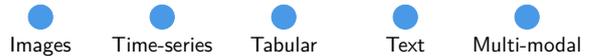

Images  Time-series  Tabular  Text  Multi-modal

**Variable:**

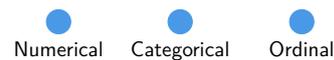

Numerical  Categorical  Ordinal

[Figure: grid showing Patient 1–4 with some missing (gray) cells]

## Prerequisites and recommendations

- Dataset must be structured in records (rows) where each record corresponds to an observation instance.
- The set of considered variables should be defined in advance.

## Pitfalls and limitations

- Does not indicate which variables are missing; only measures completeness at record level.
- Sensitive to datasets with many optional variables where missing entries are expected.
- Not meaningful if records are of variable structure or heterogeneous length.

## 9.2 Consistency

### 9.2.1 Syntactic consistency

# Syntactic accuracy

Syntactic accuracy measures how closely data values match syntactically correct domain values.

$$A = \frac{1}{N} \sum_{1}^{N} q_i$$

In a dataset with $N$ values, let $q_i$ be a boolean value, that evaluates to 1 if the value $y_i$ is in the predefined definition domain $D$ for $i \in [1, ..., N]$ and to 0 otherwise.
For database accuracy or syntactic consistency, a ratio is typically calculated between accurate values and the total number of values.

**Value range:** $[0, 1]$ ↑
A value of 1 indicates that all items in the dataset are correctly represented, while 0 indicate that no item is in the predefined value set.

## Use in METRIC-framework

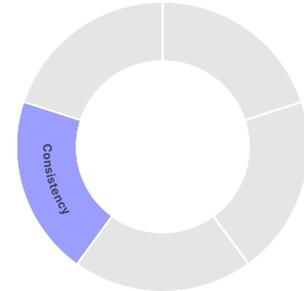

Syntactic consistency

## References
(Batini and Scannapieco, 2016)

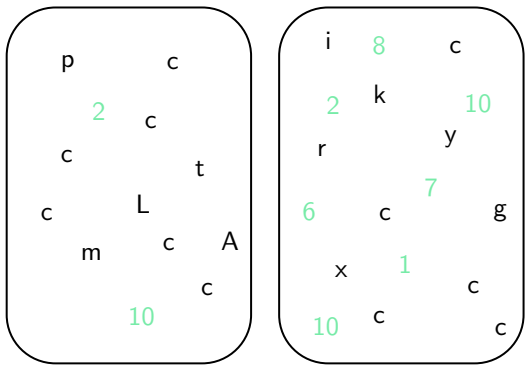

$D = \{\text{Letters}\}$

Higher ($A = 0.846$) — Lower ($A = 0.611$)

## Applicability

**Modality:**
○ Images  ○ Time-series  ● Tabular  ○ Text  ○ Multi-modal

**Variable:**
● Numerical  ● Categorical  ● Ordinal

## Prerequisites and recommendations

- For the effective use of the syntactic accuracy metric, it is essential that the dataset is well-defined with a clear domain of acceptable values.

## Pitfalls and limitations

- Sensitive to missing values, which can severely impact the accuracy assessment.
- Depends on the completeness and correctness of the predefined value sets. Incomplete or incorrect value sets can lead to misleading syntactic accuracy results.
- Generally not suitable for unstructured data types like free text, where syntactic rules are less clearly defined and more complex to implement.

### 9.2.2 Distribution drift

# Page-Hinkley Test Statistic

The PH test is used for change detection in sequential signals. It tracks cumulative deviations from the average and triggers an alarm when changes go above a certain tunable threshold.

$$m_t = \sum_{i=1}^{t}(x_i - \bar{x}_t - \delta)$$
$$PH_t = m_t - \min_{1 \leq i \leq t} m_i$$

with $\bar{x}_t = \frac{1}{t}\sum_{i=1}^{t} x_i$ and $\delta$ as the tolerable magnitude of changes.

**Value range:** $[0, \infty)$ ↑
High values indicate a significant change.

### Use in METRIC-framework

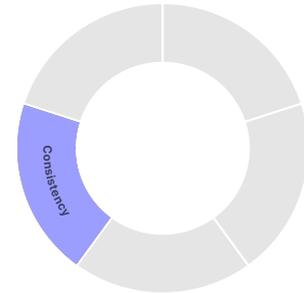

Distribution drift

### References

Mouss et al., 2004; Hinkley, 1970; Sebastião et al., 2012.

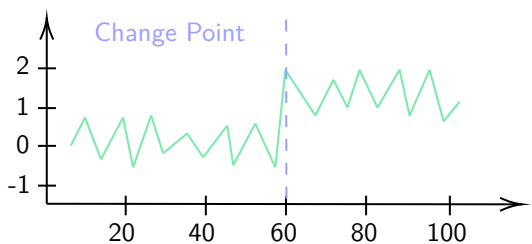

Synthetic Signal with Change Point

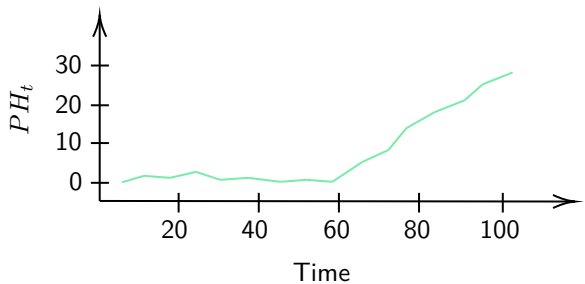

Page-Hinkley Test Statistic over Time

### Example

Monitoring of anaesthesia depth (Sebastião et al., 2012).

### Relation to other metrics

- Type of Cumulative Sum algorithm.

### Applicability

**Modality:**

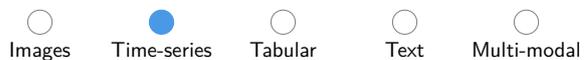

Images    Time-series    Tabular    Text    Multi-modal

**Variable:**

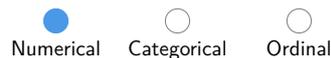

Numerical    Categorical    Ordinal

### Prerequisites and recommendations

- The sensitivity of the test depends heavily on the drift tolerance parameter $\delta$. It should be tuned based on the acceptable level of false positives and expected magnitude of changes.

### Pitfalls and limitations

- Dependence on parameter choice - Poor choice of the drift threshold ($\delta$) can result in either frequent false positives or missed detections.
- The test focuses on detecting changes in the mean. It can fail or raise false alarms if there is high variance or if the change affects only higher-order moments.
- The PH test is designed for detecting small but consistent shifts. Sudden, large changes are not necessarily detected instantly, and the test may exhibit detection lag.
- This version detects only increase of signal, not decrease. For that purpose it would be necessary to add a second trigger.

## 9.3 Representativeness

### 9.3.1 Dataset size

# Dataset size (DS)
## Synonyms: Data quantity, number of records

Dataset size (DS) measures the number of records or datapoints in a dataset.

$$DS = L, \qquad \text{for number of records } L.$$

**Value range:** $[0, +\infty)$ ↑
Value 0 indicates an empty dataset. High values indicate a larger dataset.

### Use in METRIC-framework

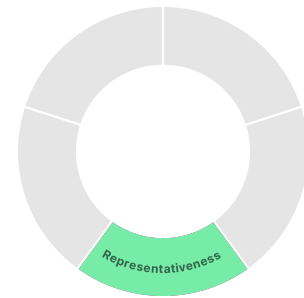

Dataset size

### References
(Han et al., 2012)

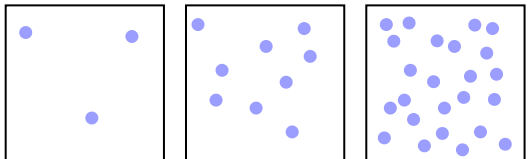

Small    Medium    Large

### Example
This study on "Sample Size Requirements for Popular Classification Algorithms in Tabular Clinical Data" lists the datasets by the number of rows, referring to dataset size (Silvey et al., 2024).

### Pitfalls and limitations
- $DS$ does not reflect the storage capacity needed to store the dataset. Size of the individual record is needed for this.
- $DS$ does not guarantee informative datapoints and hence might decrease efficiency without performance improvement.

### Applicability
**Modality:**

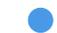 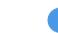 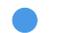 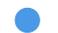 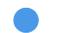
Images   Time-series   Tabular   Text   Multi-modal

**Variable:**

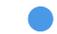 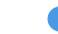 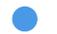
Numerical   Categorical   Ordinal

### 9.3.2 Granularity

# (Tabular) Granularity (G)
## Synonyms: Level of detail, Detail of data

Granularity reflects the level of detail for tabular data by the number of its columns.
The granularity $G$ of a tabular dataset is defined by

$$G = |Col|, \qquad \text{for set of columns } Col.$$

**Value range:** $[0, \infty) \uparrow$
The value is 0 for an empty dataset. High values of $G$ represent a dataset with many columns.

## Use in METRIC-framework

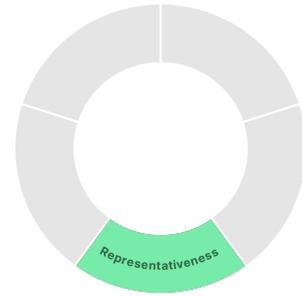

Granularity

## References
(Kimball et al., 2016)

## Example
It can be used to evaluate the granularity of clinical databases based on the number of unique systematized nomenclature (Ostropolets et al., 2021). The granularity score is calculated for each dataset, categorizing them as having low, moderate or high granularity.

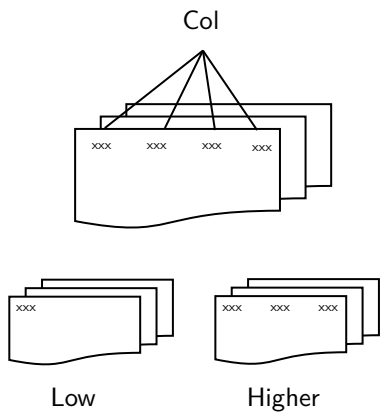

## Pitfalls and limitations
- Finer granularity does not always appear with more columns, the columns have to be independent and meaningful.
- Adding an empty or redundant column increases the value, but will affect other dimensions (e.g. completeness or feature importance).
- Finer granularity may lead to overfitting.
- Missing or noisy data can affect the depth of the data.

## Applicability
**Modality:** Tabular

**Variable:** Numerical, Categorical, Ordinal

# Sampling frequency (f)
## Synonyms: Sampling rate

The sampling frequency measures how many observations are made per unit of measurement, either in time (time-series) or space (images, volumes). It is the inverse of the sampling interval between consecutive observations or samples.
The sampling frequency $f$ can be defined as:

$$f = \frac{1}{T}, \quad \text{, where } T \text{ is the time interval between observations}$$
$$\text{or spatial sampling interval.}$$

**Value range:** $(0, \infty)$ ↑
Strictly larger than 0. Smaller values mean fewer observations per unit timeslot or spatial slot. Larger values mean more observations per unit timeslot or spatial slot, enabling finer temporal or spatial resolution.

### Use in METRIC-framework

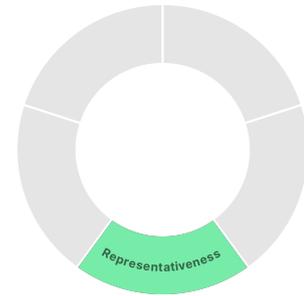

Granularity

### References
(Oppenheim et al., 1978)

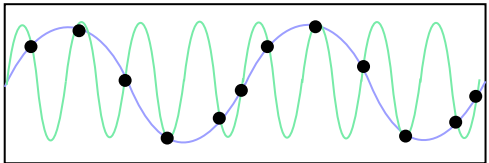

Original Signal
Reconstructed Signal

### Example
Sampling frequencies may be analyzed to evaluate how different values of T, as sampling intervals, affect heart rate variability in order to determine the optimal frequency $f$ (Kwon et al., 2018).

### Relation to other metrics
The sufficient sampling frequency $f_{min}$ ensures that the sampling frequency is high enough to accurately represent the most rapid features in the data without loss of information:

$$f_{min} = \frac{2}{T_{min}}, \quad \text{for } T_{min} \text{ being the time span.}$$

$T_{min}$ is the shortest time span of relevant morphologies in the dataset.

### Applicability
**Modality:**
○ Images  ● Time-series  ○ Tabular  ○ Text  ○ Multi-modal

**Variable:**
● Numerical  ○ Categorical  ○ Ordinal

### Prerequisites and recommendations
- The sampling frequency is defined for one time-series or one spatial dimension (image, volume).
- Knowledge of the relevant time frames, spatial scales, or resolutions is crucial.
- Consistent sampling intervals (temporal or spatial) or precise sample locations are important.
- For obtaining a dataset score, additionally inspect the distribution of sampling intervals.
- The sampling frequency should be high enough to capture meaningful changes.
- A higher sampling frequency results in more samples over the same time period or spatial domain, providing higher temporal or spatial resolution.

### Pitfalls and limitations
- Applicable only to data with approximately constant sampling intervals.
- The appropriate sampling frequency depends on the context: e.g., vital signs in ICU monitoring require higher temporal sampling than long-term blood pressure monitoring; in imaging, fine structural features require higher spatial sampling.
- Undersampling may lead to missing events or features, while oversampling may unnecessarily increase data size and computational cost.

# Image resolution (r)
## Synonyms: Picture resolution, spatial image resolution

Image resolution characterizes the sampling of an image or signal. The resolution $r$ of an image can be defined by:

$$r = (p_x, p_y) \quad \text{(for 2D)}, \quad r = (p_x, p_y, p_z) \quad \text{(for 3D)}$$

where $p_x$ is the number of samples in the width (x) direction, $p_y$ in the height (y) direction, and $p_z$ in the depth (z) direction.

**Value range:** $p_x, p_y, p_z \in (0, \infty) \quad \uparrow$
A higher number of samples along a dimension shows a higher sampling frequency, enabling finer variations to be captured, whereas fewer samples reduce representable detail.

### Use in METRIC-framework

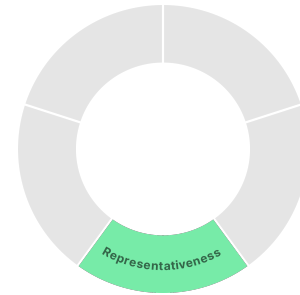

Granularity / Representativeness

### References
(Cole et al., 2022)

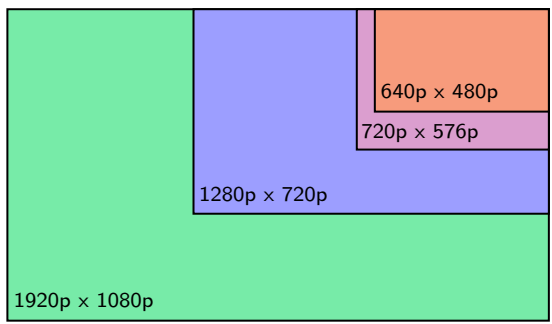

### Example
Diagnostic accuracy improves with higher spatial resolution in breast MRI, as it enables clearer delineation of small lesions, leading to more precise assessments and improved patient outcome (Eskreis-Winkler et al., 2023).

### Relation to other metrics
- Sampling Step (Synonyms: Sampling Interval, Effective Pixel/Voxel Size in 2D/3D), which equals to the sensor pixel size divided by the total system magnification in imaging

### Applicability
**Modality:** ● Images ● Time-series ○ Tabular ○ Text ○ Multi-modal

**Variable:** ● Numerical ○ Categorical ○ Ordinal

### Prerequisites and recommendations
- Resolution must always be balanced with computational capabilities and matching the resolution in real-world settings.
- Capturing the important visual features is crucial (e.g. edges, textures etc.).
- The resolution is defined for one image and for obtaining a dataset score it is useful to additionally inspect the distribution of resolutions.
- Increasing or decreasing resolution should be handled carefully to preserve the signal integrity.

### Pitfalls and limitations
- Resolution does not carry information about the real world size of the objects in the image.
- Higher resolution increases file size and computational load.
- Features might be distorted if not aspect ratio consistent.

# Label granularity (LG)

The label tree $H$ is a hierarchical data structure representing the organization of labels or classes in a dataset.
The label granularity $LG$ is defined by the number of leaves

$$LG = |leaves(H)|$$

where $H$ is a label tree graph and $leaves(H)$ is the set of all leaves in $H$.

**Value range:** $[0, \infty)$ ↑
A higher number of leaves indicate higher label granularity.

### Use in METRIC-framework

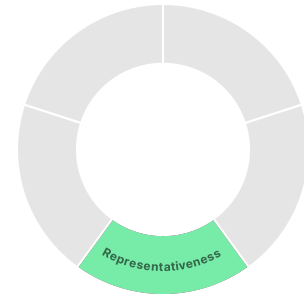

Granularity

### References
(Cole et al., 2022)

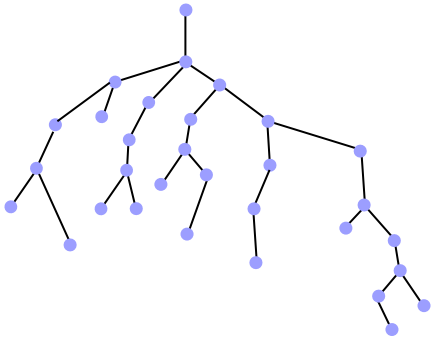

### Example
Diagnosing a respiratory issue using a three-level diagnostic hierarchy ranging from broad categories such as "abnormal" to specific conditions like "Cov19 pneumonia" involves organizing chest X-Ray findings into a structured label tree, where the most detailed diagnoses correspond to leaf nodes (Chen et al., 2020).

### Applicability

**Modality:**

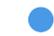 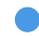 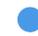 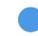 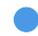
Images    Time-series   Tabular   Text   Multi-modal

**Variable:**

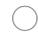 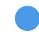 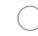
Numerical   Categorical   Ordinal

### Prerequisites and recommendations
- A structured, hierachical labeling scheme is needed.
- Consistent applying of leaves/fine-grained labels over the label tree is required.
- Fine-grained labels/leaves should be provided by experts.
- The required level of detail should be considered in its context (balanced label granularity).
- Leaves should be broad and fine-grained.

### Pitfalls and limitations
- Higher label granularity may lead to increased label noise and complexity.
- Possible imbalance by higher label granularity indicating rare classes.
- The granularity of a category is the degree to which it is specific, which can vary from coarse-grained (e.g. "animal") to fine-grained (e.g. "Persian cat").
- "Good" label granularity is difficult to define. Controversial results have been shown about the advantages of fine-grained vs. coarse-grained labels.

### 9.3.3 Target Class Balance

# Generalized Imbalance Ratio
## Synonyms: Generalized Class Ratio

The Generalized Imbalance Ratio measures the degree of class imbalance in a dataset by taking the ratio of the number of samples in the majority class to the samples in the minority class.

$$\rho = \frac{\max_i\{|C_i|\}}{\min_i\{|C_i|\}}$$

where $|C_i|$ is the number of examples in class $i$,

$\max_i\{|C_i|\}$ is the number of samples in the majority class,

$\min_i\{|C_i|\}$ is the number of samples in the minority class.

**Value range:** $[1, \infty)$ ↓
The minimum possible value of $\rho$ is 1 (indicating majority and minority classes have equal size), while there is no theoretical maximum value (as the majority class grows or the minority class shrinks, $\rho$ approaches infinity).

### Use in METRIC-framework

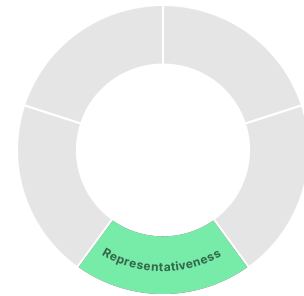

Target class balance

### References
(Buda et al., 2018)

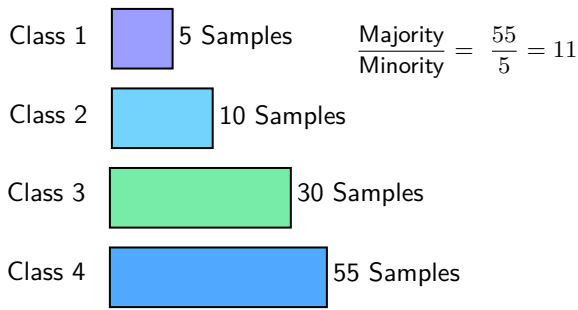

$\frac{\text{Majority}}{\text{Minority}} = \frac{55}{5} = 11$

- Class 1: 5 Samples
- Class 2: 10 Samples
- Class 3: 30 Samples
- Class 4: 55 Samples

### Example
The Generalized Imbalance Ratio was used for the class imbalance between short-(13 percent) and long-survival (87 percent) patients groups during the diagnosis of amyotrophic lateral sclerosis (Papaiz et al., 2024). Learning chest X-ray and CT images in imbalanced dataset (Iqbal et al., 2023).

### Relation to other metrics
- Multi-class generalization of Imbalance Ratio (IR)

### Applicability
**Modality:**

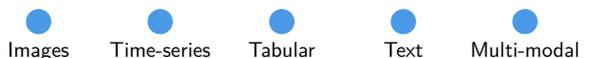

Images | Time-series | Tabular | Text | Multi-modal

**Variable:**

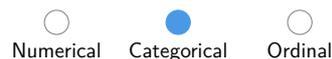

Numerical | Categorical | Ordinal

### Prerequisites and recommendations
- Well-defined class labels: The dataset must have clear, mutually exclusive class labels
- Complete class count information: We need the exact counts for all classes.

### Pitfalls and limitations
- Ignores intermediate classes: Only considers the ratio between majority and minority groups, missing nuances in class distribution.

# Imbalance Degree (ID)

The ID quantifies the deviation of an observed class distribution from a total balanced distribution by combining a distance measure with the number of minority classes. Unlike simple imbalance ratios, ID captures the overall distribution deviation and the complexity presented by minority classes.

$$\text{ID} = \frac{d(\hat{\mathbf{p}}, \mathbf{b})}{d(\mathbf{p}_m, \mathbf{b})} + (m-1)$$

**Where:**
- $C$: total number of classes
- $m$: number of minority classes (classes with $\hat{p}_c < \frac{1}{C}$)
- $\hat{\mathbf{p}} = [\hat{p}_1, \hat{p}_2, \ldots, \hat{p}_C]$: observed class distribution vector
- $\mathbf{b} = [\frac{1}{C}, \frac{1}{C}, \ldots, \frac{1}{C}]$: perfectly balanced class distribution
- $\mathbf{p}_m$: the distribution with exactly $m$ minority classes that achieves the maximum distance to $\mathbf{b}$
- $d(\cdot, \cdot)$: distance between two distributions

The maximum-distance distribution $\mathbf{p}_m$ can be realized as:

$$\mathbf{p}_m = [0, \ldots, 0, \underbrace{\frac{1}{C}, \ldots, \frac{1}{C}}_{C-m-1}, 1 - \frac{C-m-1}{C}]$$

**Value range:** $[0, \infty)$ ↓
Value 0 for perfectly balanced classes, no minority classes, higher ID for greater imbalance and/or more minority classes.

## Use in METRIC-framework

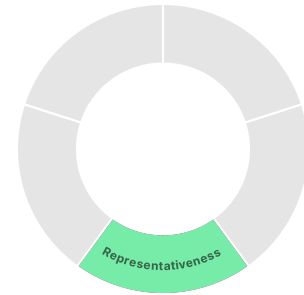

Target class balance

## References

(Ortigosa-Hernández et al., 2017)

## Prerequisites and recommendations

- Well-defined class labels: The dataset must have clear, mutually exclusive class labels.
- Choice of distance function: Selection of an appropriate distance function (like Hellinger, Total Variation, etc.) to measure the deviation from balanced distribution.
- Complete class distribution information: You need to know the number of samples in each class to calculate the empirical distribution.

## Pitfalls and limitations

- Dependence on parameter choice - the first component of ID relies on a distance metric to measure the difference between distributions.
- ID only works for comparing datasets with the same number of classes; two datasets having different numbers of classes but the same number of minority classes can still have the same ID value.

## Applicability

**Modality:** Images, Time-series, Tabular, Text, Multi-modal

**Variable:** Categorical

# Likelihood Ratio Imbalance Degree (LRID)

The LRID quantifies the deviation of a class distribution from a total balanced distribution using a likelihood-ratio test. Unlike the imbalance ratio (IR), LRID considers the entire distribution.

$$\text{LRID} = -2 \sum_{c=1}^{C} n_c \ln \frac{b_c}{\hat{p}_c} = -2 \sum_{c=1}^{C} n_c \ln \frac{N}{C n_c}$$

where:

- $C$ is the number of classes
- $n_c$ is the number of observations in class $c$
- $N = \sum_{c=1}^{C} n_c$ is the total number of observations
- $\hat{p}_c = \frac{n_c}{N}$ is the observed probability of class $c$
- $b_c = \frac{1}{C}$ is the balanced probability for class $c$
- The expression is derived from the likelihood-ratio test comparing $H_0 : \mathbf{p} = \mathbf{b}$ versus $H_1 : \mathbf{p} = \hat{\mathbf{p}}$

**Value range:** $[0, \infty) \downarrow$
For balanced data, LRID equals 0; for imbalanced data, LRID is greater than 0.

## Use in METRIC-framework

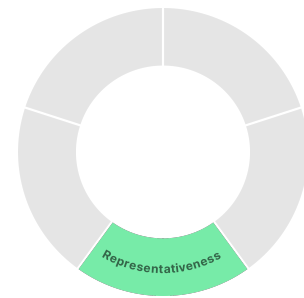

Target class balance

## References
(Zhu et al., 2018)

## Prerequisites and recommendations

- Access to class frequency information - We need the raw counts ($n_c$) of observations in each class, or at least the relative frequency distribution and total sample size.
- Each observation must belong to exactly one class - LRID assumes a traditional classification setup where each sample belongs to a single class.
- Mathematical prerequisite - When implementing handle the case of classes with zero observations carefully to avoid division by 0 issues.

## Pitfalls and limitations

- LRID cannot be used to compare class imbalance across datasets with different numbers of classes due to its dependence on the uniform reference distribution $1/C$, which changes with $C$.

## Applicability

**Modality:**
● Images  ● Time-series  ● Tabular  ● Text  ● Multi-modal

**Variable:**
○ Numerical  ● Categorical  ○ Ordinal

## 9.4 Timeliness

### 9.4.1 Currency

# Currency by Ballou et al.

Currency by Ballou et al. measures the extent to which the data values are up-to-date. It decreases with a polynomial decay of degree $s$.

$$C(v) = \left[\max\left(0,\ (1 - \frac{t^{\text{cur}} - t^{\text{update}}(v)}{T^{\text{exp}}(v)})\right)\right]^s,$$

where $T^{\text{exp}}(v)$ is the valid lifetime of the data value $v$, $t^{\text{cur}}$ is the current time, and $t^{\text{update}}(v)$ is the time of the last update of $v$, $s \in \mathbb{N}$.

**Value range:** $[0, 1]$ ↑
High values, close to 1, indicate perfectly up-to-date data (freshly updated), while values close to 0 mean the data is outdated beyond its expected lifetime.

## Use in METRIC-framework

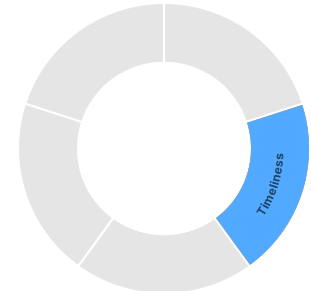

Currency

## References
(Ballou et al., 1998)

## Example
Using a small amount of recent clinical data has been shown to be more effective for predicting future clinical decisions than using larger volumes of older data, highlighting the diminishing relevance of outdated information (Chen et al., 2017).

## Applicability

**Modality:**
● Images  ● Time-series  ● Tabular  ● Text  ● Multi-modal

**Variable:**
● Numerical  ● Categorical  ● Ordinal

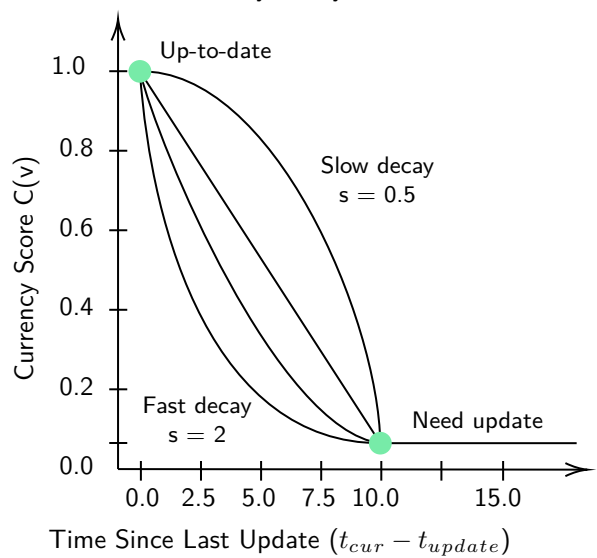

## Prerequisites and recommendations
- Each datapoint needs to include metadata indicating its last update time and its valid lifetime.

## Pitfalls and limitations
- Dependence on parameter choice - Valid lifetime has strong influence on the metric value but is subject to inaccuracies.
- Low impact of data volatility

# Currency by Li et al.

Currency by Li et al. measures the extent to which the data values are up-to-date. The concept of decreasing relevance is emphasized by incorporating both the time since a value's last update and the volatility (likelihood of change).

$$C(v) = \max[0, \left(1 - \frac{t^{\text{cur}} - t^{\text{update}}(v)}{T^{\text{exp}}(v)}\right) e^{-V}],$$

where $T^{\text{exp}}(v)$ is the valid lifetime of the data value $v$, $t^{\text{cur}}$ the current time, $t^{\text{update}}(v)$ the time of last update of $v$, and $V$ the volatility.

**Value range:** $[0,1]$ ↑
High values, close to 1, indicate perfectly up-to-date data (freshly updated), while values close to 0 mean the data is outdated beyond its expected lifetime.

## Use in METRIC-framework

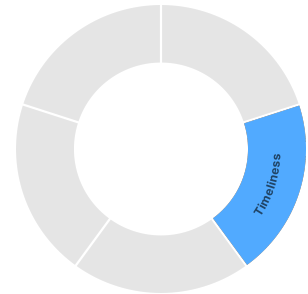

Currency

## References
(Li et al., 2012)

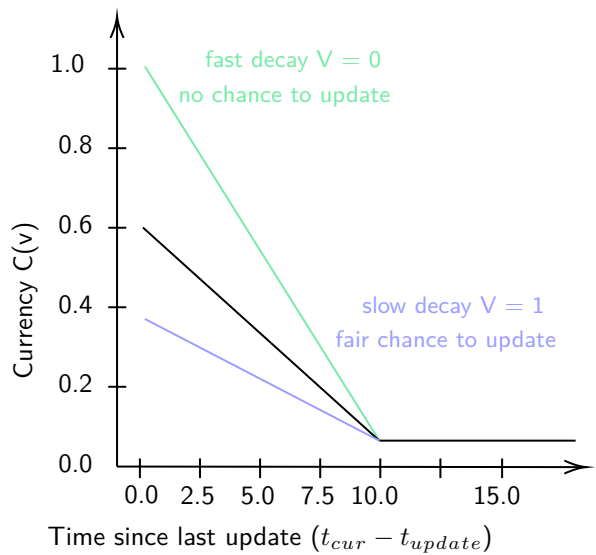

## Example
It is acknowledged that patient information becomes less informative as the time since last update increases (Chen et al., 2017, Wang et al., 2024). Therefore models incorporate various time gap-based decay functions to weigh older clinical visits or diagnosis codes less heavily in predictions.

## Applicability
**Modality:** Images, Time-series, Tabular, Text, Multi-modal

**Variable:** Numerical, Categorical, Ordinal

## Prerequisites and recommendations
- Each datapoint should include metadata indicating its last update time and its valid lifetime.
- To calculate the volatility $V$, access to the complete update history is required.

## Pitfalls and limitations
- Dependence on parameter choice
- Volatility has strong influence on the metric and hence needs to be chosen carefully.
- Valid lifetime has strong influence on the metric value but is subject to inaccuracies.

# Currency by Hinrichs

Currency by Hinrichs measures the extent to which the data values are up-to-date. The Currency of a data value $v$ is defined as inversely proportional to both its mean attribute update time $U(v)$ and its age $A(v)$.

$$C(v) = \frac{1}{U(v) \cdot A(v) + 1},$$

where $U(v)$ is the mean attribute update time, meaning the update frequency needed for the value $v$ to stay up-to-date, and $A(v)$ is the age of data value $v$.

**Value range:** $(0, 1]$ ↑
This metric asserts its maximum value when either the value is time independent (such as a field storing the names of patients) or when the data value is new. Note that this metric can not assert its minimum value, but tends to 0 for data needing frequent update and increasing age.

### Use in METRIC-framework

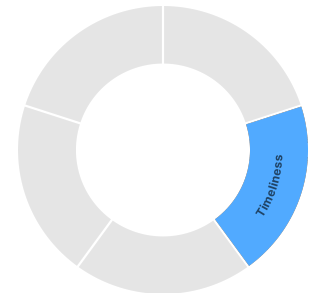

Currency

### References

(Hinrichs,2002,Heinrich et al., 2007)

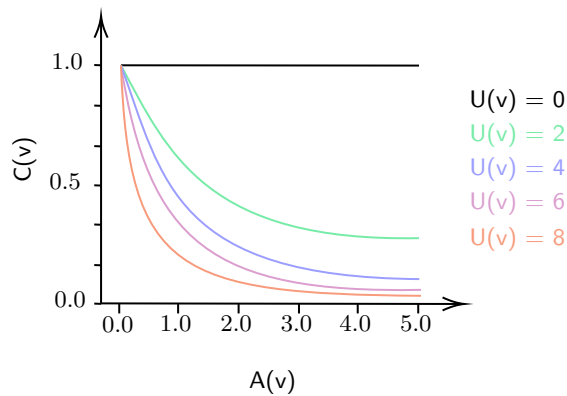

### Example

The delay between clinical events and their documentation in a patient portal for type II diabetes care has been quantified (Wiley et al., 2024). This study showed that portal users had more up-to-date records, with significantly shorter time lags between clinical actions and data entry, highlighting that more recent and less frequently updated data values exhibit higher currency.

### Applicability

**Modality:**

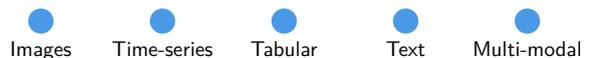

Images    Time-series    Tabular    Text    Multi-modal

**Variable:**

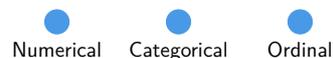

Numerical    Categorical    Ordinal

### Prerequisites and recommendations

- Each datapoint should include metadata indicating its update times.

### Pitfalls and limitations

- Dependence on parameter choice
- The historical update frequency may not be easy to measure.
- The needed update frequency associated to a variable may be varying over time.

# Currency by Heinrich et al.

Currency by Heinrich et al. measures the extent to which the data values are up-to-date. It quantifies the likelihood that a data value $v$ of an attribute $A$ is still valid at the time of retrieval. It decays exponentially with the age of the value, modulated by a decline rate $\text{decline}(A)$ specific to the attribute.

$$Q_{\text{Time}}(v, A) = \exp\left(-\text{decline}(A) \cdot \text{age}(v, A)\right)$$

**Value range:** $(0, 1]$ ↑
Value 1 is reached if the decline rate is 0 or the age of the value is 0. The metric value tends to 0 as $age(v, A)$ approaches infinity. It approaches 0 faster for higher decline rates.

## Use in METRIC-framework

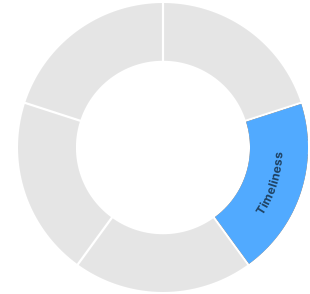

Currency

## References

(Heinrich et al., 2007)

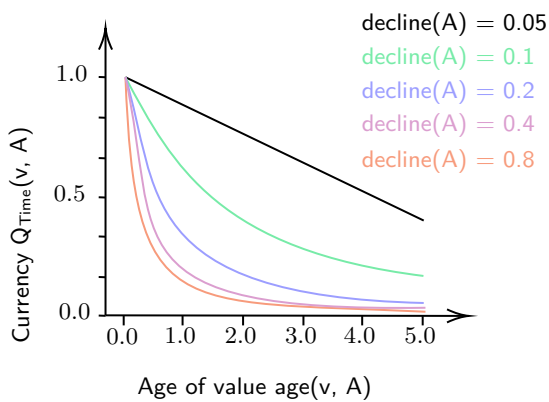

Exponential Currency Delay Over Time
- $\text{decline}(A) = 0.05$
- $\text{decline}(A) = 0.1$
- $\text{decline}(A) = 0.2$
- $\text{decline}(A) = 0.4$
- $\text{decline}(A) = 0.8$

## Applicability

**Modality:** Images, Time-series, Tabular, Text, Multi-modal

**Variable:** Numerical, Categorical, Ordinal

## Prerequisites and recommendations

- Each datapoint should include metadata indicating its age.

## Pitfalls and limitations

- Dependence on scale of the age variable (is age measured in seconds, hours, days, ...?)
- Dependence on parameter choice - Appropriate choice of decline parameter according to age scale is necessary due to exponential dependency.
- Does not compensate for update frequency or last update time
- Does not compensate for data volatility

## 9.5 Informativeness

### 9.5.1 Uniqueness

# The prevalence of duplicates (PD)

This metric represents the proportion of instances that appear more than once in the dataset, expressed as a ratio to the total number of instances:

$$PD = \frac{D}{N},$$

where $D$ is the number of duplicates and $N$ is the number of instances in the dataset.

**Value range:** $[0, 1]$ ↑
High values close to 1 are the worst case, representing a dataset with one unique instance; value 0 for every instance being unique.

## Use in METRIC-framework

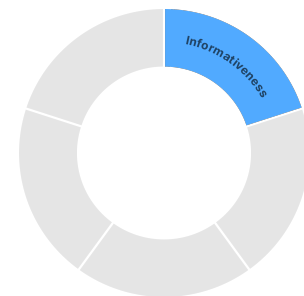

Uniqueness

## References
(Qi et al., 2013)

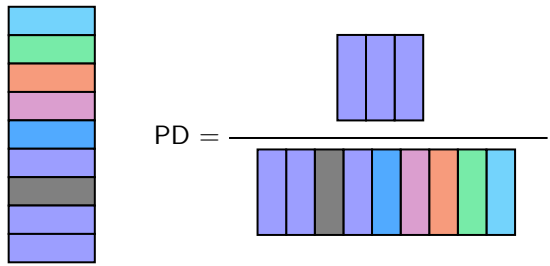

## Example
Used in data cleaning pipelines to assess data redundancy in EHRs or registries (Steinkamp, 2022). Can be easily implemented from common functions in statistical packages (e.g., SPSS, R, Python pandas|).

## Applicability
**Modality:**
● Images ● Time-series ● Tabular ● Text ● Multi-modal

**Variable:**
● Numerical ● Categorical ● Ordinal

## Prerequisites and recommendations
- It requires a clear definition of duplication logic: exact match, fuzzy match, partial match, etc.

## Pitfalls and limitations
- Unstable for small sample size
- Sensitive to missing values - may coincide with missing fields (completeness) of the data set.
- May flag legitimate redundancy (e.g., repeated visits by a patient)

# The effective sample size (ESS)

Reflects the loss of information due to serial correlation between samples.

$$ESS = \frac{N}{1 + 2\sum_{k=1}^{T} \rho_k},$$

where $N$ is the number of points in a chain/time series and $\rho_k$ is the autocorrelation at lag $k$, truncation lag $T$.

**Value range:** $(0, N]$ ↑
High values close to N, indicating low autocorrelation, samples are nearly independent. In contrast, values close to 0 indicate high autocorrelation, meaning that there are fewer effective samples and estimates are less reliable.

## Use in METRIC-framework

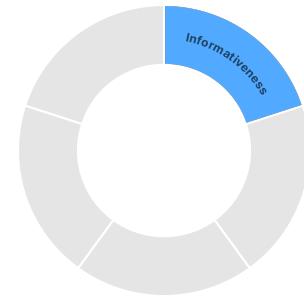

Uniqueness

## References
(Thompson, 2012)

## Applicability

**Modality:**
○ Images  ● Time-series  ○ Tabular  ○ Text  ○ Multi-modal

**Variable:**
● Numerical  ○ Categorical  ○ Ordinal

## Example
Applications in Markov chain Monte-Carlo modelling and bayesian workflow literature (Gelman et al., 2020). For a sequence of *non-independent, autocorrelated* observations, estimates the sample size necessary for the MCMC CLT to hold. For balancing under/overrepresented survey responses in epidemiology (Zhang et al, 2024). Tools (e.g., PyMC, ArviZ) include implementations of ESS.

## Prerequisites and recommendations
- Requires estimation of autocorrelations $\rho_k$ at multiple lags.
- Requires to choose a reasonable truncation lag $T$; to choose a cutoff lag $T$ up to which to include $\rho_k$ in the ESS formula (e.g. using Geyer's initial positive sequence or setting a threshold for significance of autocorrelation).

## Pitfalls and limitations
- Dependence on parameter choice - sensitive to lag selection $T$: too small $T$ miss autocorrelation at higher lags that still contributes to dependency, too high $T$ the sum unnecessarily large due to inclusion of random fluctuations at high lags.
- Unstable for small sample size.
- There are alternative definitions of ESS which are applicable to non-sequential data. The paper by Acosta et al, 2021 introduces a definition of ESS (with a block likelihood estimation approach) which is appropriate for large spatial datasets and reduces to the original ESS when the number of blocks is equal to one.

### 9.5.2 Informative missingness

# Little's test

A chi-square-based test to assess whether missing data are missing completely at random (MCAR) (Schafer and Graham) or if there are category-dependant patterns in missing data. Under the MCAR null-hypothesis at each time point the calculated means of the observed data should be the same irrespective of the pattern of missingness.

$$d_0^2 = \sum_{p=1}^{P} N_p \cdot \left(\bar{X}_p - \mu_p\right)^T \Sigma_p^{-1} \left(\bar{X}_p - \mu_p\right),$$

where $P$ is the number of distinct missing data patterns, $N_p$ the number of observations in pattern $p$, $\bar{X}_p$ the sample mean vector of observed variables in pattern $p$, $\mu_p$ the corresponding subvector of the global mean vector $\mu$ restricted to observed components in pattern $p$, and $\Sigma_p$ the estimated covariance matrix of the observed components in pattern $p$.

Under the null hypothesis (MCAR), $d_0^2$ approximately follows a chi-squared distribution with degrees of freedom $\sum_{j=1}^{P} p_j - p$, where $p_j$ is the number of observed variables in pattern $p$, and $p$ is the total number of variables in the dataset.

**Value range (associated p-value):** $[0, 1]$ ↑
A non-significant p-value (typically p>0,05) indicates that the data is consistent with the MCAR assumption. A significant p-value indicates that it may not be MCAR.

### Use in METRIC-framework

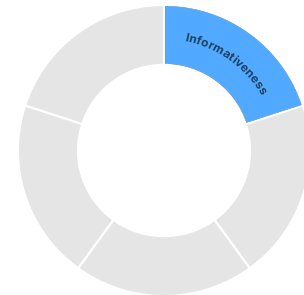

Informative missingness

### References
(Little, 1988)

### Example
Quantifying and qualifying the amount of missing data in surveys of healthcare workers (Synsky et al, 2021) and patients (Cooper et al, 2021) during the COVID pandemic.

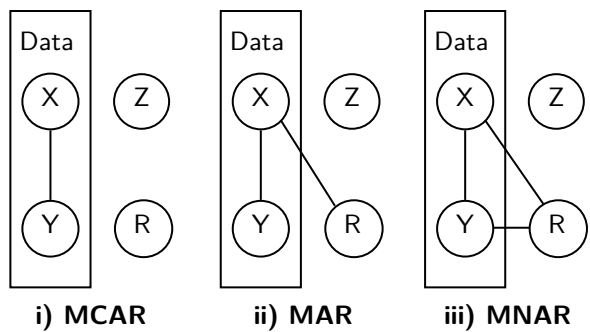

i) MCAR    ii) MAR    iii) MNAR

Source: Modified from Schafer and Graham

### Applicability

**Modality:**
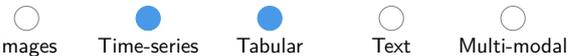
Images    Time-series    Tabular    Text    Multi-modal

**Variable:**
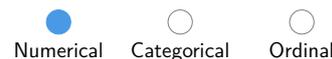
Numerical    Categorical    Ordinal

### Prerequisites and recommendations

- The test assumes multivariate normality of data. Violations of this assumption can lead to unreliable results, especially with highly skewed or categorical data.
- Little's test can be applied directly to ordinal data (which is discrete and bounded), even though this is a violation of its assumptions. However, if the ordinal variables have enough categories (typically 5+), such variables can be treated as continuous, making it possible to apply Little's test.
- Otherwise, instead of Little's test, diagnostic measures can be used from multiple imputation procedures designed for categorical/ordinal data.

### Pitfalls and limitations

- Sensitive to outliers - outliers can significantly impact covariance estimates and test results in multivariate settings.
- Dependence on parameter choice
- Violations of the normality assumption can lead to unreliable results, especially with highly skewed or categorical data.
- Cannot distinguish MAR from MNAR - If the MCAR hypothesis is rejected, the test will not tell if data is Missing At Random (MAR) or Missing Not At Random (MNAR).

# Joint likelihood model for informative dropout (ID)

Characterizes *informative dropout* (ID), when the missingness of data depends on the data itself – in opposition to MCAR (Schafer and Graham).
Joint likelihood model for the *outcome*:

$$Y = f(X, \beta) + \epsilon$$

with Y the outcome variable, X the observed covariates, $\beta$ the regression coefficient for the predictors and $\epsilon$ the error term.
Joint likelihood model for the *missingness*:

$$P(M = 1|Y, X) = g(Y, X, \delta)$$

with M the missingness indicator (M=1 if missing, M=0 if observed), g a link function (e.g. logistic) and $\delta$ the parameters capturing the dependence of missingness on Y and X.
The joint likelihood is modeled by

$$L(\beta, \delta) = \prod_{i=1}^{n} f(Y_i|X_i, \beta) \times P(M_i|Y_i, X_i, \delta)$$

and maximized to estimate the parameters $\beta$ and $\delta$.

**Value range:** $[0, K] \uparrow$
The upper bound $K$ for the joint likelihood is determined by the specific distributional assumptions, but is always finite.

## Use in METRIC-framework

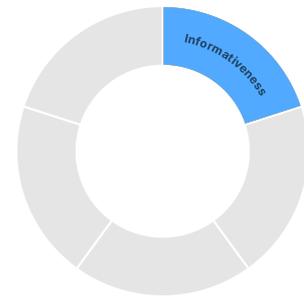

Informative missingness

## References

(Diggle and Kenward, 1994)

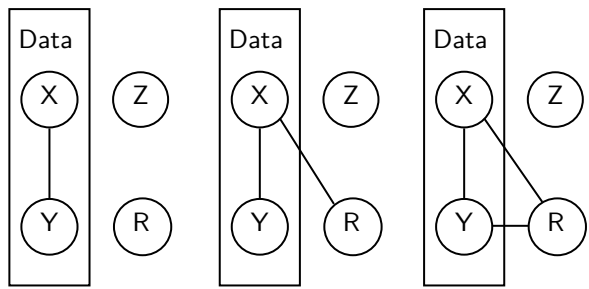

**i) MCAR**    **ii) MAR**    **iii) MNAR**

Source: Modified from Schafer and Graham

## Applicability

**Modality:**

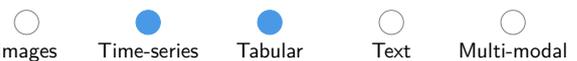

Images    Time-series    Tabular    Text    Multi-modal

**Variable:**

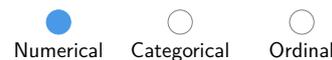

Numerical    Categorical    Ordinal

## Prerequisites and recommendations

- The selection model models the joint likelihood of the outcome and the missing indicator by interconnecting two model equations. Evaluation involves interpreting the coefficients of the missingness model to understand how unobserved outcomes affect the probability of missingness.
- A significant association between missingness and unobserved outcomes supports the presence of MNAR (missing not at random). Model fit is assessed using likelihood ratio tests, AIC/BIC, or cross-validation, ensuring that the joint model appropriately captures the data structure and missingness pattern.

## Pitfalls and limitations

- Dependence on parameter choice - correct specification of both the measurement model and the dropout mechanism is crucial
- Unstable for imbalanced data - if groups are highly imbalanced, dropout mechanism estimation can be unstable, affecting MAR vs. MNAR inference.
- Very often, insufficient information in the data does not allow to distinguish between missing at random (MAR) and missing not at random (MNAR) mechanisms.
- Results depend heavily on the chosen functional form relating the outcome to the dropout probability.

## 9.6 Distribution metrics

# Range

The range is defined as difference between the maximum and the minimum value of a variable in the data and describes its spread.

$$r = max_{v \in V}(v) - min_{v \in V}(v), \qquad \text{with values } v \in V.$$

**Value range:** $[0, +\infty)$ ↑
Low values indicate that only a small range is covered, high values indicate that a large range is covered.

### Use in METRIC-framework

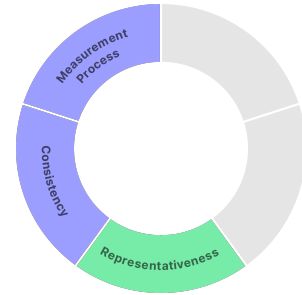

Variety
Target class balance
Homogeneity
Distribution drift

### References
(Freedman, 2007, Fahrmeier et al., 2016)

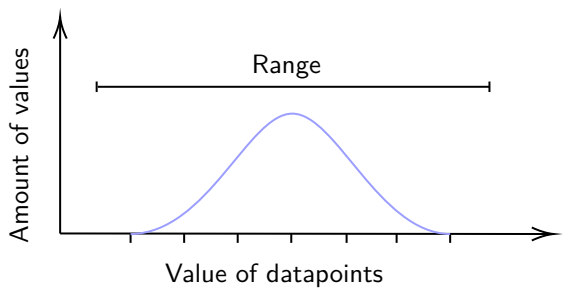

Distribution of values in a dataset

### Example
In a kinematic analysis on lumbar spine motion the range of motion (ROM) was calculated by measuring the difference between the maximum and minimum trunk angles during flexion and extension tasks (Villalba-Meneses et al., 2024). The range quantified mobility, identified abnormal motion patterns and was used for classification models to distinguish between pathological and normal movements.

### Applicability
**Modality:**
● Images ● Time-series ● Tabular ○ Text ○ Multi-modal

**Variable:**
● Numerical ○ Categorical ● Ordinal

### Prerequisites and recommendations
- Most useful for symmetric distributions with no extreme outliers.

### Pitfalls and limitations
- Unstable for small sample size.
- Sensitive to outliers.
- Does not provide information about the shape of distribution.

# Interquartile range (IQR)

The difference between the 75th percentile ($Q_3$) and the 25th percentile ($Q_1$) of a variable in the data describes its spread.

$$IQR = Q_3 - Q_1$$

**Value range:** $[0, +\infty)$ ↑
Low values indicate that majority of the values are within a narrow range, high values indicate that majority of the values covered by large range.

## Use in METRIC-framework

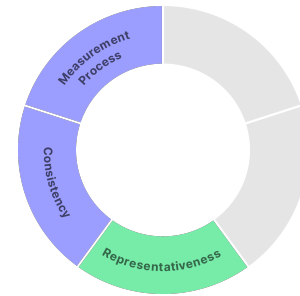

Variety
Target class balance
Homogeneity
Distribution drift

## References

(Freedman, 2007, Fahrmeier et al., 2016)

## Example

The IQR served as a quantitative imaging feature to characterize the spread of voxel intensity values in brain MRI images, enabling the differentiation between glioma subtypes (brain tumors). IQR showed the highest diagnostic performance in distinguishing between astrocytoma and glioblastoma (Sun et al., 2023).

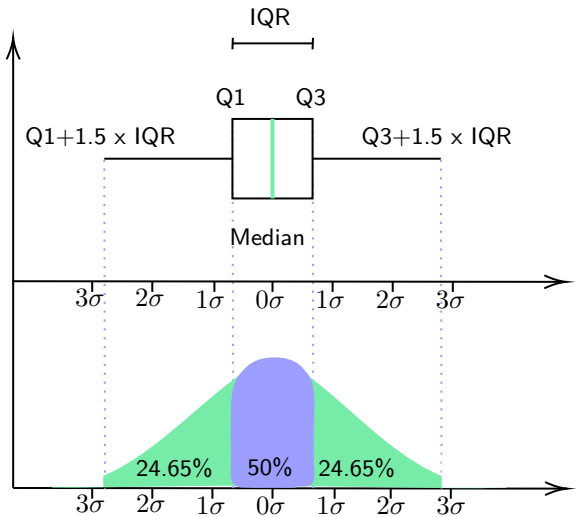

## Applicability

**Modality:**

● Images  ● Time-series  ● Tabular  ○ Text  ○ Multi-modal

**Variable:**

● Numerical  ○ Categorical  ● Ordinal

## Prerequisites and recommendations

- More stable than the range in values between the minimum and the maximum.
- Might be a better choice than standard deviation for distributions with outliers.

## Pitfalls and limitations

- Unstable for small sample size.
- Does not provide information about the shape of distribution.

# Mean and standard deviation

The standard deviation measures the spread of values $V = (v_1, ..., v_N) \in \mathbb{R}$ around their mean value $\mu$. It quantifies the variability of values in $V$.

$$\sigma = \sqrt{\frac{1}{n}\sum_{i=1}^{n}(v_i - \mu)^2}$$

$$\mu = \frac{1}{n}\sum_{i=1}^{n} v_i$$

**Value range:** $[0, +\infty) \uparrow$
A low standard deviation implies a low spread of values, a high standard deviation implies a high spread of values.

## Use in METRIC-framework

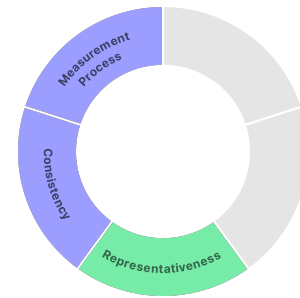

Variety
Target class balance
Homogeneity
Distribution drift

## References

(Freedman, 2007, Fahrmeier et al., 2016)

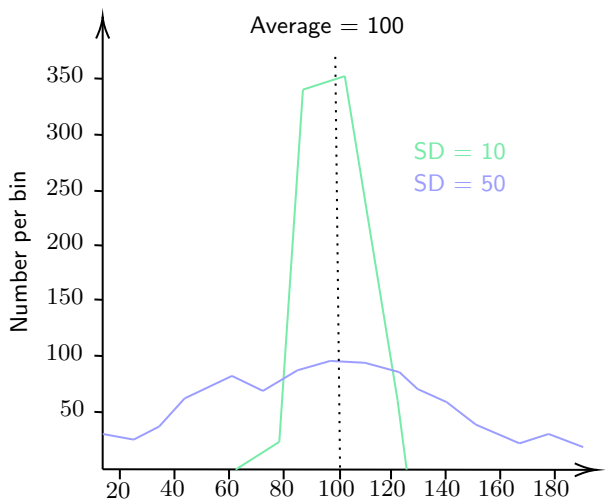

Example of samples from two populations with the same mean but different standard devitations. Green population has mean 100 and SD 10; but blue population has mean 100 and SD 50.

## Example

In population studies, the standard deviation is calculated using the population size, individual observations and the population mean. It quantifies the average deviation of each data point from the mean, offering insights into the spread and consistency of measurements within the population (El Omeda et al., 2025).

## Relation to other metrics

- Standard deviation is the square root of variance.

## Applicability

**Modality:**
● Images   ● Time-series   ● Tabular   ○ Text   ○ Multi-modal

**Variable:**
● Numerical   ○ Categorical   ○ Ordinal

## Prerequisites and recommendations

- When a comparison between two distributions is attempted in terms of standard deviation, care should be taken to verify that the distributions have a similar shape.

## Pitfalls and limitations

- Sensitive to outliers.
- If the distribution of the data is not symmetrical or outliers are included in the data, the standard deviation might not provide an useful result for its spread. Consider using IQR instead.

# Hill numbers
## Synonyms: True diversity, Effective number of types

The Hill numbers quantify diversity by comparing the relative abundance of different types in a dataset using the generalized mean.
The Hill numbers $^qD$ of order $q$ and $v_n \in \{1,...,R\}$ are defined by:

$$^qD = (\sum_{i=1}^{R} p_i^q)^{\frac{1}{1-q}}$$

with $p_i = \frac{|v \in V : v = \sigma_i|}{N}$, values $V = (v_1,...,v_N) \in \Omega^N$ and $\Omega = (\sigma_1,...,\sigma_R)$.

**Value range:** $[1, R]$ ↑
Value 1 means that there is only one type in the data. Value $R$ is attained for equal proportional abundance of all types, where $R \in \mathbb{N}$ is the number of types expected in the data.

### Use in METRIC-framework

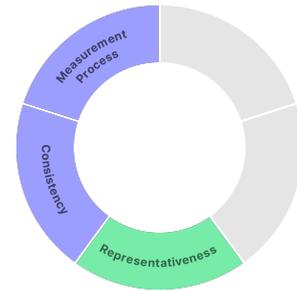

Variety
Target class balance
Homogeneity
Distribution drift

### References
(Ricotta et al., 2021)

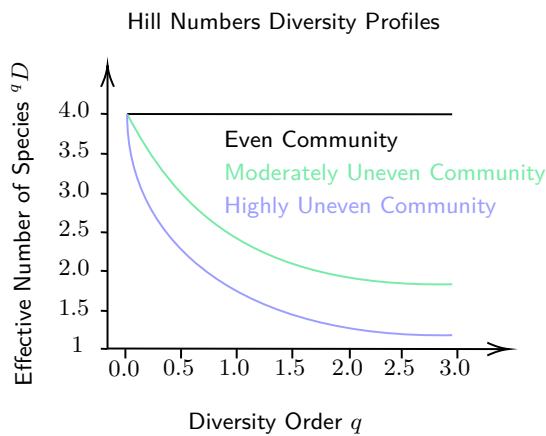

### Example
Microbiome diversity can be quantified using Hill numbers to assess microbiological heterogeneity between patients with ulcerative colitis, a microbiome-associated disease, and healthy patients (Li et al., 2021).

### Relation to other metrics
- $q = 0$: Species richness
  (Counts different types in dataset.)
- $q = 1$: Shannon index
  (Balances common and rare types.)
- $q = 2$: Simpson index
  (Gives more weight to dominant types.)
- Gini-Simpson index $= 1 - \frac{1}{^2D}$
- Rényi entropy:

$$^qH = \frac{1}{1-q} \ln(\sum_{i=1}^{R} p_i^q)$$

### Applicability
**Modality:**
○ Images  ○ Time-series  ● Tabular  ○ Text  ○ Multi-modal

**Variable:**
○ Numerical  ● Categorical  ● Ordinal

### Prerequisites and recommendations
- Assumes independent and well-defined types.
- Use lower values of $q$ to focus on rare types, higher values of $q$ to focus on dominant types.
- Compare several values of $q$ to gain a broader picture.

### Pitfalls and limitations
- Dependence on parameter choice - choice of diversity order $q$.
- Comparing Hill numbers for datasets with a different number of types might lead to misinterpretation.
- Discretizing numerical features into buckets and calculating Hill numbers can lead to misinterpretation, as it introduces sensitivity to bucket size and disregards the inherent order of the data.
- Promotes uniform distribution, which might not always be desirable.

# Maximum Mean Discrepancy
## Synonyms: MMD

Maximum Mean Discrepancy measures distance between two distributions.

$$\text{MMD}^2(P, Q) = \|\mu_P - \mu_Q\|_{\mathcal{H}}^2$$

with $P$, $Q$ as the distributions of subsets of variables in the dataset, $\mathcal{H}$ as the reproducing kernel Hilbert space, $\mu_P$, $\mu_Q$ as the mean embeddings of $P$ and $Q$ in $\mathcal{H}$. MMD quantifies the distance between two distributions by calculating the distance between their mean embeddings in a reproducing kernel Hilbert space. MMD is sometimes used in the context of GANs, e.g. as a simple discriminator or to assess their performance.

**Value range:** $[0, \infty)$
Higher values indicate dissimilarity, 0 indicates that the two distributions are identical.

### Use in METRIC-framework

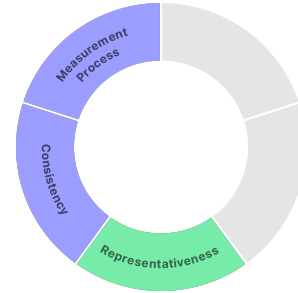

Variety
Target class balance
Homogeneity
Distribution drift

### References
(Gretton, 2012)

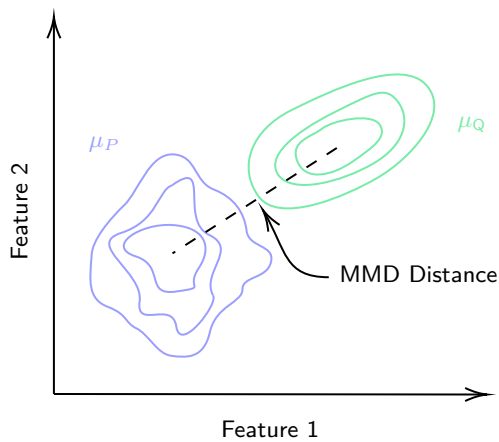

### Relation to other metrics
- Wasserstein distance (both measure distributional differences)

### Applicability
**Modality:**

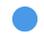 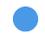 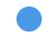 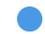 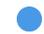
Images   Time-series   Tabular   Text   Multi-modal

**Variable:**

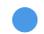 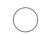 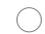
Numerical   Categorical   Ordinal

### Prerequisites and recommendations
- Statistical hypothesis tests can be performed with MMD.

### Pitfalls and limitations
- Dependence on parameter choice - The parameters of the kernel, such as bandwidth in the Gaussian kernel, need careful tuning. Improper parameters can lead to an inability to detect significant differences or overly sensitive results that suggest differences where none exist.
- Biased and linear variants, although easier to compute, yield statistical tests with a lower power than the unbiased MMD.
- Calculation for large datasets is computationally expensive.

# Cohen's $d$
## Synonyms: Effect size

Cohen's d is a measure of effect size that quantifies the difference between two group means in terms of standard deviation units.

$$\text{Cohen's } d = \frac{\bar{x}_1 - \bar{x}_2}{s}$$

with $s = \sqrt{\frac{(n_1-1)s_1^2 + (n_2-1)s_2^2}{n_1+n_2-2}}$ as the pooled standard deviation, $s_1^2, s_2^2$ the variances of the two groups, $\bar{x}_1$ and $\bar{x}_2$ the population means. It is helpful to assess the practical relevance of significant differences in means.

**Value range:** $(-\infty, \infty)$ ↑
Value 0 refers to identical distribution means, larger absolute values refer to larger difference between distributions.

### Use in METRIC-framework

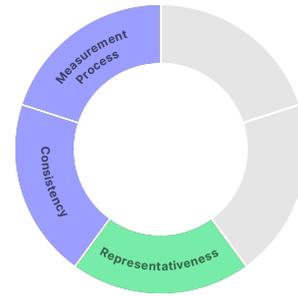

Variety
Target class balance
Homogeneity
Distribution drift

### References
(Cohen, 1988; Lakens, 2013)

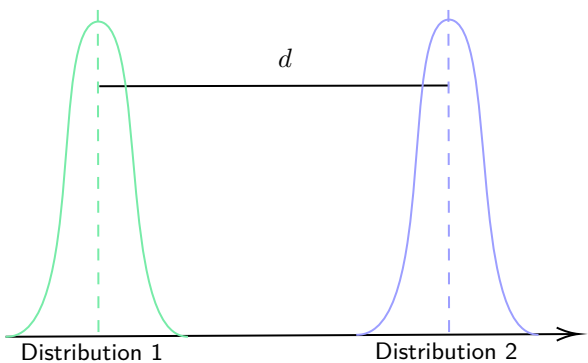

Distribution 1    Distribution 2

### Example
A study by Larner et al. compared multiple cognitive screening instruments using data from five separate clinical diagnostic cohorts. They calculated Cohen's d to assess diagnostic consistency (Larner, 2014).

### Relation to other metrics
Cohen'd is the univariate special case of Mahalanobis distance.

### Applicability
**Modality:**

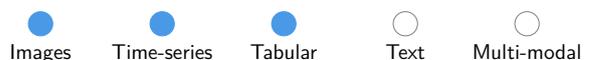

Images   Time-series   Tabular   Text   Multi-modal

**Variable:**

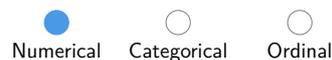

Numerical   Categorical   Ordinal

### Prerequisites and recommendations
- A good interpretation of Cohen's $d$ is to compare the observed effect size to other effect sizes in the literature and assess their impact of the observed effect size.
- Cohen's $d$ of 0.5 means the difference in means equals half a (pooled) standard deviation.
- Interestingly, Cohen's $d$ can be directly linked to the probability of superiority (the probability that a random individual sample in a population is greater than another random individual sample in another population).

### Pitfalls and limitations
- Sensitive to outliers - as a mean-based measure, Cohen's $d$ can be heavily influenced by extreme values in small samples.
- Unstable for small sample size - Cohen's $d$ tends to overestimate the effect size when calculated from small samples. This is because the sample standard deviation (SD) is typically lower than the population SD, leading to an overestimated $d$ value.
- A small Cohen's $d$ does not mean that the difference is not important (depending on the nature of the data) or statistically significant.

# Energy Distance (ED)

Energy Distance is a metric that measures the distance between the distributions of random vectors.

$$ED_\rho(P, Q) = 2\mathbb{E}[\rho(X, Y)] - \mathbb{E}[\rho(X, X')] - \mathbb{E}[\rho(Y, Y')]$$

where

$X, X'$ are independent and identically distributed random variables with distribution $P$

$Y, Y'$ are independent and identically distributed random variables with distribution $Q$

$P, Q$ are probability distributions on space $Z$

$\rho$ is a semimetric of negative type on $Z$

**Value range:** $[0, \infty)$
Higher values indicate dissimilarity, 0 indicates that the two distributions are identical.

## Use in METRIC-framework

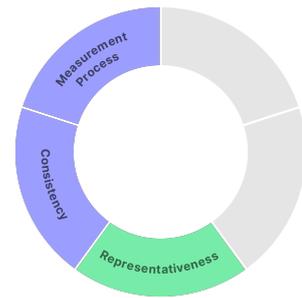

Variety
Target class balance
Homogeneity
Distribution drift

## References

(Székely and Rizzo, 2013)

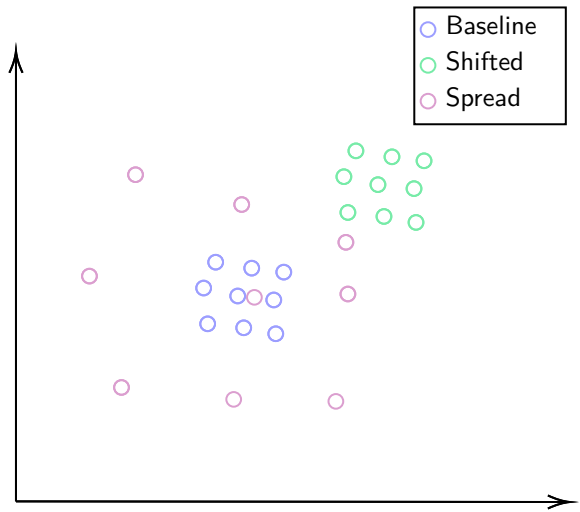

## Example

Used by Rahman et al., 2023, for evaluation of semantic segmentation models for medical imagery.

## Relation to other metrics

$$ED_\rho(P, Q) = 2 \cdot MMD_k^2(P, Q)$$

where:

$X, X' \sim P$ independently

$Y, Y' \sim Q$ independently

$\rho$ is a semimetric of negative type

$k$ generates $\rho$ via: $\rho(x, y) = k(x, x) + k(y, y) - 2k(x, y)$

## Applicability

**Modality:**

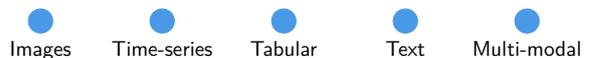

Images  Time-series  Tabular  Text  Multi-modal

**Variable:**

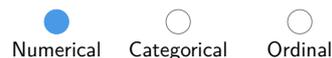

Numerical  Categorical  Ordinal

## Prerequisites and recommendations

- The distributions must have finite first moments with respect to the chosen semimetric.
- The samples drawn from each distribution must be independent.
- The data should be compatible with your chosen semimetric. For example, if using Euclidean distance, the data should be in a space where Euclidean distance is meaningful.

## Pitfalls and limitations

- Sensitive to outliers - Energy Distance uses expectations of distances, it can be sensitive to outliers, especially when using Euclidean distance.
- Sensitive to missing values.
- Higher computational costs for large datasets.

# Kullback-Leibler Divergence (KL)
## Synonyms: KL Divergence, Relative Entropy

An information-theoretic measure of the average difference between two probability distributions.

$$\text{Discrete:} \quad D_{KL}(P\|Q) = \sum_{x \in X} P(x) \log\left(\frac{P(x)}{Q(x)}\right)$$

$$\text{Continuous:} \quad D_{KL}(P\|Q) = \int_{-\infty}^{\infty} p(x) \log\left(\frac{p(x)}{q(x)}\right) dx$$

with $P$, $Q$ being the probability distributions and $p(x), q(x)$ being the probability density functions.

**Value Range:** $[0, +\infty) \downarrow$
KL Divergence is always a non-negative number. A value of 0 means both distributions are identical. Larger values indicate greater divergence between them.
The interpretation has to be considered in its context.

### Use in METRIC-framework

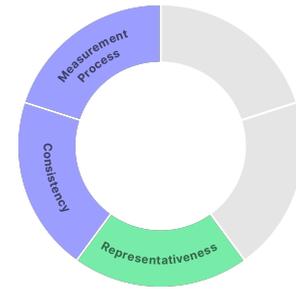

Variety
Target class balance
Homogeneity
Distribution drift

### References
(Joyce, 2011; Kullback and Leibler, 1951; Lin, 1991; Basterrech and Wozniak, 2022)

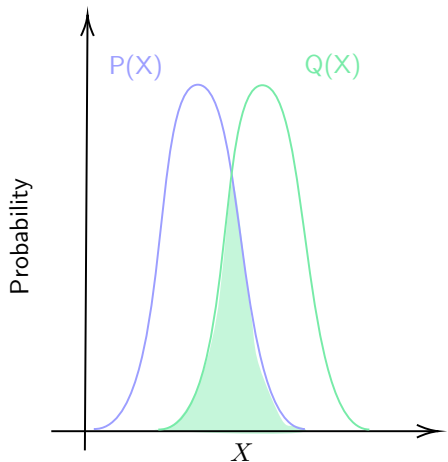

### Example
Kullback–Leibler divergence was used to quantify information loss between direct and indirect treatment effect estimates in network meta-analyses, allowing them to detect and interpret medical inconsistencies. (Spineli, 2024)
Practical implementation: *kl_div* function in scipy stats Python package

### Relation to other metrics
- Fisher Information
- Shannon Entropy
- Cross Entropy
- Mutual Information
- Jensen-Shannon divergence

### Applicability
**Modality:**
● Images   ● Time-series   ● Tabular   ● Text   ● Multi-modal

**Variable:**
● Numerical   ○ Categorical   ○ Ordinal

### Prerequisites and recommendations
- Input distributions $P$ and $Q$ have to be well-defined.
- KL Divergence is only defined where $Q(x) > 0$ if $P(x) > 0$.
- Normalizing both P and Q distributions before.
- Using smoothing techniques.
- Using a sufficient sample size.

### Pitfalls and limitations
- Sensitive to outliers or rare events.
- Dependence on parameter choice - Estimation of probability distributions needs to be done before computation.
- Sensitive to missing values.
- This measure is asymmetric, such that $D_{\mathrm{KL}}(P\|Q) \neq D_{\mathrm{KL}}(Q\|P)$.
- Does not satisfy the triangle inequality.
- As $D_{\mathrm{KL}}$ does not satisfies the triangle inequality and is asymmetric, it is not a true metric in the mathematical sense.
- KL divergence can result in infinite values if there are points where $P(x) > 0$ and $Q(x) = 0$. Careful handling of these cases is needed.

# Population Stability Index (PSI)
## Synonyms: Jeffreys divergence, symmetrized KL Divergence

Population Stability Index assesses the distinction between a given probability distribution $P$ and a reference distribution $Q$.

$$PSI = D_{\mathrm{KL}}(P \parallel Q) + D_{\mathrm{KL}}(Q \parallel P)$$

PSI can be understood as the symmetrized KL divergence.

**Value range:** $[0, \infty)$ ↑

If a variable shows exactly the same distribution in a validation sample as in the modeling sample, its PSI is equal to 0. The following rule of thumb indicates to what extent a variable has shifted in distribution:

- $< 0.1$: Very slight change.
- 0.1- 0.2: some minor change.
- $> 0.2$: Significant change.

### Use in METRIC-framework

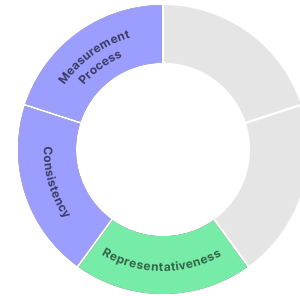

Variety
Target class balance
Homogeneity
Distribution drift

### References
(Kurian and Allali, 2024, Lin, 2017)

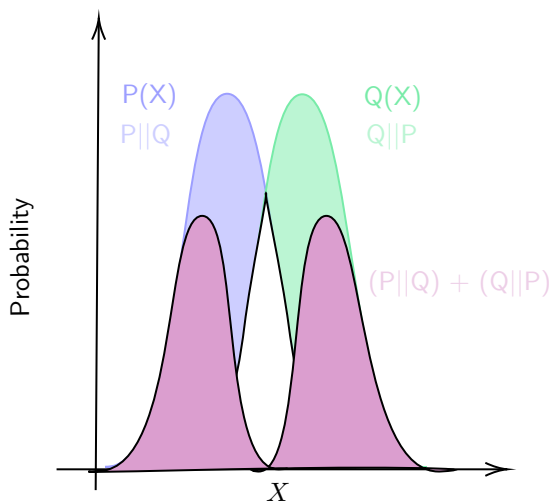

### Example
Dong et al., 2022, uses the PSI to evaluate homogeneity between test and training sets of HIV related data. Lu et al., 2025, use the PSI to measure distribution shifts in cancer data across time.

### Relation to other metrics
- KL Divergence
- Jensen–Shannon divergence

### Applicability
**Modality:**
○ Images  ○ Time-series  ● Tabular  ○ Text  ○ Multi-modal

**Variable:**
● Numerical  ● Categorical  ● Ordinal

### Prerequisites and recommendations
- This measure is symmetric, such that $PSI(P,Q) = PSI(Q,P)$.

### Pitfalls and limitations
- Dependence on parameter choice - Estimation of probability distributions needs to be done before computation.
- Sensitive to missing values.
- Does not satisfy the triangle inequality.
- KL divergence can result in infinite values if there are points where $P(x) > 0$ and $Q(x) = 0$. Careful handling of these cases is needed.

# Jensen Shannon Divergence

The Jensen-Shannon Divergence is a symmetric metric that measures the relative entropy or difference in information represented by two distributions $P$ and $Q$.

$$D_{\text{JS}} = \frac{1}{2} D_{\text{KL}}\left(P \,\|\, \frac{P+Q}{2}\right) + \frac{1}{2} D_{\text{KL}}\left(Q \,\|\, \frac{P+Q}{2}\right)$$

Where $D_{\text{KL}}$ is the Kullback-Leibler divergence. It can be thought of as measuring the distance between two data distributions or the total divergence to the average distribution.

**Value range:** $[0, \ln(2)]$ ↑
Higher values indicate dissimilarity, 0 indicate that the two distributions are identical.

## Use in METRIC-framework

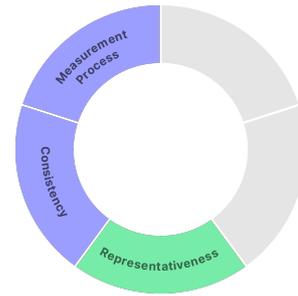

Variety
Target class balance
Homogeneity
Distribution drift

## References
(Nielsen, 2019)

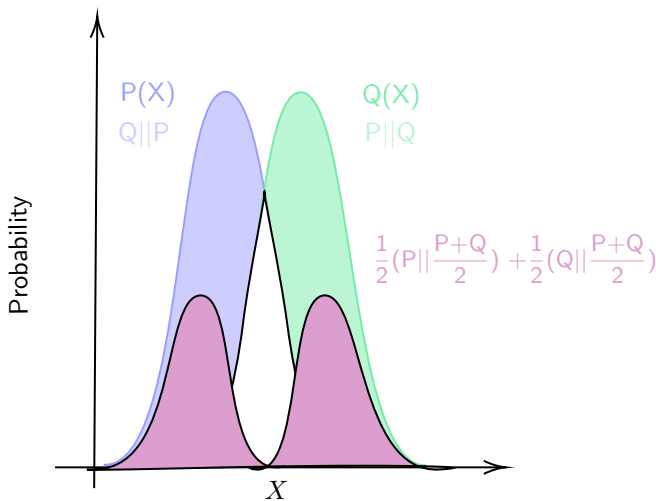

## Example
Used for evaluating LLM-generated summarizations of clinical notes (Kanval et al., 2022) and for analysis of deep brain stimulation data (Kaymak et al., 2025).

## Relation to other metrics
- KL Divergence
- PSI

## Applicability
**Modality:**
Images   Time-series   Tabular   Text   Multi-modal

**Variable:**
Numerical   Categorical   Ordinal

## Prerequisites and recommendations
- This measure is symmetric, such that $D_{\text{JS}}(P \,\|\, Q) = D_{\text{JS}}(Q \,\|\, P)$.
- Better suited for small sample sizes, as even rare events in either distribution produce do not $\infty$-values.

## Pitfalls and limitations
- Dependence on parameter choice - Estimation of probability distributions needs to be done before computation.
- Sensitive to missing values.
- Does not satisfy the triangle inequality.
- As $D_{\text{KL}}$ does not satisfies the triangle inequality, it is not a true metric in the mathematical sense.
- Does not capture very rare events as well as other metrics (like PSI).

# Kolmogorov-Smirnov Test Statistic

Kolmogorov-Smirnov (KS) Test is a nonparametric test which can be utilized to test whether two samples came from the same distribution.

$$D_{n,m} = \sup_x |F_n(x) - G_m(x)|$$

$F_n(x)$ and $G_m(x)$ being two cumulative distribution functions.
Its test statistic can also be used to quantify distribution dissimilarity.

**Value range:** $[0, 1]$ ↑
High values indicate dissimilarity; values close to 0 indicate close alignment.

## Use in METRIC-framework

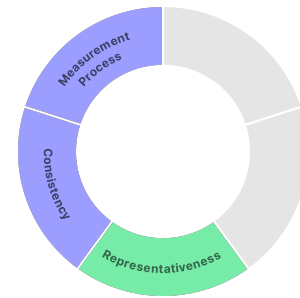

Variety
Target class balance
Homogeneity
Distribution drift

## References

(Freedman, 2007)

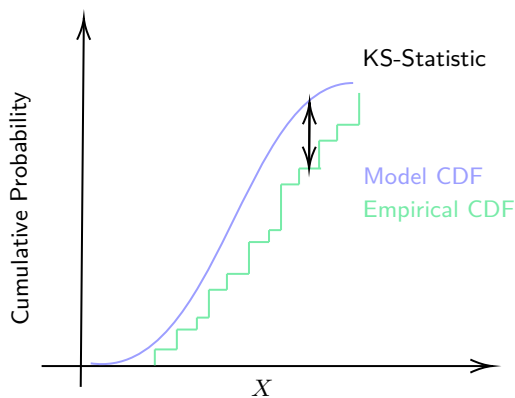

## Example

Closas et al., 2012, use the KS test on EHR data to detect Influenza epidemics.

## Relation to other metrics

- Anderson–Darling test statistic
- Cramér–von Mises criterion

## Applicability

**Modality:**
○ Images  ● Time-series  ● Tabular  ○ Text  ○ Multi-modal

**Variable:**
● Numerical  ○ Categorical  ○ Ordinal

## Prerequisites and recommendations

- The KS test statistic can be computed in one-sample or two-sample cases.
- It can be used to perform a statistical test.
- It is non parametric.
- The probability distributions should be continuous.

## Pitfalls and limitations

- Unstable for small sample size
- Sensitive to missing values
- This test typically has lower power compared to other parametric tests.

# Epps-Singleton test

Epps-Singleton test is a statistical test used primarily for comparing two sets of samples to determine if they come from the same probability distribution, without having to specify what that distribution is.

$$W^2 = \frac{nm}{n+m} \sum_{k=1}^{K} w_k |\phi_n(t_k) - \phi_m(t_k)|^2$$

$$\phi_n(t) = \frac{1}{n} \sum_{i=1}^{n} \exp(it \cdot X_i)$$

$$\phi_m(t) = \frac{1}{m} \sum_{j=1}^{m} \exp(it \cdot Y_j)$$

Where:

$n$ is the size of first sample

$m$ is the size of second sample

$K$ is the number of evaluation points (typically 4)

$t_k$ are the preselected evaluation points

$w_k$ are weights, inverse variances of $|\phi_n(t_k) - \phi_m(t_k)|^2$

$i$ is the imaginary unit

$X_i$ is the $i$th observation from first sample

$Y_j$ is the $j$th observation from second sample

Under null hypothesis: $W^2 \sim \chi^2(2K)$

**Value range:** $[0, 1) \uparrow$
The test statistic is non-negative because it's based on squared differences between characteristic functions. Theoretically, there is no upper bound.

### Use in METRIC-framework

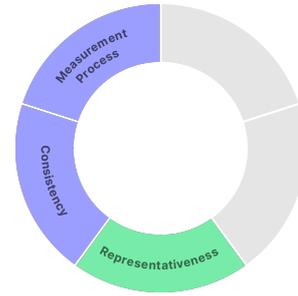

Variety
Target class balance
Homogeneity
Distribution drift

### References

(Epps and Singleton, 1986)

### Relation to other metrics

- t-test
- Mann-Whitney U-test
- Kolmogorov-Smirnov test

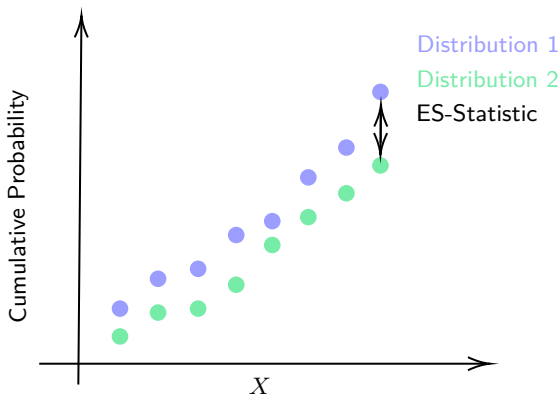

### Example

Used to evaluate the kernel density estimates in a study on the relationship of nurse staffing and mortality in hospitals (Mark et al., 2007).

### Applicability

**Modality:**
○ Images   ○ Time-series   ● Tabular   ○ Text   ○ Multi-modal

**Variable:**
● Numerical   ○ Categorical   ○ Ordinal

### Prerequisites and recommendations

- The original paper suggests using K=4 evaluation points, with values typically scaled to the data's standard deviation.
- The observations in each sample should be independent.
- It is non-parametric.

### Pitfalls and limitations

- Sensitive to missing values
- The test can be computationally demanding for very large datasets.
- The test is mathematically sophisticated and not widely implemented in standard statistical packages, leading to potential coding errors.

# K-Sample Anderson-Darling Test Statistic

The k-sample Anderson-Darling test is used to compare multiple samples, determining whether they come from the same underlying distribution.

$$A_{kN}^2 = \frac{1}{N} \sum_{j=1}^{k} \sum_{i=1}^{n_j} \frac{(N_{ij} - i)^2}{i(N - i)}$$

where $N = \sum_{j=1}^{k} n_j$ is the total sample size across all $k$ samples $n_j$ is the size of the $j$th sample
$N_{ij}$ is the number of observations in all samples combined that are less than or equal to the $i$th observation in the $j$th sample.

**Value range:** $[0, \infty)$ ↑
The test statistic equals 0 only when the distributions of all k samples are identical. As the differences between the distributions increase, the test statistic increases. No fixed upper limit, but practically bounded by sample size.

### Use in METRIC-framework

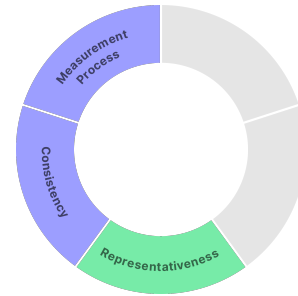

Variety
Target class balance
Homogeneity
Distribution drift

### References

(Scholz and Stephens, 1987, Engmann and Cousineau, 2011)

### Applicability

**Modality:**
○ Images  ○ Time-series  ● Tabular  ○ Text  ○ Multi-modal

**Variable:**
● Numerical  ○ Categorical  ○ Ordinal

### Example

Used to analyse the similarity in genetic markers in human bone marrow for immunotherapy (Zhang et. al, 2024) and to evaluate features extracted from histopathological data by a neural network (Hölscher et al., 2023).

### Relation to other metrics

- Kolmogorov-Smirnov test statistic
- Cramér–von Mises criterion

### Prerequisites and recommendations

- The k samples should be independent of each other. This means that observations in one sample should not influence or be related to observations in other samples.
- It is non parametric.

### Pitfalls and limitations

- Sensitive to missing values
- The probability distributions should be continuous.
- A significant result only indicates that differences exist somewhere, not which specific distributions differ or how.

# Chi-Squared Test Statistic

The Chi-Squared test assesses whether different groups have the same distribution across categories.

$$\chi^2 = \sum_{i=1}^{k} \frac{(O_i - E_i)^2}{E_i}$$

Where:
- $O_i$ is the observed frequency
- $E_i$ is the expected frequency
- $\sum$ means "sum over all categories"

The $\chi^2$ test statistic compares the frequencies of co-occurrence of two variables. It can also be used to quantify distribution dissimilarity: the higher the test statistic, the higher the discrepancy between two distributions.

**Value range:** $[0, \infty)$ ↓
A value of zero indicates that the proportions of each class are exactly as expected, and higher values indicate more deviation from the expected proportion as the test statistic increases.

## Use in METRIC-framework

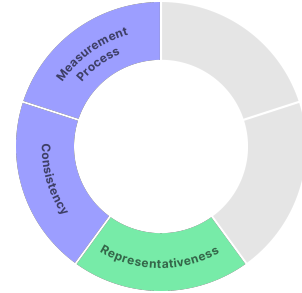

Variety
Target class balance
Homogeneity
Distribution drift

## References
(Freedman, 2007)

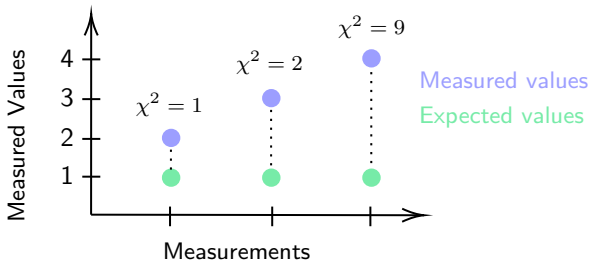

## Example
Homogeneity of binary variables (e.g., sex, smoking status, hypertension) within two dataset and across positive and negative bladder cancer status was assessed using the chi-square test. (Cabon, 2025)

## Relation to other metrics
- G-test statistic
- Pearson chi-square statistic

## Applicability
**Modality:** ○ Images  ○ Time-series  ● Tabular  ○ Text  ○ Multi-modal

**Variable:** ○ Numerical  ● Categorical  ● Ordinal

## Prerequisites and recommendations
- Observed and expected frequencies need to be computed beforehand.

## Pitfalls and limitations
- Unstable for small sample size
- Sensitive to missing values
- Only applicable to categorical data. Also works with ordinal data if the cardinality of datasets is not too important, but there the order relation between samples is not considered.
- Not applicable if the expected frequencies of any category are less than 5.

# Fréchet Inception Distance (FID)

Developed for evaluating the semantic quality of generated images. It compares the statistics of generated and reference images in feature space of an Inception v3 model, which is semantically richer than pixel space.

$$\text{FID} = \|\mu - \mu'\|_2^2 + \text{Tr}\left(\Sigma + \Sigma' - 2(\Sigma\Sigma')^{1/2}\right)$$

Where:

$$\begin{aligned}
\mu, \Sigma &: \text{mean and covariance of Inception features for real images} \\
\mu', \Sigma' &: \text{mean and covariance of Inception features} \\
&\quad \text{for generated images} \\
\text{Tr}(\cdot) &: \text{trace of a matrix} \\
(\Sigma\Sigma')^{1/2} &: \text{matrix square root of the product of covariance matrices} \\
\|\cdot\|_2 &: \text{Euclidean (L2) norm of the difference of the means}
\end{aligned}$$

**Value range:** $[0, \infty)$ ↑
Small values indicate similar distributions (0 when feature distribution are identical), an higher value indicates more dissimilarity between feature distributions.

## Use in METRIC-framework

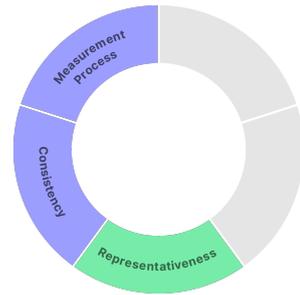

Variety
Target class balance
Homogeneity
Distribution drift

## References

(Heusel et al., 2017)

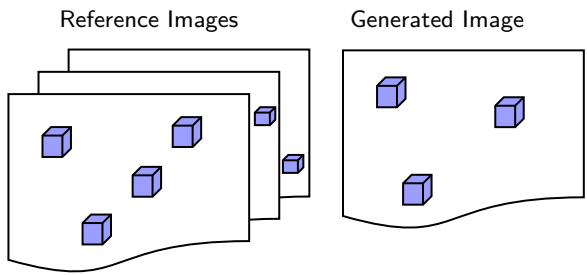

Comparisson of high-level features such as "blue boxes"

## Example

Used for the evaluation of generated synthetic chest radiographies, using also a X-ray specific model instead of the ImageNet-based Inception (Lee et al., 2024).

## Relation to other metrics

- Special case of Earth-movers distance for two normally distributed variables.

## Applicability

**Modality:**
● Images  ○ Time-series  ○ Tabular  ○ Text  ○ Multi-modal

**Variable:**
● Numerical  ○ Categorical  ○ Ordinal

## Prerequisites and recommendations

- Both compared datasets must be processed identically.
- All sample images need to be encoded via an Inception-v3 model.
- A sufficient number of samples should be processed to the underlying gaussian estimates are stable.

## Pitfalls and limitations

- Depends on parameter choice - Results depend on the pretrained Inception-v3 model. Different implementations or feature layers may yield incompatible FID values
- FID is best used for relative comparison between models or iterations. A "good" FID value depends on the data and task — it cannot be interpreted as an absolute measure of quality.
- FID assumes both distributions are multivariate Gaussians in the Inception feature space. This approximation is reasonable for natural images but may not hold for out-of-distribution content.

# Kernel Inception Distance (KID)

The Kernel Inception Distance measures the Maximum Mean Discrepancy (MMD) between the distributions of real and generated images in the feature space of a pre-trained Inception network.

$$\text{KID} = \text{MMD}^2(X, Y) =$$
$$\frac{1}{m(m-1)} \sum_{\substack{i,j=1 \\ i \neq j}}^{m} k(x_i, x_j) + \frac{1}{n(n-1)} \sum_{\substack{i,j=1 \\ i \neq j}}^{n} k(y_i, y_j)$$
$$- \frac{2}{mn} \sum_{i=1}^{m} \sum_{j=1}^{n} k(x_i, y_j),$$
$$X = \{x_1, x_2, \ldots, x_m\}, \quad Y = \{y_1, y_2, \ldots, y_n\},$$

$x_i, y_j \in \mathbb{R}^d$, (features from a pre-trained network, e.g., Inception),

$k(x, y) = \left( \frac{x^\top y}{d} + 1 \right)^3$, $d$ = dimensionality of the feature vectors.

**Value range:** $[0, 1]$ ↑
A value of 0 indicates perfect similarity between real and generated distributions. Larger values indicate greater dissimilarity between distributions.

## Use in METRIC-framework

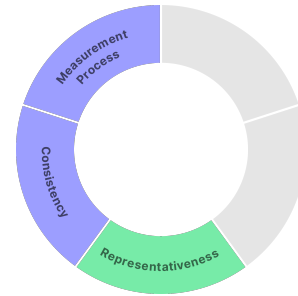

Variety
Target class balance
Homogeneity
Distribution drift

## References
(Binkowski et al., 2018)

## Relation to other metrics
- The Kernel Inception Distance is the squared Mean Maximum Discrepancy (MMD) between Inception representations.

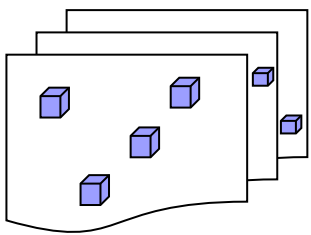 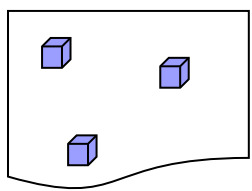

Comparisson of high-level features such as "blue boxes"

## Applicability

**Modality:**

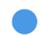 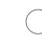 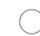 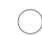 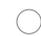
Images  Time-series  Tabular  Text  Multi-modal

**Variable:**

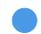 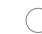 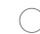
Numerical  Categorical  Ordinal

## Prerequisites and recommendations
- Use the same number of real and generated samples for more reliable comparisons.
- Access to a pre-trained Inception network (typically Inception-v3) to extract features..
- Use KID alongside other metrics like FID or Inception Score for more comprehensive evaluation.

## Pitfalls and limitations
- Depends on parameter choice - Results can vary based on the chosen kernel and its parameters, introducing a subjective element to the evaluation.
- KID relies on features extracted from the Inception network, which may not be optimal for all image domains or types.
- The absolute values of KID scores lack intuitive interpretation beyond comparative analysis.

# Mann-Whitney U-rank test
## Synonyms: Wilcoxon rank-sum test

The Mann-Whitney U test is used to determine whether two independent samples were selected from populations having the same distribution.

Given:

Sample 1: $X = \{X_1, X_2, \ldots, X_{n_1}\}$ with size $n_1$

Sample 2: $Y = \{Y_1, Y_2, \ldots, Y_{n_2}\}$ with size $n_2$

Step 1: Combine samples and rank all values from lowest to highest

Step 2: Calculate rank sums:

$R_1 = \sum (\text{ranks assigned to sample } X)$

$R_2 = \sum (\text{ranks assigned to sample } Y)$

Step 3: Calculate $U$ statistics:

$U_1 = n_1 n_2 + \dfrac{n_1(n_1+1)}{2} - R_1$

$U_2 = n_1 n_2 + \dfrac{n_2(n_2+1)}{2} - R_2$

Step 4: The test statistic is the minimum:

$U = \min(U_1, U_2)$

**Value range:** $[0, \infty) \uparrow$
The minimum value is 0, which occurs when all values in one sample are smaller than all values in the other sample. The maximum value is $n_1 n_2$, which occurs in the opposite extreme case.

### Use in METRIC-framework

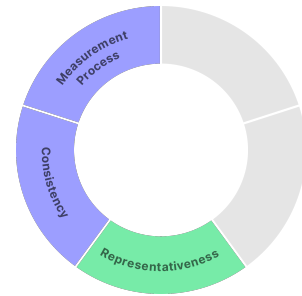

Variety
Target class balance
Homogeneity
Distribution drift

### References
(Mann and Whitney, 1947; Wilcoxon, 1945)

### Example
Used for evaluating variables which don't follow normal distributions for both diabetes risk prediction (Whang et al., 2023) and in prediction of chronic heart failure (Xu et al, 2025).

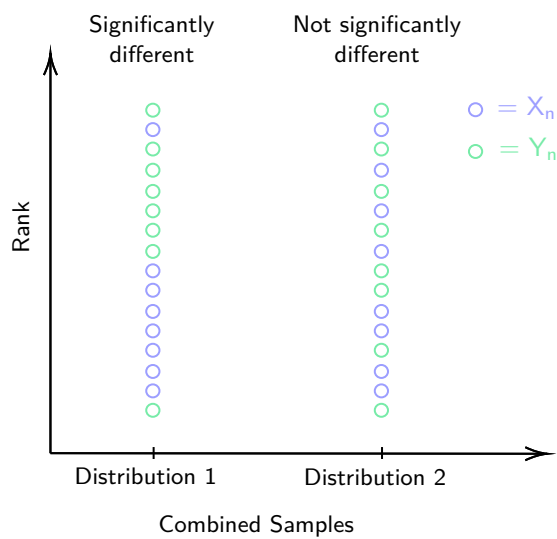

### Applicability

**Modality:** Images, Time-series, **Tabular**, Text, Multi-modal

**Variable:** **Numerical**, Categorical, **Ordinal**

### Prerequisites and recommendations
- Observations in one group must not be related to observations in the other group.
- Measurements must be at least on an ordinal scale.
- It is non parametric.

### Pitfalls and limitations
- When many tied ranks occur, standard calculations may yield inaccurate p-values without proper correction.
- The test doesn't specifically compare medians, but rather tests whether one population tends to produce larger values than the other.

# Wasserstein Distance
## Synonyms: Earth Mover's Distance

Models the distance between two probability distributions as the minimum cost of turning one into the other. It provides a robust way to compare distributions, particularly when their shapes and locations differ significantly.

$$W_p(P,Q) = \left( \inf_{\gamma \in \Gamma(P,Q)} \int_{\Omega \times \Omega} d(x,y)^p \, d\gamma(x,y) \right)^{1/p}$$

with $\Gamma(P,Q)$ as the set of all possible couplings (joint distributions) with marginals P and Q. Each coupling represents a possible way of pairing elements from P with elements from $Q$. $d(x,y)$ is the distance metric between points $x$ and $y$ (e.g., Euclidean distance). $p$ is a parameter defining the order of the Wasserstein distance.

**Value range:** $[0, \infty)$ ↑
Higher values indicate dissimilarity, 0 indicates that the two distributions are identical.

### Use in METRIC-framework

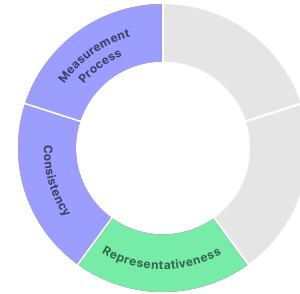

Variety
Target class balance
Homogeneity
Distribution drift

### References
(Arjovsky & Bottou, 2017; Cédric Villani, 2009; Panaretos & Zemel, 2018)

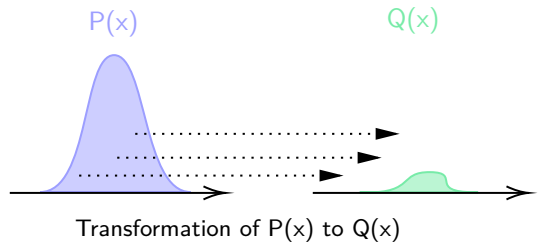

P(x)   Q(x)

Transformation of P(x) to Q(x)

### Relation to other metrics
- for $p=1$: 1-Wasserstein distance is known as the Earth Mover's distance.
- Variations which are (actually) computable include sliced Wasserstein Distance, generalized Wasserstein distance.

### Applicability
**Modality:**
● Images  ● Time-series  ● Tabular  ○ Text  ○ Multi-modal

**Variable:**
● Numerical  ○ Categorical  ○ Ordinal

### Prerequisites and recommendations
- For empirical use (e.g., on histograms or samples), make sure both distributions are normalized to valid probability distributions (sum to 1).

### Pitfalls and limitations
- The exact computation involves solving an optimal transport problem, which scales cubically with the number of samples. This limits its use in large-scale or real-time systems.
- Except for certain special cases (e.g., Gaussian distributions), there is no closed-form expression for Wasserstein distances.
- Computing Wasserstein distances exactly is computationally expensive. For large-scale problems, consider approximations (e.g., sliced Wasserstein).

## 9.7 Correlation coefficients

# Pearson's Correlation Coefficient (PCC)
## Synonyms: Pearson's R, Linear correlation coefficient

PCC is a descriptive and inferential statistical metric that measures the strength and direction of the linear relationship between two continuous variables, and is commonly used to assess inter-agreement or agreement in measurements when the relationship is expected to be linear.

$$p_{x,y} := \frac{cov(x,y)}{\sigma_x \sigma_y}$$

cov(x,y) is the covariance of the two variables, $\sigma$ the standard deviations.

**Value Range:** $[-1, 1]$ ↑
A value between 0 to +1 indicates a positive correlation (changing in the same direction), a value of 0 no correlation and a value between 0 to -1 indicates a negative correlation (changing in the opposite directions) between the two variables. For interpreting the PCC, the following references could be used: Chan, Y H., 2003; Akoglu, 2018.

### Use in METRIC-framework

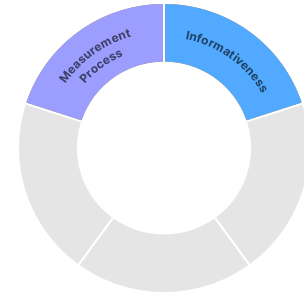

Accuracy
Noisy labels
Feature importance

### References
(Akoglu, 2018; Fahrmeir et al., 2016)

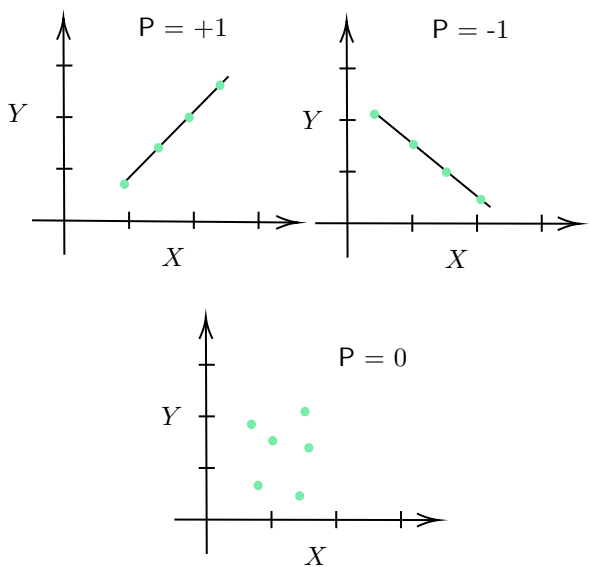

### Example
Assessing the correlation between systolic blood pressure measured via a cuffless (prototype) and a cuffed (gold standard) blood pressure device (ISO 81060-2:2018).

Often used to remove highly correlated features in the dataset (Prelaj et al.) In Stolarova, 2014. study assessing expressive vocabulary in toddlers, researchers used PCC as a measure of inter-rater agreement between standardized vocabulary scores assigned by parents and those assigned by daycare teachers.

Common in feature selection and data exploration in ML workflows (Hall, 1999; Nasir et al., 2020; Yu L. and Liu H., 2003).

Practical implementation: Build-in for dataframe of pandas in Python and *pearsonr* function in scipy stats Python package.

### Relation to other metrics
- Concordance Correlation Coefficient
- Spearman´s rank correlation coefficient
- Kendall´s rank correlation coefficient
- Mutual Information

### Applicability
**Modality:**
● Images   ● Time-series   ● Tabular   ○ Text   ○ Multi-modal

**Variable:**
● Numerical   ○ Categorical   ○ Ordinal

### Prerequisites and recommendations
- Both variables must be continuos, numerical and paired.
- The relation between the variables should be linear.
- Variables should be approximately normally distributed.
- Outliers should be identified and handled properly.
- Best used for inter-rater agreement of continuous scales.

### Pitfalls and limitations
- Sensitive to outliers that can significantly influence the results.
- Only measuring linear correlation between variables.
- In case of range restrictions, misleading results can be produced.
- Not causal — correlation does not imply causation.

# Concordance Correlation Coefficient (CCC)
## Synonyms: Lin´s Concordance Correlation Coefficient

A statistical metric for assessing the agreement between two measurement methods, accounting for both correlation and systematic differences, to evaluate reproducibility and inter-rater reliability.

$$\rho'_c = \frac{(2\rho\sigma_x\sigma_y)}{\sigma_x^2 + \sigma_y^2 + (\mu_x - \mu_y)^2}$$

with $\rho$ being the Pearson's correlation coefficient between two variables, $\mu$ the means and $\sigma^2$ the variances for the two measurements x and y.
A statistical metric for assessing the agreement between two methods to evaluate reproducibility and inter-rater reliability.

**Value Range:** $[-1, 1]$ ↑
A value of +1 indicates a perfect positive agreement, 0 no agreement and -1 a perfect negative agreement.
For interpreting different scales, the following references could be used:
Altman, 1990;
Landis & Koch, 1977.

### Use in METRIC-framework

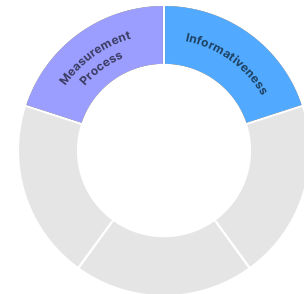

Accuracy
Noisy labels
Feature importance

### References
(Altman, 1990; Akoglu, 2018; Fahrmeir et al., 2016; Lin LI., 1989)

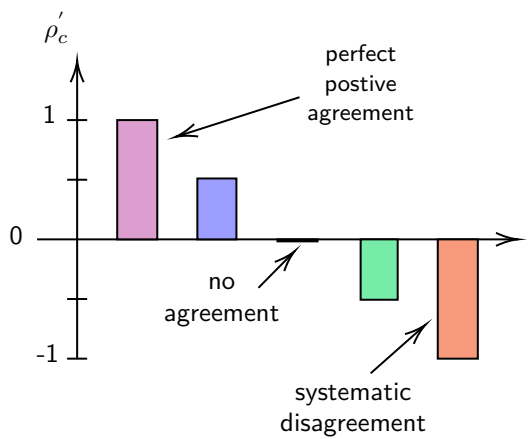

### Example
Rosenkrantz et al., 2013, CCC was used to assess inter-rater reproducibility of prostate MRI interpretations among radiologists, it provided a quantitative measure of exact agreement between radiologists scores across prostate regions. The main goal was to evaluate how consistently experienced and less experienced readers interpreted MRI findings.

### Relation to other metrics
- Cohen´s Kappa
- Fleiss´ Kappa
- Krippendorff´s Alpha
- Pearson correlation coefficient

### Applicability
**Modality:**
● Images  ● Time-series  ● Tabular  ○ Text  ○ Multi-modal

**Variable:**
● Numerical  ○ Categorical  ○ Ordinal

### Prerequisites and recommendations
- Variables must be continuous and on the same scale.
- The data should be approximately normally distributed.
- Pearson´s correlation coefficient should be calculated first.
- Best used for paired data with repeated measurements, e.g. two observers.

### Pitfalls and limitations
- Sensitive to outliers that can significantly influence the results.
- Not interpretable in isolation, needs comparisons.

# Goodman-Kruskal's gamma
## Synonyms: Kruskal Gamma, $\gamma$

A non-parametric measure of association between two ordinal variables, based on the difference between the number of concordant and discordant pairs, making no adjustment for ties.

$$G = \frac{N_c - N_d}{N_c + N_d}$$

$N_c$ refers to the number of concordant (i.e., both variables increase or decrease together) and $N_d$ to the number of reversed pairs (i.e., one variable increases while the other decreases).

**Value Range:** $[-1, 1]$ ↑
A value of $+1$ indicates a perfect positive association, 0 no association and a value of -1 a perfect negative association.

### Use in METRIC-framework

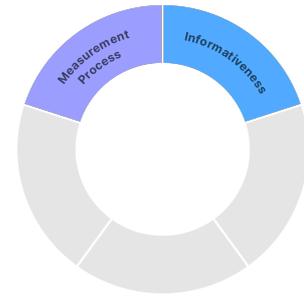

Accuracy
Noisy labels
Feature importance

### References
(Barbiero and Hitaj, 2020; Goodman and Kruskal, 1954)

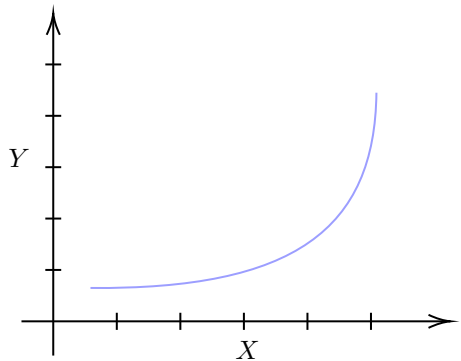

Positive monotic Spearman's $\rho = 1$

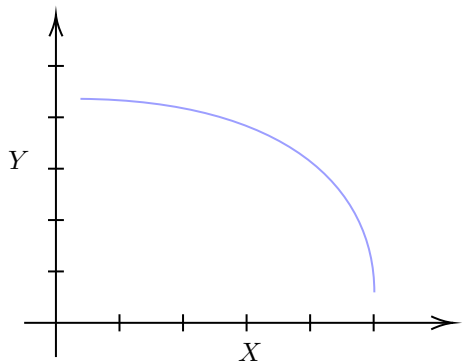

Negative monotic Spearman's $\rho = 1$

### Example
Practical implementation: GoodmanKruskal function in R package.

### Relation to other metrics
- Spearman´s rank correlation coefficient
- Kendall´s rank correlation coefficient

### Applicability
**Modality:**
● Images  ● Time-series  ● Tabular  ● Text  ○ Multi-modal

**Variable:**
○ Numerical  ● Categorical  ● Ordinal

### Prerequisites and recommendations
- Data should reflect a monotonic relationship.
- No ties or only non-critical ties.
- Suitable especially for a small to moderate dataset.

### Pitfalls and limitations
- Ignoring ties which can lead to overestimation of the association.
- Requires (manually) ordering if the data are not ordinal.
- Captures direction of association, not magnitude of differences.

# Kendall's rank correlation coefficient (Tau-b)
## Synonyms: Kendall's Tau, $\tau$

Kendall´s Tau is a non-parametric rank correlation coefficient that measures the strength and direction of association between two ordinal or continuous variables, based on concordant and discordant pairs, and explicitly accounts for ties.

$$\tau_b = \frac{n_c - n_d}{\sqrt{(n_c + n_d + t_x)(n_c + n_d + t_y)}}$$

$n_c$ is the number of concordant pairs, $n_d$ the number of discordant pairs, $t_x$ the number of tied values in x and $t_j$ the number of tied values in y.

**Value Range:** $[-1, 1]$ ↑
The values reflect the direction of the monotonic association: A value of +1 shows a perfect agreement in ranking, 0 no association and a value of -1 a perfect inverse ranking.

## Use in METRIC-framework

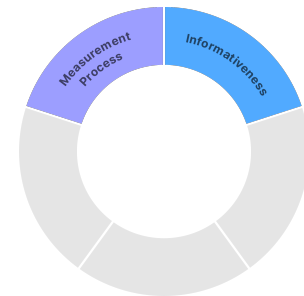

Accuracy
Noisy labels
Feature importance

## References
(Akoglu, 2018; Kendall, 1938)

## Example
Practical implementation: *kendalltau* function in scipy stats Python package

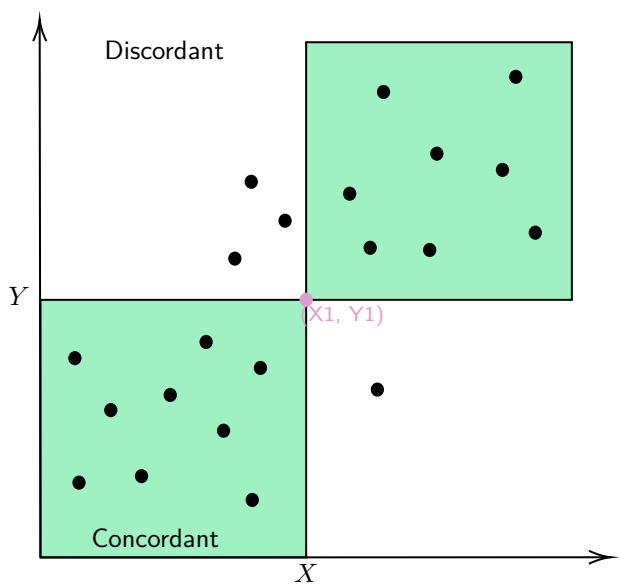

## Relation to other metrics
- Spearman´s rank correlation coefficient
- Concordance correlation coefficient (CCC)
- Goodman-Kruskal's gamma
- Pearson's correlation coefficient (Linear)

## Applicability
**Modality:** Images, Time-series, Tabular, Text (Multi-modal: no)

**Variable:** Numerical, Categorical, Ordinal

## Prerequisites and recommendations
- Variables have to be ordinal or ranked (numerical, categorical).
- Data should reflect a monotonic relationship.
- Robust estimator for outliers and non-normalities.
- Suitable especially for a small dataset.
- Handles ties better than Spearman´s rank correlation coefficient.

## Pitfalls and limitations
- Underestimating the associations in case of many tied ranks.
- Slower to compute for large datasets (pairwise comparisons).

# Spearman´s rank correlation coefficient
## Synonyms: Spearman´s rho, $\rho_s$

Spearman´s rank correlation coefficient is a non-parametric measure of the strength and direction of a monotonic relationship between two variables, based on their ranks rather than their actual values.

$$r_s = \rho(R[x], R[y]) = \frac{\text{cov}(R[x], R[y])}{\sigma_{R[x]} \sigma_{R[y]}}$$

with $\rho$ being the Pearson´s correlation coefficient applied to the rank variables $R[x]$ and $R[y]$, cov the covariance and $\sigma$ the standard deviations of the ranks.

**Value Range:** $[-1, 1]$ ↑
A high value, close to $+1$, indicates a similar rank between the two variables (increasing monotonic relationship). A low value close to -1 indicates a dissimilar rank (perfect decreasing monotonic relationship). 0 indicates no monotonic relationship.

### Use in METRIC-framework

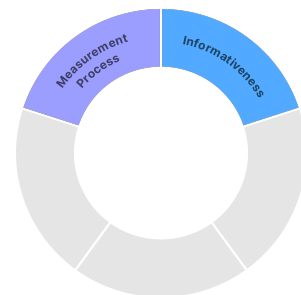

Accuracy
Noisy labels
Feature importance

### References
(Akoglu, 2018; Spearman, 1904)

### Example
Practical implementation: Build-in for dataframe of pandas in Python, *spearmanr* function in scipy stats Python package

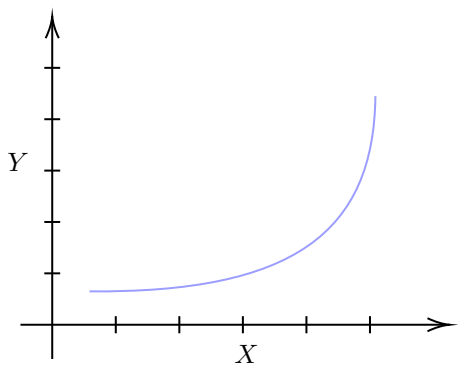

Positive monotic Spearman's $\rho = 1$

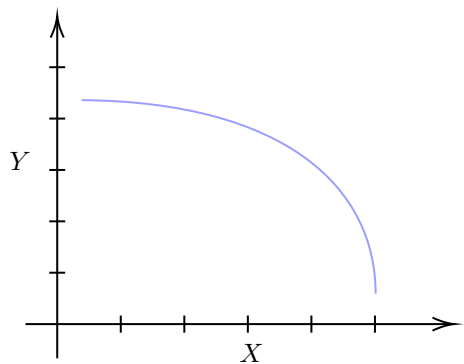

Negative monotic Spearman's $\rho = 1$

### Relation to other metrics
- Pearson´s correlation coefficient
- Concordance correlation coefficient
- Kendall´s rank correlation coefficient
- Goodman-Kruskal´s gamma

### Applicability
**Modality:**
● Images  ● Time-series  ● Tabular  ○ Text  ● Multi-modal

**Variable:**
● Numerical  ○ Categorical  ○ Ordinal

### Prerequisites and recommendations
- They have to be monotonic, can be non-linear and not normally distributed.
- The dataset can include outliers as ranks reduce the influence on the result.

### Pitfalls and limitations
- Unstable for small sample size
- Sensitive to outliers
- Sensitive to tied-ranks.
- Less interpretable for small sample size.
- Not suitable when scales are not comparable.

# Intraclass Correlation Coefficient (ICC)

ICC is a reliability index that quantifies the consistency or agreement of multiple raters or measurements, across repeated tests, or within the same rater. It reflects both the degree of correlation and agreement among measurements, and is commonly used to assess inter-rater, intra-rater, and test–retest reliability.

$$ICC = \frac{\text{variance between subjects}}{\text{total variance}}$$

**Value range:** $[0, 1]$ ↑
A high ICC (close to 1) indicates high precision with the measurements being consistent with minimal variability, whereas a low ICC (close to 0) indicates low precision with the measurements being inconsistent and having large variability

## Use in METRIC-framework

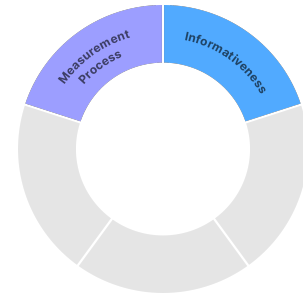

Accuracy
Noisy labels
Feature importance

## References
(Bland JM, 1986)

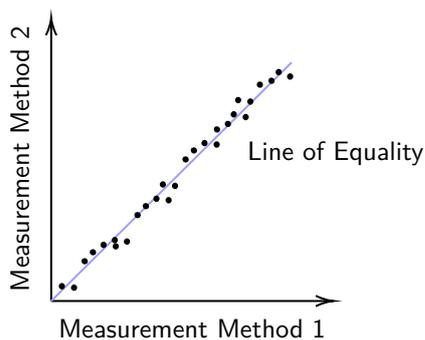

## Example
In Pleila, 2018, describing non-invasive biomarker measurements, ICC was used to quantify the consistency and reliability of repeated biomarker measurements across subjects, ensuring that observed variability reflects true individual differences rather than measurement error. Practical implementation: *intraclass_corr* function in Pingouin Python package

## Relation to other metrics

Pearson's Correlation Coefficient (PCC)
Concordance Correlation Coefficient (CCC)

## Applicability

**Modality:**
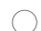 Images  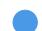 Time-series  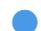 Tabular  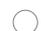 Text  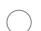 Multi-modal

**Variable:**
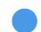 Numerical  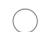 Categorical  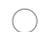 Ordinal

## Prerequisites and recommendations
- The right ICC model has to be chosen carefully.

## Pitfalls and limitations
- The interpretation of ICC values is dependent on the field of study/experimental design.
- A high correlation between the results must not mean that the two methods agree. As for example a change in the scale of measurement does not effect the correlation but affects the agreement.
- The ICC is not able to detect a bias.

# Cramer's V

Cramér's V is a statistical measure of association between two categorical variables. It quantifies the strength of dependence based on the chi-squared statistic and is independent of the table size, making it suitable for contingency tables of arbitrary dimensions.

$$V = \sqrt{\frac{\chi^2}{n \cdot \min(k-1, r-1)}}$$

where $\chi^2$ is the chi-squared statistic of the contingency table, $n$ is the total number of observations, $k$ is the number of columns, and $r$ is the number of rows.

**Value Range:** $[0, 1]$ ↑
A value of 0 indicates no association between the variables, while a value of 1 indicates a perfect association. Higher values correspond to stronger associations.

## Use in METRIC-framework

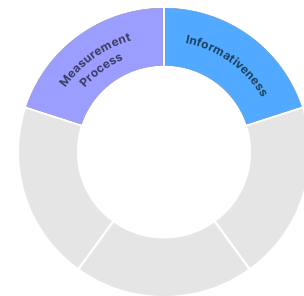

Accuracy
Noisy labels
Feature importance

## References

(Akoglu, 2018)

## Relation to other metrics

- Chi-squared statistic
- Pearson´s correlation coefficient
- Concordance correlation coefficient
- Kendall´s rank correlation coefficient
- Goodman-Kruskal´s gamma

## Applicability

**Modality:**

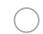 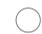 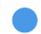 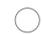 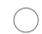
Images   Time-series   Tabular   Text   Multi-modal

**Variable:**

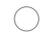 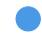 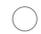
Numerical   Categorical   Ordinal

## Prerequisites and recommendations

- Observations should be independent.
- Expected cell counts should not be excessively small.
- Best interpreted together with contingency table inspection.

## Pitfalls and limitations

- Sensitive to sample size through the chi-squared statistic.
- Unstable for highly imbalanced contingency tables.
- Does not indicate direction or causality of association.



\end{document}

%% file: tables/metrics-table.tex
\centering
\resizebox{\linewidth}{!}{%
\begin{tabular}{!{\color{white!80!black}\vline}l|p{0.25cm}p{0.25cm}p{0.25cm}|p{0.25cm}p{0.25cm}p{0.25cm}|p{0.25cm}p{0.25cm}p{0.25cm}p{0.25cm}|p{0.25cm}|p{0.25cm}p{0.25cm}p{0.25cm}|p{0.25cm}|p{0.25cm}|}
\arrayrulecolor{white!80!black}\hline
 &
  \begin{sideways}Accuracy \end{sideways} &
  \begin{sideways}Noisy labels \end{sideways}&
  \begin{sideways}Completeness \end{sideways}&
  \begin{sideways}Syntactic consistency \end{sideways}&
  \begin{sideways}Homogeneity \end{sideways}&
  \begin{sideways}Distribution drift \end{sideways}&
  \begin{sideways}Dataset size \end{sideways}&
  \begin{sideways}Granularity \end{sideways}&
  \begin{sideways}Variety \end{sideways}&
  \begin{sideways}Target class balance \end{sideways}&
  \begin{sideways}Currency \end{sideways}&
  \begin{sideways}Uniqueness \end{sideways}&
  \begin{sideways}Informative missingness \end{sideways}&
  \begin{sideways}Feature importance \end{sideways}&
  \begin{sideways}Distribution metrics \end{sideways}&
  \begin{sideways}Correlation coefficients \end{sideways}\\ \arrayrulecolor{black}\hline
Entropy &
  \cellcolor[HTML]{9B9B9B} &
   &
   &
   &
   &
   &
   &
   &
   &
   &
   &
   &
   &
   &
   &
   \\  \arrayrulecolor{gray}\arrayrulecolor{white!80!black}\hline
Limit of Detection &
  \cellcolor[HTML]{9B9B9B} &
   &
   &
   &
   &
   &
   &
   &
   &
   &
   &
   &
   &
   &
   &
   \\ \arrayrulecolor{white!80!black}\hline
Limit of Quantification &
  \cellcolor[HTML]{9B9B9B} &
   &
   &
   &
   &
   &
   &
   &
   &
   &
   &
   &
   &
   &
   &
   \\ \arrayrulecolor{white!80!black}\hline
Systematic error in instruments &
  \cellcolor[HTML]{9B9B9B} &
   &
   &
   &
   &
   &
   &
   &
   &
   &
   &
   &
   &
   &
   &
   \\ \arrayrulecolor{white!80!black}\hline
Random error in instruments &
  \cellcolor[HTML]{9B9B9B} &
   &
   &
   &
   &
   &
   &
   &
   &
   &
   &
   &
   &
   &
   &
   \\ \arrayrulecolor{white!80!black}\hline
Bland-Altman coefficient of repeatability &
  \cellcolor[HTML]{9B9B9B} &
   &
   &
   &
   &
   &
   &
   &
   &
   &
   &
   &
   &
   &
   &
   \\ \arrayrulecolor{white!80!black}\hline
Repeatability coefficient of variation &
  \cellcolor[HTML]{9B9B9B} &
   &
   &
   &
   &
   &
   &
   &
   &
   &
   &
   &
   &
   &
   &
   \\ \arrayrulecolor{white!80!black}\hline
Reproducibility variance &
  \cellcolor[HTML]{9B9B9B} &
   &
   &
   &
   &
   &
   &
   &
   &
   &
   &
   &
   &
   &
   &
   \\ \arrayrulecolor{white!80!black}\hline
Cohen's kappa &
   &
  \cellcolor[HTML]{9B9B9B} &
   &
   &
   &
   &
   &
   &
   &
   &
   &
   &
   &
   &
   &
   \\ \arrayrulecolor{white!80!black}\hline
Fleiss' kappa &
   &
  \cellcolor[HTML]{9B9B9B} &
   &
   &
   &
   &
   &
   &
   &
   &
   &
   &
   &
   &
   &
   \\ \arrayrulecolor{white!80!black}\hline
Kendall's coefficient of concordance &
   &
  \cellcolor[HTML]{9B9B9B} &
   &
   &
   &
   &
   &
   &
   &
   &
   &
   &
   &
   &
   &
   \\ \arrayrulecolor{white!80!black}\hline
Krippendorff's alpha &
   &
  \cellcolor[HTML]{9B9B9B} &
   &
   &
   &
   &
   &
   &
   &
   &
   &
   &
   &
   &
   &
   \\ \arrayrulecolor{white!80!black}\hline
Dice similarity score &
   &
  \cellcolor[HTML]{9B9B9B} &
   &
   &
   &
   &
   &
   &
   &
   &
   &
   &
   &
   &
   &
   \\ \arrayrulecolor{white!80!black}\hline
Intersection over Union &
   &
  \cellcolor[HTML]{9B9B9B} &
   &
   &
   &
   &
   &
   &
   &
   &
   &
   &
   &
   &
   &
   \\ \arrayrulecolor{white!80!black}\hline
Completeness &
   &
   &
  \cellcolor[HTML]{9B9B9B} &
   &
   &
   &
   &
   &
   &
   &
   &
   &
   &
   &
   &
   \\ \arrayrulecolor{white!80!black}\hline
Patient-level completeness &
   &
   &
  \cellcolor[HTML]{9B9B9B} &
   &
   &
   &
   &
   &
   &
   &
   &
   &
   &
   &
   &
   \\ \arrayrulecolor{white!80!black}\hline
Record completeness &
   &
   &
  \cellcolor[HTML]{9B9B9B} &
   &
   &
   &
   &
   &
   &
   &
   &
   &
   &
   &
   &
   \\ \arrayrulecolor{white!80!black}\hline
Syntactic accuracy &
   &
   &
   &
  \cellcolor[HTML]{9B9B9B} &
   &
   &
   &
   &
   &
   &
   &
   &
   &
   &
   &
   \\ \arrayrulecolor{white!80!black}\hline
Page-Hinkley test statistic &
   &
   &
   &
   &
   &
  \cellcolor[HTML]{9B9B9B} &
   &
   &
   &
   &
   &
   &
   &
   &
   &
   \\ \arrayrulecolor{white!80!black}\hline
Dataset size &
   &
   &
   &
   &
   &
   &
  \cellcolor[HTML]{9B9B9B} &
   &
   &
   &
   &
   &
   &
   &
   &
   \\ \arrayrulecolor{white!80!black}\hline
Granularity &
   &
   &
   &
   &
   &
   &
   &
  \cellcolor[HTML]{9B9B9B} &
   &
   &
   &
   &
   &
   &
   &
   \\ \arrayrulecolor{white!80!black}\hline
Sampling frequency &
   &
   &
   &
   &
   &
   &
   &
  \cellcolor[HTML]{9B9B9B} &
   &
   &
   &
   &
   &
   &
   &
   \\ \arrayrulecolor{white!80!black}\hline
Resolution &
   &
   &
   &
   &
   &
   &
   &
  \cellcolor[HTML]{9B9B9B} &
   &
   &
   &
   &
   &
   &
   &
   \\ \arrayrulecolor{white!80!black}\hline
Label granularity &
   &
   &
   &
   &
   &
   &
   &
  \cellcolor[HTML]{9B9B9B} &
   &
   &
   &
   &
   &
   &
   &
   \\ \arrayrulecolor{white!80!black}\hline
Generalized imbalance ratio &
   &
   &
   &
   &
   &
   &
   &
   &
   &
  \cellcolor[HTML]{9B9B9B} &
   &
   &
   &
   &
   &
   \\ \arrayrulecolor{white!80!black}\hline
Imbalance degree &
   &
   &
   &
   &
   &
   &
   &
   &
   &
  \cellcolor[HTML]{9B9B9B} &
   &
   &
   &
   &
   &
   \\ \arrayrulecolor{white!80!black}\hline
Likelihood ratio imbalance degree &
   &
   &
   &
   &
   &
   &
   &
   &
   &
  \cellcolor[HTML]{9B9B9B} &
   &
   &
   &
   &
   &
   \\ \arrayrulecolor{white!80!black}\hline
Currency Ballou et al. &
   &
   &
   &
   &
   &
   &
   &
   &
   &
   &
  \cellcolor[HTML]{9B9B9B} &
   &
   &
   &
   &
   \\ \arrayrulecolor{white!80!black}\hline
Currency Li et al. &
   &
   &
   &
   &
   &
   &
   &
   &
   &
   &
  \cellcolor[HTML]{9B9B9B} &
   &
   &
   &
   &
   \\ \arrayrulecolor{white!80!black}\hline
Currency Hinrichs &
   &
   &
   &
   &
   &
   &
   &
   &
   &
   &
  \cellcolor[HTML]{9B9B9B} &
   &
   &
   &
   &
   \\ \arrayrulecolor{white!80!black}\hline
Currency Heinrich et al. &
   &
   &
   &
   &
   &
   &
   &
   &
   &
   &
  \cellcolor[HTML]{9B9B9B} &
   &
   &
   &
   &
   \\ \arrayrulecolor{white!80!black}\hline
Prevalence of duplicates &
   &
   &
   &
   &
   &
   &
   &
   &
   &
   &
   &
  \cellcolor[HTML]{9B9B9B} &
   &
   &
   &
   \\ \arrayrulecolor{white!80!black}\hline
Effective sample size &
   &
   &
   &
   &
   &
   &
   &
   &
   &
   &
   &
  \cellcolor[HTML]{9B9B9B} &
   &
   &
   &
   \\ \arrayrulecolor{white!80!black}\hline
Little's test &
   &
   &
   &
   &
   &
   &
   &
   &
   &
   &
   &
   &
  \cellcolor[HTML]{9B9B9B} &
   &
   &
   \\ \arrayrulecolor{white!80!black}\hline
Likelihood model for informative dropout &
   &
   &
   &
   &
   &
   &
   &
   &
   &
   &
   &
   &
  \cellcolor[HTML]{9B9B9B} &
   &
   &
   \\ \arrayrulecolor{white!80!black}\hline
Range &
  \cellcolor[HTML]{9B9B9B} &
  \cellcolor[HTML]{9B9B9B} &
   &
   &
  \cellcolor[HTML]{9B9B9B} &
  \cellcolor[HTML]{9B9B9B} &
   &
   &
  \cellcolor[HTML]{9B9B9B} &
  \cellcolor[HTML]{9B9B9B} &
   &
   &
   &
   &
  \cellcolor[HTML]{9B9B9B} &
   \\ \arrayrulecolor{white!80!black}\hline
Interquartile range &
  \cellcolor[HTML]{9B9B9B} &
  \cellcolor[HTML]{9B9B9B} &
   &
   &
  \cellcolor[HTML]{9B9B9B} &
  \cellcolor[HTML]{9B9B9B} &
   &
   &
  \cellcolor[HTML]{9B9B9B} &
  \cellcolor[HTML]{9B9B9B} &
   &
   &
   &
   &
  \cellcolor[HTML]{9B9B9B} &
   \\ \arrayrulecolor{white!80!black}\hline
Mean and standard deviation &
  \cellcolor[HTML]{9B9B9B} &
  \cellcolor[HTML]{9B9B9B} &
   &
   &
  \cellcolor[HTML]{9B9B9B} &
  \cellcolor[HTML]{9B9B9B} &
   &
   &
  \cellcolor[HTML]{9B9B9B} &
  \cellcolor[HTML]{9B9B9B} &
   &
   &
   &
   &
  \cellcolor[HTML]{9B9B9B} &
   \\ \arrayrulecolor{white!80!black}\hline
Hill numbers &
  \cellcolor[HTML]{9B9B9B} &
  \cellcolor[HTML]{9B9B9B} &
   &
   &
  \cellcolor[HTML]{9B9B9B} &
  \cellcolor[HTML]{9B9B9B} &
   &
   &
  \cellcolor[HTML]{9B9B9B} &
  \cellcolor[HTML]{9B9B9B} &
   &
   &
   &
   &
  \cellcolor[HTML]{9B9B9B} &
   \\ \arrayrulecolor{white!80!black}\hline
Maximum mean discrepancy &
  \cellcolor[HTML]{9B9B9B} &
  \cellcolor[HTML]{9B9B9B} &
   &
   &
  \cellcolor[HTML]{9B9B9B} &
  \cellcolor[HTML]{9B9B9B} &
   &
   &
  \cellcolor[HTML]{9B9B9B} &
  \cellcolor[HTML]{9B9B9B} &
   &
   &
   &
   &
  \cellcolor[HTML]{9B9B9B} &
   \\ \arrayrulecolor{white!80!black}\hline
Cohen's d &
  \cellcolor[HTML]{9B9B9B} &
  \cellcolor[HTML]{9B9B9B} &
   &
   &
  \cellcolor[HTML]{9B9B9B} &
  \cellcolor[HTML]{9B9B9B} &
   &
   &
  \cellcolor[HTML]{9B9B9B} &
  \cellcolor[HTML]{9B9B9B} &
   &
   &
   &
   &
  \cellcolor[HTML]{9B9B9B} &
   \\ \arrayrulecolor{white!80!black}\hline
Energy distance &
  \cellcolor[HTML]{9B9B9B} &
  \cellcolor[HTML]{9B9B9B} &
   &
   &
  \cellcolor[HTML]{9B9B9B} &
  \cellcolor[HTML]{9B9B9B} &
   &
   &
  \cellcolor[HTML]{9B9B9B} &
  \cellcolor[HTML]{9B9B9B} &
   &
   &
   &
   &
  \cellcolor[HTML]{9B9B9B} &
   \\ \arrayrulecolor{white!80!black}\hline
Kulback-Leibler divergence &
  \cellcolor[HTML]{9B9B9B} &
  \cellcolor[HTML]{9B9B9B} &
   &
   &
  \cellcolor[HTML]{9B9B9B} &
  \cellcolor[HTML]{9B9B9B} &
   &
   &
  \cellcolor[HTML]{9B9B9B} &
  \cellcolor[HTML]{9B9B9B} &
   &
   &
   &
   &
  \cellcolor[HTML]{9B9B9B} &
   \\ \arrayrulecolor{white!80!black}\hline
Population stability index &
  \cellcolor[HTML]{9B9B9B} &
  \cellcolor[HTML]{9B9B9B} &
   &
   &
  \cellcolor[HTML]{9B9B9B} &
  \cellcolor[HTML]{9B9B9B} &
   &
   &
  \cellcolor[HTML]{9B9B9B} &
  \cellcolor[HTML]{9B9B9B} &
   &
   &
   &
   &
  \cellcolor[HTML]{9B9B9B} &
   \\ \arrayrulecolor{white!80!black}\hline
Jensen Shannon divergence &
  \cellcolor[HTML]{9B9B9B} &
  \cellcolor[HTML]{9B9B9B} &
   &
   &
  \cellcolor[HTML]{9B9B9B} &
  \cellcolor[HTML]{9B9B9B} &
   &
   &
  \cellcolor[HTML]{9B9B9B} &
  \cellcolor[HTML]{9B9B9B} &
   &
   &
   &
   &
  \cellcolor[HTML]{9B9B9B} &
   \\ \arrayrulecolor{white!80!black}\hline
Kolmogorov-Smirnov test statistic &
  \cellcolor[HTML]{9B9B9B} &
  \cellcolor[HTML]{9B9B9B} &
   &
   &
  \cellcolor[HTML]{9B9B9B} &
  \cellcolor[HTML]{9B9B9B} &
   &
   &
  \cellcolor[HTML]{9B9B9B} &
  \cellcolor[HTML]{9B9B9B} &
   &
   &
   &
   &
  \cellcolor[HTML]{9B9B9B} &
   \\ \arrayrulecolor{white!80!black}\hline
Epps-Singelton test &
  \cellcolor[HTML]{9B9B9B} &
  \cellcolor[HTML]{9B9B9B} &
   &
   &
  \cellcolor[HTML]{9B9B9B} &
  \cellcolor[HTML]{9B9B9B} &
   &
   &
  \cellcolor[HTML]{9B9B9B} &
  \cellcolor[HTML]{9B9B9B} &
   &
   &
   &
   &
  \cellcolor[HTML]{9B9B9B} &
   \\ \arrayrulecolor{white!80!black}\hline
K-Sample Anderson-Darling test statistic &
  \cellcolor[HTML]{9B9B9B} &
  \cellcolor[HTML]{9B9B9B} &
   &
   &
  \cellcolor[HTML]{9B9B9B} &
  \cellcolor[HTML]{9B9B9B} &
   &
   &
  \cellcolor[HTML]{9B9B9B} &
  \cellcolor[HTML]{9B9B9B} &
   &
   &
   &
   &
  \cellcolor[HTML]{9B9B9B} &
   \\ \arrayrulecolor{white!80!black}\hline
Chi-squared test statistic &
  \cellcolor[HTML]{9B9B9B} &
  \cellcolor[HTML]{9B9B9B} &
   &
   &
  \cellcolor[HTML]{9B9B9B} &
  \cellcolor[HTML]{9B9B9B} &
   &
   &
  \cellcolor[HTML]{9B9B9B} &
  \cellcolor[HTML]{9B9B9B} &
   &
   &
   &
   &
  \cellcolor[HTML]{9B9B9B} &
   \\ \arrayrulecolor{white!80!black}\hline
Fréchet inception distance &
  \cellcolor[HTML]{9B9B9B} &
  \cellcolor[HTML]{9B9B9B} &
   &
   &
  \cellcolor[HTML]{9B9B9B} &
  \cellcolor[HTML]{9B9B9B} &
   &
   &
  \cellcolor[HTML]{9B9B9B} &
  \cellcolor[HTML]{9B9B9B} &
   &
   &
   &
   &
  \cellcolor[HTML]{9B9B9B} &
   \\ \arrayrulecolor{white!80!black}\hline
Kernel inception distance &
  \cellcolor[HTML]{9B9B9B} &
  \cellcolor[HTML]{9B9B9B} &
   &
   &
  \cellcolor[HTML]{9B9B9B} &
  \cellcolor[HTML]{9B9B9B} &
   &
   &
  \cellcolor[HTML]{9B9B9B} &
  \cellcolor[HTML]{9B9B9B} &
   &
   &
   &
   &
  \cellcolor[HTML]{9B9B9B} &
   \\ \arrayrulecolor{white!80!black}\hline
Mann-Whitney U-rank test &
  \cellcolor[HTML]{9B9B9B} &
  \cellcolor[HTML]{9B9B9B} &
   &
   &
  \cellcolor[HTML]{9B9B9B} &
  \cellcolor[HTML]{9B9B9B} &
   &
   &
  \cellcolor[HTML]{9B9B9B} &
  \cellcolor[HTML]{9B9B9B} &
   &
   &
   &
   &
  \cellcolor[HTML]{9B9B9B} &
   \\ \arrayrulecolor{white!80!black}\hline
Wasserstein distance &
  \cellcolor[HTML]{9B9B9B} &
  \cellcolor[HTML]{9B9B9B} &
   &
   &
  \cellcolor[HTML]{9B9B9B} &
  \cellcolor[HTML]{9B9B9B} &
   &
   &
  \cellcolor[HTML]{9B9B9B} &
  \cellcolor[HTML]{9B9B9B} &
   &
   &
   &
   &
  \cellcolor[HTML]{9B9B9B} &
   \\ \arrayrulecolor{white!80!black}\hline
Pearson's correlation coefficient &
  \cellcolor[HTML]{9B9B9B} &
  \cellcolor[HTML]{9B9B9B} &
   &
   &
   &
   &
   &
   &
   &
   &
   &
   &
   &
  \cellcolor[HTML]{9B9B9B} &
   &
  \cellcolor[HTML]{9B9B9B} \\ \arrayrulecolor{white!80!black}\hline
Concordance correlation coefficient &
  \cellcolor[HTML]{9B9B9B} &
  \cellcolor[HTML]{9B9B9B} &
   &
   &
   &
   &
   &
   &
   &
   &
   &
   &
   &
  \cellcolor[HTML]{9B9B9B} &
   &
  \cellcolor[HTML]{9B9B9B} \\ \arrayrulecolor{white!80!black}\hline
Goodman-Kruskal's gamma &
  \cellcolor[HTML]{9B9B9B} &
  \cellcolor[HTML]{9B9B9B} &
   &
   &
   &
   &
   &
   &
   &
   &
   &
   &
   &
  \cellcolor[HTML]{9B9B9B} &
   &
  \cellcolor[HTML]{9B9B9B} \\ \arrayrulecolor{white!80!black}\hline
Kendall's rank correlation coefficient &
  \cellcolor[HTML]{9B9B9B} &
  \cellcolor[HTML]{9B9B9B} &
   &
   &
   &
   &
   &
   &
   &
   &
   &
   &
   &
  \cellcolor[HTML]{9B9B9B} &
   &
  \cellcolor[HTML]{9B9B9B} \\ \arrayrulecolor{white!80!black}\hline
Spearman's rank correlation coefficient &
  \cellcolor[HTML]{9B9B9B} &
  \cellcolor[HTML]{9B9B9B} &
   &
   &
   &
   &
   &
   &
   &
   &
   &
   &
   &
  \cellcolor[HTML]{9B9B9B} &
   &
  \cellcolor[HTML]{9B9B9B} \\ \arrayrulecolor{white!80!black}\hline
Intraclass correlation coefficient &
  \cellcolor[HTML]{9B9B9B} &
  \cellcolor[HTML]{9B9B9B} &
   &
   &
   &
   &
   &
   &
   &
   &
   &
   &
   &
  \cellcolor[HTML]{9B9B9B} &
   &
  \cellcolor[HTML]{9B9B9B} \\ \arrayrulecolor{white!80!black}\hline
Cramer's V &
  \cellcolor[HTML]{9B9B9B} &
  \cellcolor[HTML]{9B9B9B} &
   &
   &
   &
   &
   &
   &
   &
   &
   &
   &
   &
  \cellcolor[HTML]{9B9B9B} &
   &
  \cellcolor[HTML]{9B9B9B} \\ \arrayrulecolor{black}\hline
\end{tabular}%
}

%% file: decision-trees/accuracy.tex
\tikzset{every picture/.style={line width=0.75pt}} 

\begin{tikzpicture}[x=0.75pt,y=0.75pt,yscale=-1,xscale=1]

\draw  [color={rgb, 255:red, 116; green, 232; blue, 164 }  ,draw opacity=1 ][fill={rgb, 255:red, 173; green, 243; blue, 203 }  ,fill opacity=1 ] (17.17,10.6) .. controls (17.17,8.39) and (18.96,6.6) .. (21.17,6.6) -- (473.17,6.6) .. controls (475.38,6.6) and (477.17,8.39) .. (477.17,10.6) -- (477.17,22.6) .. controls (477.17,24.81) and (475.38,26.6) .. (473.17,26.6) -- (21.17,26.6) .. controls (18.96,26.6) and (17.17,24.81) .. (17.17,22.6) -- cycle ;
\draw  [color={rgb, 255:red, 156; green, 158; blue, 255 }  ,draw opacity=1 ][fill={rgb, 255:red, 196; green, 197; blue, 255 }  ,fill opacity=1 ] (137.9,37.37) .. controls (137.9,35.17) and (139.69,33.37) .. (141.9,33.37) -- (358.9,33.37) .. controls (361.11,33.37) and (362.9,35.17) .. (362.9,37.37) -- (362.9,49.37) .. controls (362.9,51.58) and (361.11,53.37) .. (358.9,53.37) -- (141.9,53.37) .. controls (139.69,53.37) and (137.9,51.58) .. (137.9,49.37) -- cycle ;
\draw    (400.36,78.3) -- (400.36,83.3) ;
\draw    (400.36,104.3) -- (400.36,183.2) ;
\draw    (100.5,183.87) -- (409.5,183.87) ;
\draw    (100.5,183.87) -- (100.5,188.87) ;
\draw    (409.5,183.87) -- (409.5,188.87) ;
\draw    (250.36,184.45) -- (250.36,189.45) ;
\draw    (100.5,58.88) -- (400.5,58.88) ;
\draw    (100.5,58.88) -- (100.5,63.88) ;
\draw    (400.5,58.88) -- (400.5,63.88) ;
\draw    (250.36,53.45) -- (250.36,58.45) ;

\draw  [color={rgb, 255:red, 156; green, 158; blue, 255 }  ,draw opacity=1 ][fill={rgb, 255:red, 196; green, 197; blue, 255 }  ,fill opacity=1 ] (8.6,87.19) .. controls (8.6,84.96) and (10.41,83.15) .. (12.65,83.15) -- (229.55,83.15) .. controls (231.79,83.15) and (233.6,84.96) .. (233.6,87.19) -- (233.6,99.33) .. controls (233.6,101.56) and (231.79,103.37) .. (229.55,103.37) -- (12.65,103.37) .. controls (10.41,103.37) and (8.6,101.56) .. (8.6,99.33) -- cycle ;
\draw  [color={rgb, 255:red, 156; green, 158; blue, 255 }  ,draw opacity=1 ][fill={rgb, 255:red, 196; green, 197; blue, 255 }  ,fill opacity=1 ] (267.5,88.19) .. controls (267.5,85.96) and (269.31,84.15) .. (271.55,84.15) -- (488.45,84.15) .. controls (490.69,84.15) and (492.5,85.96) .. (492.5,88.19) -- (492.5,100.33) .. controls (492.5,102.56) and (490.69,104.37) .. (488.45,104.37) -- (271.55,104.37) .. controls (269.31,104.37) and (267.5,102.56) .. (267.5,100.33) -- cycle ;
\draw    (100.36,77.3) -- (100.36,82.3) ;
\draw    (68.5,108.88) -- (195.5,108.88) ;
\draw    (68.5,108.88) -- (68.5,113.88) ;
\draw    (195.5,108.88) -- (195.5,113.88) ;
\draw    (100.36,103.45) -- (100.36,108.45) ;
\draw  [color={rgb, 255:red, 81; green, 170; blue, 255 }  ,draw opacity=1 ][fill={rgb, 255:red, 150; green, 204; blue, 255 }  ,fill opacity=1 ] (133,129.15) -- (253.4,129.15) -- (253.4,143.5) -- (133,143.5) -- cycle ;
\draw  [color={rgb, 255:red, 81; green, 170; blue, 255 }  ,draw opacity=1 ][fill={rgb, 255:red, 150; green, 204; blue, 255 }  ,fill opacity=1 ] (328,209.5) -- (483,209.5) -- (483,224.5) -- (328,224.5) -- cycle ;
\draw    (100.36,204.3) -- (100.36,262.2) ;
\draw  [color={rgb, 255:red, 156; green, 158; blue, 255 }  ,draw opacity=1 ][fill={rgb, 255:red, 196; green, 197; blue, 255 }  ,fill opacity=1 ] (17.5,266.19) .. controls (17.5,263.96) and (19.31,262.15) .. (21.55,262.15) -- (217.95,262.15) .. controls (220.19,262.15) and (222,263.96) .. (222,266.19) -- (222,278.33) .. controls (222,280.56) and (220.19,282.38) .. (217.95,282.38) -- (21.55,282.38) .. controls (19.31,282.38) and (17.5,280.56) .. (17.5,278.33) -- cycle ;
\draw    (51.5,287.88) -- (409.5,287.88) ;
\draw    (51.5,287.88) -- (51.5,292.88) ;
\draw    (409.5,287.88) -- (409.5,292.88) ;
\draw    (169.36,288.45) -- (169.36,290.45) ;
\draw    (105.5,282.88) -- (105.5,287.88) ;
\draw    (281.36,288.45) -- (281.36,293.45) ;
\draw  [color={rgb, 255:red, 116; green, 232; blue, 164 }  ,draw opacity=1 ][fill={rgb, 255:red, 173; green, 243; blue, 203 }  ,fill opacity=1 ] (350,315.88) .. controls (350,313.67) and (351.79,311.88) .. (354,311.88) -- (479,311.88) .. controls (481.21,311.88) and (483,313.67) .. (483,315.88) -- (483,327.88) .. controls (483,330.08) and (481.21,331.88) .. (479,331.88) -- (354,331.88) .. controls (351.79,331.88) and (350,330.08) .. (350,327.88) -- cycle ;
\draw  [color={rgb, 255:red, 81; green, 170; blue, 255 }  ,draw opacity=1 ][fill={rgb, 255:red, 150; green, 204; blue, 255 }  ,fill opacity=1 ] (128,339.5) -- (286,339.5) -- (286,354.5) -- (128,354.5) -- cycle ;
\draw  [color={rgb, 255:red, 81; green, 170; blue, 255 }  ,draw opacity=1 ][fill={rgb, 255:red, 150; green, 204; blue, 255 }  ,fill opacity=1 ] (12,364.5) -- (208.6,364.5) -- (208.6,379.5) -- (12,379.5) -- cycle ;
\draw    (51.5,327.88) -- (51.5,363.78) ;
\draw    (169.36,330.45) -- (169.36,339.45) ;
\draw  [color={rgb, 255:red, 81; green, 170; blue, 255 }  ,draw opacity=1 ][fill={rgb, 255:red, 150; green, 204; blue, 255 }  ,fill opacity=1 ] (133,148.15) -- (253.4,148.15) -- (253.4,162.5) -- (133,162.5) -- cycle ;
\draw  [color={rgb, 255:red, 81; green, 170; blue, 255 }  ,draw opacity=1 ][fill={rgb, 255:red, 150; green, 204; blue, 255 }  ,fill opacity=1 ] (9,128.75) -- (127.4,128.75) -- (127.4,143.1) -- (9,143.1) -- cycle ;
\draw  [color={rgb, 255:red, 81; green, 170; blue, 255 }  ,draw opacity=1 ][fill={rgb, 255:red, 150; green, 204; blue, 255 }  ,fill opacity=1 ] (12,384.5) -- (208.6,384.5) -- (208.6,399.5) -- (12,399.5) -- cycle ;
\draw  [color={rgb, 255:red, 116; green, 232; blue, 164 }  ,draw opacity=1 ][fill={rgb, 255:red, 173; green, 243; blue, 203 }  ,fill opacity=1 ] (215,315.88) .. controls (215,313.67) and (216.79,311.88) .. (219,311.88) -- (344,311.88) .. controls (346.21,311.88) and (348,313.67) .. (348,315.88) -- (348,327.88) .. controls (348,330.08) and (346.21,331.88) .. (344,331.88) -- (219,331.88) .. controls (216.79,331.88) and (215,330.08) .. (215,327.88) -- cycle ;
\draw  [color={rgb, 255:red, 81; green, 170; blue, 255 }  ,draw opacity=1 ][fill={rgb, 255:red, 150; green, 204; blue, 255 }  ,fill opacity=1 ] (170,209.5) -- (325,209.5) -- (325,224.5) -- (170,224.5) -- cycle ;

\draw (250,43.12) node  [font=\footnotesize] [align=left] {\begin{minipage}[lt]{147.56pt}\setlength\topsep{0pt}
\begin{center}
Does a repeated measurement/ ground truth exist?
\end{center}

\end{minipage}};
\draw (100.5,71.17) node  [font=\footnotesize] [align=left] {\begin{minipage}[lt]{20.4pt}\setlength\topsep{0pt}
\begin{center}
no
\end{center}

\end{minipage}};
\draw (400.2,70.92) node  [font=\footnotesize] [align=left] {\begin{minipage}[lt]{20.4pt}\setlength\topsep{0pt}
\begin{center}
yes
\end{center}

\end{minipage}};
\draw (281.25,303.88) node  [font=\footnotesize] [align=left] {\begin{minipage}[lt]{48.62pt}\setlength\topsep{0pt}
\begin{center}
correlation
\end{center}

\end{minipage}};
\draw (409.25,303.38) node  [font=\footnotesize] [align=left] {\begin{minipage}[lt]{48.62pt}\setlength\topsep{0pt}
\begin{center}
distribution
\end{center}

\end{minipage}};
\draw (251.27,17) node  [font=\footnotesize] [align=left] {\begin{minipage}[lt]{96.87pt}\setlength\topsep{0pt}
\begin{center}
Accuracy
\end{center}

\end{minipage}};
\draw (120.7,93.26) node  [font=\footnotesize] [align=left] {\begin{minipage}[lt]{147.5pt}\setlength\topsep{0pt}
\begin{center}
Does a blank sample measurement exist?
\end{center}

\end{minipage}};
\draw (380,94.26) node  [font=\footnotesize] [align=left] {\begin{minipage}[lt]{147.5pt}\setlength\topsep{0pt}
\begin{center}
Accuracy vs trueness vs precision?
\end{center}

\end{minipage}};
\draw (195.45,136.32) node  [font=\footnotesize] [align=left] {\begin{minipage}[lt]{78.27pt}\setlength\topsep{0pt}
\begin{center}
Limit of quantification
\end{center}

\end{minipage}};
\draw (100.75,200.38) node  [font=\footnotesize] [align=left] {\begin{minipage}[lt]{48.62pt}\setlength\topsep{0pt}
\begin{center}
accuracy
\end{center}

\end{minipage}};
\draw (250.75,200.38) node  [font=\footnotesize] [align=left] {\begin{minipage}[lt]{48.62pt}\setlength\topsep{0pt}
\begin{center}
trueness
\end{center}

\end{minipage}};
\draw (409.75,199.38) node  [font=\footnotesize] [align=left] {\begin{minipage}[lt]{48.62pt}\setlength\topsep{0pt}
\begin{center}
precision
\end{center}

\end{minipage}};
\draw (405.5,217) node  [font=\footnotesize] [align=left] {\begin{minipage}[lt]{105.4pt}\setlength\topsep{0pt}
\begin{center}
Random error in instruments
\end{center}

\end{minipage}};
\draw (119.75,272.26) node  [font=\footnotesize] [align=left] {\begin{minipage}[lt]{128.18pt}\setlength\topsep{0pt}
\begin{center}
Which approach and condition?
\end{center}

\end{minipage}};
\draw (416.3,321.88) node  [font=\footnotesize] [align=left] {\begin{minipage}[lt]{85.89pt}\setlength\topsep{0pt}
\begin{center}
Distribution Metrics
\end{center}

\end{minipage}};
\draw (54,309.51) node  [font=\footnotesize] [align=left] {\begin{minipage}[lt]{72.08pt}\setlength\topsep{0pt}
\begin{center}
repeated measurements\\(same conditions)
\end{center}

\end{minipage}};
\draw (164.1,311.01) node  [font=\footnotesize] [align=left] {\begin{minipage}[lt]{70.58pt}\setlength\topsep{0pt}
\begin{center}
repeated measurements\\(changed conditions)
\end{center}

\end{minipage}};
\draw (207,347) node  [font=\footnotesize] [align=left] {\begin{minipage}[lt]{107.44pt}\setlength\topsep{0pt}
\begin{center}
Reproducibility variance
\end{center}

\end{minipage}};
\draw (110.3,372) node  [font=\footnotesize] [align=left] {\begin{minipage}[lt]{133.69pt}\setlength\topsep{0pt}
\begin{center}
Repeatability coefficient of variation
\end{center}

\end{minipage}};
\draw (195.65,155.32) node  [font=\footnotesize] [align=left] {\begin{minipage}[lt]{78.54pt}\setlength\topsep{0pt}
\begin{center}
Limit of detection
\end{center}

\end{minipage}};
\draw (68.2,135.92) node  [font=\footnotesize] [align=left] {\begin{minipage}[lt]{80.51pt}\setlength\topsep{0pt}
\begin{center}
Entropy
\end{center}

\end{minipage}};
\draw (405.5,237) node  [font=\footnotesize] [align=left] {\begin{minipage}[lt]{105.4pt}\setlength\topsep{0pt}
\begin{center}

\end{center}

\end{minipage}};
\draw (247.5,237) node  [font=\footnotesize] [align=left] {\begin{minipage}[lt]{105.4pt}\setlength\topsep{0pt}
\begin{center}

\end{center}

\end{minipage}};
\draw (110.3,392) node  [font=\footnotesize] [align=left] {\begin{minipage}[lt]{133.69pt}\setlength\topsep{0pt}
\begin{center}
Bland-Altman coefficient
\end{center}

\end{minipage}};
\draw (281.5,321.88) node  [font=\footnotesize] [align=left] {\begin{minipage}[lt]{85.89pt}\setlength\topsep{0pt}
\begin{center}
Correlation Coefficients
\end{center}

\end{minipage}};
\draw (247.5,217) node  [font=\footnotesize] [align=left] {\begin{minipage}[lt]{125.4pt}\setlength\topsep{0pt}
\begin{center}
Systematic error in instruments
\end{center}

\end{minipage}};
\draw (68.5,121.38) node  [font=\footnotesize] [align=left] {\begin{minipage}[lt]{20.4pt}\setlength\topsep{0pt}
\begin{center}
no
\end{center}

\end{minipage}};
\draw (195.4,122.25) node  [font=\footnotesize] [align=left] {\begin{minipage}[lt]{20.4pt}\setlength\topsep{0pt}
\begin{center}
yes
\end{center}

\end{minipage}};

\end{tikzpicture}

%% file: decision-trees/noisy-labels.tex
\tikzset{every picture/.style={line width=0.75pt}} 

\begin{tikzpicture}[x=0.75pt,y=0.75pt,yscale=-1,xscale=1]

\draw  [color={rgb, 255:red, 116; green, 232; blue, 164 }  ,draw opacity=1 ][fill={rgb, 255:red, 173; green, 243; blue, 203 }  ,fill opacity=1 ] (20.67,8.53) .. controls (20.67,6.32) and (22.46,4.53) .. (24.67,4.53) -- (476.67,4.53) .. controls (478.88,4.53) and (480.67,6.32) .. (480.67,8.53) -- (480.67,20.53) .. controls (480.67,22.73) and (478.88,24.53) .. (476.67,24.53) -- (24.67,24.53) .. controls (22.46,24.53) and (20.67,22.73) .. (20.67,20.53) -- cycle ;
\draw  [color={rgb, 255:red, 156; green, 158; blue, 255 }  ,draw opacity=1 ][fill={rgb, 255:red, 196; green, 197; blue, 255 }  ,fill opacity=1 ] (137,34.9) .. controls (137,32.69) and (138.79,30.9) .. (141,30.9) -- (358,30.9) .. controls (360.21,30.9) and (362,32.69) .. (362,34.9) -- (362,46.9) .. controls (362,49.11) and (360.21,50.9) .. (358,50.9) -- (141,50.9) .. controls (138.79,50.9) and (137,49.11) .. (137,46.9) -- cycle ;
\draw [color={rgb, 255:red, 156; green, 158; blue, 255 }  ,draw opacity=1 ][fill={rgb, 255:red, 196; green, 197; blue, 255 }  ,fill opacity=1 ]   (399.86,75.83) -- (399.86,80.83) ;
\draw  [color={rgb, 255:red, 156; green, 158; blue, 255 }  ,draw opacity=1 ][fill={rgb, 255:red, 196; green, 197; blue, 255 }  ,fill opacity=1 ] (189,85.9) .. controls (189,83.69) and (190.79,81.9) .. (193,81.9) -- (477,81.9) .. controls (479.21,81.9) and (481,83.69) .. (481,85.9) -- (481,97.9) .. controls (481,100.11) and (479.21,101.9) .. (477,101.9) -- (193,101.9) .. controls (190.79,101.9) and (189,100.11) .. (189,97.9) -- cycle ;
\draw  [color={rgb, 255:red, 156; green, 158; blue, 255 }  ,draw opacity=1 ][fill={rgb, 255:red, 196; green, 197; blue, 255 }  ,fill opacity=1 ] (22,165.9) .. controls (22,163.69) and (23.79,161.9) .. (26,161.9) -- (310,161.9) .. controls (312.21,161.9) and (314,163.69) .. (314,165.9) -- (314,177.9) .. controls (314,180.11) and (312.21,181.9) .. (310,181.9) -- (26,181.9) .. controls (23.79,181.9) and (22,180.11) .. (22,177.9) -- cycle ;
\draw    (100,127.4) -- (100,161.3) ;
\draw  [color={rgb, 255:red, 81; green, 170; blue, 255 }  ,draw opacity=1 ][fill={rgb, 255:red, 150; green, 204; blue, 255 }  ,fill opacity=1 ] (331,210.4) -- (481,210.4) -- (481,225.4) -- (331,225.4) -- cycle ;
\draw  [color={rgb, 255:red, 81; green, 170; blue, 255 }  ,draw opacity=1 ][fill={rgb, 255:red, 150; green, 204; blue, 255 }  ,fill opacity=1 ] (175,210.4) -- (325,210.4) -- (325,225.4) -- (175,225.4) -- cycle ;
\draw  [color={rgb, 255:red, 156; green, 158; blue, 255 }  ,draw opacity=1 ][fill={rgb, 255:red, 196; green, 197; blue, 255 }  ,fill opacity=1 ] (22.4,238.5) .. controls (22.4,236.29) and (24.19,234.5) .. (26.4,234.5) -- (310.4,234.5) .. controls (312.61,234.5) and (314.4,236.29) .. (314.4,238.5) -- (314.4,250.5) .. controls (314.4,252.71) and (312.61,254.5) .. (310.4,254.5) -- (26.4,254.5) .. controls (24.19,254.5) and (22.4,252.71) .. (22.4,250.5) -- cycle ;
\draw    (99.65,208.93) -- (99.65,233.9) ;
\draw  [color={rgb, 255:red, 81; green, 170; blue, 255 }  ,draw opacity=1 ][fill={rgb, 255:red, 150; green, 204; blue, 255 }  ,fill opacity=1 ] (325,281.4) -- (475,281.4) -- (475,296.4) -- (325,296.4) -- cycle ;
\draw  [color={rgb, 255:red, 156; green, 158; blue, 255 }  ,draw opacity=1 ][fill={rgb, 255:red, 196; green, 197; blue, 255 }  ,fill opacity=1 ] (23.4,308.7) .. controls (23.4,306.49) and (25.19,304.7) .. (27.4,304.7) -- (311.4,304.7) .. controls (313.61,304.7) and (315.4,306.49) .. (315.4,308.7) -- (315.4,320.7) .. controls (315.4,322.91) and (313.61,324.7) .. (311.4,324.7) -- (27.4,324.7) .. controls (25.19,324.7) and (23.4,322.91) .. (23.4,320.7) -- cycle ;
\draw  [color={rgb, 255:red, 81; green, 170; blue, 255 }  ,draw opacity=1 ][fill={rgb, 255:red, 150; green, 204; blue, 255 }  ,fill opacity=1 ] (24,355.4) -- (174,355.4) -- (174,370.4) -- (24,370.4) -- cycle ;
\draw  [color={rgb, 255:red, 81; green, 170; blue, 255 }  ,draw opacity=1 ][fill={rgb, 255:red, 150; green, 204; blue, 255 }  ,fill opacity=1 ] (325.6,355.5) -- (475.6,355.5) -- (475.6,370.5) -- (325.6,370.5) -- cycle ;
\draw  [color={rgb, 255:red, 81; green, 170; blue, 255 }  ,draw opacity=1 ][fill={rgb, 255:red, 150; green, 204; blue, 255 }  ,fill opacity=1 ] (325.4,375.5) -- (475.4,375.5) -- (475.4,390.5) -- (325.4,390.5) -- cycle ;
\draw  [color={rgb, 255:red, 116; green, 232; blue, 164 }  ,draw opacity=1 ][fill={rgb, 255:red, 173; green, 243; blue, 203 }  ,fill opacity=1 ] (175,137.4) .. controls (175,135.19) and (176.79,133.4) .. (179,133.4) -- (321,133.4) .. controls (323.21,133.4) and (325,135.19) .. (325,137.4) -- (325,149.4) .. controls (325,151.61) and (323.21,153.4) .. (321,153.4) -- (179,153.4) .. controls (176.79,153.4) and (175,151.61) .. (175,149.4) -- cycle ;
\draw  [color={rgb, 255:red, 116; green, 232; blue, 164 }  ,draw opacity=1 ][fill={rgb, 255:red, 173; green, 243; blue, 203 }  ,fill opacity=1 ] (331.5,137.9) .. controls (331.5,135.69) and (333.29,133.9) .. (335.5,133.9) -- (477.5,133.9) .. controls (479.71,133.9) and (481.5,135.69) .. (481.5,137.9) -- (481.5,149.9) .. controls (481.5,152.11) and (479.71,153.9) .. (477.5,153.9) -- (335.5,153.9) .. controls (333.29,153.9) and (331.5,152.11) .. (331.5,149.9) -- cycle ;
\draw    (249.5,127.83) -- (249.5,132.83) ;
\draw    (100,107.4) -- (400,107.4) ;
\draw    (100,107.4) -- (100,112.4) ;
\draw    (400,107.4) -- (400,112.4) ;
\draw    (249.86,101.97) -- (249.86,106.97) ;
\draw    (249.86,107.97) -- (249.86,112.97) ;

\draw    (100,187.4) -- (400,187.4) ;
\draw    (100,187.4) -- (100,192.4) ;
\draw    (400,187.4) -- (400,192.4) ;
\draw    (249.86,181.97) -- (249.86,186.97) ;
\draw    (249.86,187.97) -- (249.86,192.97) ;

\draw    (100,260.4) -- (400,260.4) ;
\draw    (100,260.4) -- (100,265.4) ;
\draw    (400,260.4) -- (400,265.4) ;
\draw    (249.86,254.97) -- (249.86,259.97) ;

\draw    (400.5,128.83) -- (400.5,133.83) ;
\draw    (99.65,278.93) -- (99.65,303.9) ;
\draw    (100,330.4) -- (400,330.4) ;
\draw    (100,330.4) -- (100,335.4) ;
\draw    (400,330.4) -- (400,335.4) ;
\draw    (249.86,324.97) -- (249.86,329.97) ;

\draw    (100,56.4) -- (400,56.4) ;
\draw    (100,56.4) -- (100,61.4) ;
\draw    (400,56.4) -- (400,61.4) ;
\draw    (249.86,50.97) -- (249.86,55.97) ;

\draw (249.5,40.65) node  [font=\footnotesize] [align=left] {\begin{minipage}[lt]{147.56pt}\setlength\topsep{0pt}
\begin{center}
How was your dataset labeled?
\end{center}

\end{minipage}};
\draw (109,70.5) node  [font=\footnotesize] [align=left] {\begin{minipage}[lt]{170pt}\setlength\topsep{0pt}
\begin{center}
single annotator: no reference
\end{center}

\end{minipage}};
\draw (364.65,71.55) node  [font=\footnotesize] [align=left] {\begin{minipage}[lt]{256.72pt}\setlength\topsep{0pt}
\begin{center}
multiple annotators
\end{center}

\end{minipage}};
\draw (344,91.9) node  [font=\footnotesize] [align=left] {\begin{minipage}[lt]{193.12pt}\setlength\topsep{0pt}
\begin{center}
How are you assessing agreement or similarity?
\end{center}

\end{minipage}};
\draw (100.25,122.9) node  [font=\footnotesize] [align=left] {\begin{minipage}[lt]{48.62pt}\setlength\topsep{0pt}
\begin{center}
inter-rater
\end{center}

\end{minipage}};
\draw (249.75,122.4) node  [font=\footnotesize] [align=left] {\begin{minipage}[lt]{48.62pt}\setlength\topsep{0pt}
\begin{center}
correlation
\end{center}

\end{minipage}};
\draw (399.75,122.9) node  [font=\footnotesize] [align=left] {\begin{minipage}[lt]{48.62pt}\setlength\topsep{0pt}
\begin{center}
distribution
\end{center}

\end{minipage}};
\draw (168,171.9) node  [font=\footnotesize] [align=left] {\begin{minipage}[lt]{193.12pt}\setlength\topsep{0pt}
\begin{center}
Which ML task are you addressing?
\end{center}

\end{minipage}};
\draw (99.5,202.4) node  [font=\footnotesize] [align=left] {\begin{minipage}[lt]{48.62pt}\setlength\topsep{0pt}
\begin{center}
classification
\end{center}

\end{minipage}};
\draw (250,202.9) node  [font=\footnotesize] [align=left] {\begin{minipage}[lt]{106.62pt}\setlength\topsep{0pt}
\begin{center}
semantic segmentation
\end{center}

\end{minipage}};
\draw (399.88,202.4) node  [font=\footnotesize] [align=left] {\begin{minipage}[lt]{66.47pt}\setlength\topsep{0pt}
\begin{center}
object detection
\end{center}

\end{minipage}};
\draw (406,217.9) node  [font=\footnotesize] [align=left] {\begin{minipage}[lt]{102pt}\setlength\topsep{0pt}
\begin{center}
Intersection over union
\end{center}

\end{minipage}};
\draw (250,217.9) node  [font=\footnotesize] [align=left] {\begin{minipage}[lt]{102pt}\setlength\topsep{0pt}
\begin{center}
Dice similarity score
\end{center}

\end{minipage}};
\draw (168.4,244.5) node  [font=\footnotesize] [align=left] {\begin{minipage}[lt]{193.12pt}\setlength\topsep{0pt}
\begin{center}
Is there a substantial number of missing labels?
\end{center}

\end{minipage}};
\draw (100,273.38) node  [font=\footnotesize] [align=left] {\begin{minipage}[lt]{73.44pt}\setlength\topsep{0pt}
\begin{center}
no
\end{center}

\end{minipage}};
\draw (400.4,273.53) node  [font=\footnotesize] [align=left] {\begin{minipage}[lt]{71.81pt}\setlength\topsep{0pt}
\begin{center}
yes
\end{center}

\end{minipage}};
\draw (400,288.9) node  [font=\footnotesize] [align=left] {\begin{minipage}[lt]{102pt}\setlength\topsep{0pt}
\begin{center}
Krippendorff's alpha
\end{center}

\end{minipage}};
\draw (169.4,314.7) node  [font=\footnotesize] [align=left] {\begin{minipage}[lt]{193.12pt}\setlength\topsep{0pt}
\begin{center}
How many raters are in your dataset?
\end{center}

\end{minipage}};
\draw (100,345.83) node  [font=\footnotesize] [align=left] {\begin{minipage}[lt]{73.44pt}\setlength\topsep{0pt}
\begin{center}
two
\end{center}

\end{minipage}};
\draw (400.4,345.63) node  [font=\footnotesize] [align=left] {\begin{minipage}[lt]{71.81pt}\setlength\topsep{0pt}
\begin{center}
more than two
\end{center}

\end{minipage}};
\draw (99,362.9) node  [font=\footnotesize] [align=left] {\begin{minipage}[lt]{102pt}\setlength\topsep{0pt}
\begin{center}
Cohen's kappa
\end{center}

\end{minipage}};
\draw (400.6,363) node  [font=\footnotesize] [align=left] {\begin{minipage}[lt]{102pt}\setlength\topsep{0pt}
\begin{center}
Fleiss' kappa
\end{center}

\end{minipage}};
\draw (400.4,383) node  [font=\footnotesize] [align=left] {\begin{minipage}[lt]{102pt}\setlength\topsep{0pt}
\begin{center}
Kendall's W
\end{center}

\end{minipage}};
\draw (249.77,143.4) node  [font=\footnotesize] [align=left] {\begin{minipage}[lt]{96.87pt}\setlength\topsep{0pt}
\begin{center}
Correlation Coefficients
\end{center}

\end{minipage}};
\draw (406.28,143.9) node  [font=\footnotesize] [align=left] {\begin{minipage}[lt]{96.87pt}\setlength\topsep{0pt}
\begin{center}
Distribution Metrics
\end{center}

\end{minipage}};
\draw (250.77,14.53) node  [font=\footnotesize] [align=left] {\begin{minipage}[lt]{96.87pt}\setlength\topsep{0pt}
\begin{center}
Noisy Labels
\end{center}

\end{minipage}};

\end{tikzpicture}

%% file: decision-trees/completeness.tex
\tikzset{every picture/.style={line width=0.75pt}} 

\begin{tikzpicture}[x=0.75pt,y=0.75pt,yscale=-1,xscale=1]

\draw  [color={rgb, 255:red, 116; green, 232; blue, 164 }  ,draw opacity=1 ][fill={rgb, 255:red, 173; green, 243; blue, 203 }  ,fill opacity=1 ] (10.85,17) .. controls (10.85,14.79) and (12.64,13) .. (14.85,13) -- (466.85,13) .. controls (469.06,13) and (470.85,14.79) .. (470.85,17) -- (470.85,29) .. controls (470.85,31.21) and (469.06,33) .. (466.85,33) -- (14.85,33) .. controls (12.64,33) and (10.85,31.21) .. (10.85,29) -- cycle ;
\draw  [color={rgb, 255:red, 156; green, 158; blue, 255 }  ,draw opacity=1 ][fill={rgb, 255:red, 196; green, 197; blue, 255 }  ,fill opacity=1 ] (127.58,43.78) .. controls (127.58,41.57) and (129.37,39.78) .. (131.58,39.78) -- (348.58,39.78) .. controls (350.79,39.78) and (352.58,41.57) .. (352.58,43.78) -- (352.58,55.78) .. controls (352.58,57.98) and (350.79,59.78) .. (348.58,59.78) -- (131.58,59.78) .. controls (129.37,59.78) and (127.58,57.98) .. (127.58,55.78) -- cycle ;
\draw    (100.18,65.28) -- (380.18,65.28) ;
\draw    (100.18,65.28) -- (100.18,70.28) ;
\draw    (380.18,65.28) -- (380.18,70.28) ;
\draw    (240.04,65.85) -- (240.04,70.85) ;
\draw    (240.04,59.85) -- (240.04,64.85) ;
\draw  [color={rgb, 255:red, 81; green, 170; blue, 255 }  ,draw opacity=1 ][fill={rgb, 255:red, 150; green, 204; blue, 255 }  ,fill opacity=1 ] (37.13,89.4) -- (163.91,89.4) -- (163.91,104.4) -- (37.13,104.4) -- cycle ;
\draw  [color={rgb, 255:red, 81; green, 170; blue, 255 }  ,draw opacity=1 ][fill={rgb, 255:red, 150; green, 204; blue, 255 }  ,fill opacity=1 ] (172.42,89.4) -- (306.9,89.4) -- (306.9,104.4) -- (172.42,104.4) -- cycle ;
\draw  [color={rgb, 255:red, 81; green, 170; blue, 255 }  ,draw opacity=1 ][fill={rgb, 255:red, 150; green, 204; blue, 255 }  ,fill opacity=1 ] (316.13,89.4) -- (442.91,89.4) -- (442.91,104.4) -- (316.13,104.4) -- cycle ;

\draw (240.38,49.14) node  [font=\footnotesize] [align=left] {\begin{minipage}[lt]{152.59pt}\setlength\topsep{0pt}
\begin{center}
Which completeness are you interested in?
\end{center}

\end{minipage}};
\draw (240.85,23) node  [font=\footnotesize] [align=left] {\begin{minipage}[lt]{96.87pt}\setlength\topsep{0pt}
\begin{center}
Completeness
\end{center}

\end{minipage}};
\draw (99.43,81.78) node  [font=\footnotesize] [align=left] {\begin{minipage}[lt]{48.62pt}\setlength\topsep{0pt}
\begin{center}
general
\end{center}

\end{minipage}};
\draw (240.43,81.78) node  [font=\footnotesize] [align=left] {\begin{minipage}[lt]{48.62pt}\setlength\topsep{0pt}
\begin{center}
patient-level
\end{center}

\end{minipage}};
\draw (380.43,80.78) node  [font=\footnotesize] [align=left] {\begin{minipage}[lt]{48.62pt}\setlength\topsep{0pt}
\begin{center}
record
\end{center}

\end{minipage}};
\draw (100.52,96.9) node  [font=\footnotesize] [align=left] {\begin{minipage}[lt]{86.2pt}\setlength\topsep{0pt}
\begin{center}
Completeness
\end{center}

\end{minipage}};
\draw (239.66,96.9) node  [font=\footnotesize] [align=left] {\begin{minipage}[lt]{91.45pt}\setlength\topsep{0pt}
\begin{center}
Patient-level completeness
\end{center}

\end{minipage}};
\draw (379.52,96.9) node  [font=\footnotesize] [align=left] {\begin{minipage}[lt]{86.2pt}\setlength\topsep{0pt}
\begin{center}
Record completeness
\end{center}

\end{minipage}};

\end{tikzpicture}

%% file: decision-trees/syntatic-consistency.tex
\tikzset{every picture/.style={line width=0.75pt}} 

\begin{tikzpicture}[x=0.75pt,y=0.75pt,yscale=-1,xscale=1]

\draw  [color={rgb, 255:red, 116; green, 232; blue, 164 }  ,draw opacity=1 ][fill={rgb, 255:red, 173; green, 243; blue, 203 }  ,fill opacity=1 ] (20,19.6) .. controls (20,17.39) and (21.79,15.6) .. (24,15.6) -- (476,15.6) .. controls (478.21,15.6) and (480,17.39) .. (480,19.6) -- (480,31.6) .. controls (480,33.81) and (478.21,35.6) .. (476,35.6) -- (24,35.6) .. controls (21.79,35.6) and (20,33.81) .. (20,31.6) -- cycle ;
\draw  [color={rgb, 255:red, 81; green, 170; blue, 255 }  ,draw opacity=1 ][fill={rgb, 255:red, 150; green, 204; blue, 255 }  ,fill opacity=1 ] (186.29,45.4) -- (313.06,45.4) -- (313.06,60.4) -- (186.29,60.4) -- cycle ;

\draw (250.11,26) node  [font=\footnotesize] [align=left] {\begin{minipage}[lt]{96.87pt}\setlength\topsep{0pt}
\begin{center}
Syntactic Consistency
\end{center}

\end{minipage}};
\draw (249.67,52.9) node  [font=\footnotesize] [align=left] {\begin{minipage}[lt]{86.2pt}\setlength\topsep{0pt}
\begin{center}
Syntactic accuracy
\end{center}

\end{minipage}};

\end{tikzpicture}

%% file: decision-trees/homogenity.tex
\tikzset{every picture/.style={line width=0.75pt}} 

\begin{tikzpicture}[x=0.75pt,y=0.75pt,yscale=-1,xscale=1]

\draw  [color={rgb, 255:red, 116; green, 232; blue, 164 }  ,draw opacity=1 ][fill={rgb, 255:red, 173; green, 243; blue, 203 }  ,fill opacity=1 ] (20,21.6) .. controls (20,19.39) and (21.79,17.6) .. (24,17.6) -- (476,17.6) .. controls (478.21,17.6) and (480,19.39) .. (480,21.6) -- (480,33.6) .. controls (480,35.81) and (478.21,37.6) .. (476,37.6) -- (24,37.6) .. controls (21.79,37.6) and (20,35.81) .. (20,33.6) -- cycle ;
\draw  [color={rgb, 255:red, 116; green, 232; blue, 164 }  ,draw opacity=1 ][fill={rgb, 255:red, 173; green, 243; blue, 203 }  ,fill opacity=1 ] (141,55.72) .. controls (141,53.06) and (143.16,50.9) .. (145.82,50.9) -- (353.88,50.9) .. controls (356.54,50.9) and (358.7,53.06) .. (358.7,55.72) -- (358.7,70.18) .. controls (358.7,72.84) and (356.54,75) .. (353.88,75) -- (145.82,75) .. controls (143.16,75) and (141,72.84) .. (141,70.18) -- cycle ;

\draw (250.11,28) node  [font=\footnotesize] [align=left] {\begin{minipage}[lt]{96.87pt}\setlength\topsep{0pt}
\begin{center}
Homogeneity
\end{center}

\end{minipage}};
\draw (249.52,62.45) node  [font=\footnotesize] [align=left] {\begin{minipage}[lt]{140.59pt}\setlength\topsep{0pt}
\begin{center}
Distribution Metrics\\(Internal distribution as reference)
\end{center}

\end{minipage}};

\end{tikzpicture}

%% file: decision-trees/distribution-drift.tex
\tikzset{every picture/.style={line width=0.75pt}} 

\begin{tikzpicture}[x=0.75pt,y=0.75pt,yscale=-1,xscale=1]

\draw  [color={rgb, 255:red, 116; green, 232; blue, 164 }  ,draw opacity=1 ][fill={rgb, 255:red, 173; green, 243; blue, 203 }  ,fill opacity=1 ] (21,19.6) .. controls (21,17.39) and (22.79,15.6) .. (25,15.6) -- (477,15.6) .. controls (479.21,15.6) and (481,17.39) .. (481,19.6) -- (481,31.6) .. controls (481,33.81) and (479.21,35.6) .. (477,35.6) -- (25,35.6) .. controls (22.79,35.6) and (21,33.81) .. (21,31.6) -- cycle ;
\draw  [color={rgb, 255:red, 156; green, 158; blue, 255 }  ,draw opacity=1 ][fill={rgb, 255:red, 196; green, 197; blue, 255 }  ,fill opacity=1 ] (138.73,46.37) .. controls (138.73,44.17) and (140.52,42.37) .. (142.73,42.37) -- (359.73,42.37) .. controls (361.94,42.37) and (363.73,44.17) .. (363.73,46.37) -- (363.73,58.37) .. controls (363.73,60.58) and (361.94,62.37) .. (359.73,62.37) -- (142.73,62.37) .. controls (140.52,62.37) and (138.73,60.58) .. (138.73,58.37) -- cycle ;
\draw    (109.33,67.88) -- (389.33,67.88) ;
\draw    (109.33,67.88) -- (109.33,72.88) ;
\draw    (389.33,67.88) -- (389.33,72.88) ;
\draw    (249.2,62.45) -- (249.2,67.45) ;

\draw  [color={rgb, 255:red, 116; green, 232; blue, 164 }  ,draw opacity=1 ][fill={rgb, 255:red, 173; green, 243; blue, 203 }  ,fill opacity=1 ] (313,91.9) .. controls (313,89.69) and (314.79,87.9) .. (317,87.9) -- (459,87.9) .. controls (461.21,87.9) and (463,89.69) .. (463,91.9) -- (463,103.9) .. controls (463,106.11) and (461.21,107.9) .. (459,107.9) -- (317,107.9) .. controls (314.79,107.9) and (313,106.11) .. (313,103.9) -- cycle ;
\draw  [color={rgb, 255:red, 81; green, 170; blue, 255 }  ,draw opacity=1 ][fill={rgb, 255:red, 150; green, 204; blue, 255 }  ,fill opacity=1 ] (46.29,88.4) -- (173.06,88.4) -- (173.06,103.4) -- (46.29,103.4) -- cycle ;

\draw (250.83,52.12) node  [font=\footnotesize] [align=left] {\begin{minipage}[lt]{147.56pt}\setlength\topsep{0pt}
\begin{center}
Do you want to detect a change in signal or difference in distribution?
\end{center}

\end{minipage}};
\draw (108.83,81.84) node  [font=\footnotesize] [align=left] {\begin{minipage}[lt]{51pt}\setlength\topsep{0pt}
\begin{center}
signal
\end{center}

\end{minipage}};
\draw (251.11,26) node  [font=\footnotesize] [align=left] {\begin{minipage}[lt]{96.87pt}\setlength\topsep{0pt}
\begin{center}
Distribution Drift
\end{center}

\end{minipage}};
\draw (389.51,81.07) node  [font=\footnotesize] [align=left] {\begin{minipage}[lt]{56.88pt}\setlength\topsep{0pt}
\begin{center}
distribution
\end{center}

\end{minipage}};
\draw (387.78,97.9) node  [font=\footnotesize] [align=left] {\begin{minipage}[lt]{96.87pt}\setlength\topsep{0pt}
\begin{center}
Distribution metrics
\end{center}

\end{minipage}};
\draw (109.67,95.9) node  [font=\footnotesize] [align=left] {\begin{minipage}[lt]{86.2pt}\setlength\topsep{0pt}
\begin{center}
Page-Hinkley test
\end{center}

\end{minipage}};

\end{tikzpicture}

%% file: decision-trees/dataset-size.tex
\tikzset{every picture/.style={line width=0.75pt}} 

\begin{tikzpicture}[x=0.75pt,y=0.75pt,yscale=-1,xscale=1]

\draw  [color={rgb, 255:red, 116; green, 232; blue, 164 }  ,draw opacity=1 ][fill={rgb, 255:red, 173; green, 243; blue, 203 }  ,fill opacity=1 ] (22,8.6) .. controls (22,6.39) and (23.79,4.6) .. (26,4.6) -- (478,4.6) .. controls (480.21,4.6) and (482,6.39) .. (482,8.6) -- (482,20.6) .. controls (482,22.81) and (480.21,24.6) .. (478,24.6) -- (26,24.6) .. controls (23.79,24.6) and (22,22.81) .. (22,20.6) -- cycle ;
\draw  [color={rgb, 255:red, 81; green, 170; blue, 255 }  ,draw opacity=1 ][fill={rgb, 255:red, 150; green, 204; blue, 255 }  ,fill opacity=1 ] (188.29,34.4) -- (315.06,34.4) -- (315.06,49.4) -- (188.29,49.4) -- cycle ;

\draw (252.11,15) node  [font=\footnotesize] [align=left] {\begin{minipage}[lt]{96.87pt}\setlength\topsep{0pt}
\begin{center}
Dataset Size
\end{center}

\end{minipage}};
\draw (251.67,41.9) node  [font=\footnotesize] [align=left] {\begin{minipage}[lt]{86.2pt}\setlength\topsep{0pt}
\begin{center}
Dataset size
\end{center}

\end{minipage}};

\end{tikzpicture}

%% file: decision-trees/granularity.tex
\tikzset{every picture/.style={line width=0.75pt}} 

\begin{tikzpicture}[x=0.75pt,y=0.75pt,yscale=-1,xscale=1]

\draw  [color={rgb, 255:red, 116; green, 232; blue, 164 }  ,draw opacity=1 ][fill={rgb, 255:red, 173; green, 243; blue, 203 }  ,fill opacity=1 ] (21,17.6) .. controls (21,15.39) and (22.79,13.6) .. (25,13.6) -- (477,13.6) .. controls (479.21,13.6) and (481,15.39) .. (481,17.6) -- (481,29.6) .. controls (481,31.81) and (479.21,33.6) .. (477,33.6) -- (25,33.6) .. controls (22.79,33.6) and (21,31.81) .. (21,29.6) -- cycle ;
\draw  [color={rgb, 255:red, 156; green, 158; blue, 255 }  ,draw opacity=1 ][fill={rgb, 255:red, 196; green, 197; blue, 255 }  ,fill opacity=1 ] (138.73,44.37) .. controls (138.73,42.17) and (140.52,40.37) .. (142.73,40.37) -- (359.73,40.37) .. controls (361.94,40.37) and (363.73,42.17) .. (363.73,44.37) -- (363.73,56.37) .. controls (363.73,58.58) and (361.94,60.37) .. (359.73,60.37) -- (142.73,60.37) .. controls (140.52,60.37) and (138.73,58.58) .. (138.73,56.37) -- cycle ;
\draw    (69,65.88) -- (430.68,65.88) ;
\draw    (69,65.88) -- (69,70.88) ;
\draw    (430.68,65.88) -- (430.68,70.88) ;
\draw    (250.16,60.45) -- (250.16,65.45) ;
\draw    (190,65.88) -- (190,70.88) ;
\draw    (311,65.88) -- (311,70.88) ;
\draw  [color={rgb, 255:red, 81; green, 170; blue, 255 }  ,draw opacity=1 ][fill={rgb, 255:red, 150; green, 204; blue, 255 }  ,fill opacity=1 ] (257.27,87.4) -- (363.9,87.4) -- (363.9,102.4) -- (257.27,102.4) -- cycle ;
\draw  [color={rgb, 255:red, 81; green, 170; blue, 255 }  ,draw opacity=1 ][fill={rgb, 255:red, 150; green, 204; blue, 255 }  ,fill opacity=1 ] (376.27,87.4) -- (482.9,87.4) -- (482.9,102.4) -- (376.27,102.4) -- cycle ;
\draw  [color={rgb, 255:red, 81; green, 170; blue, 255 }  ,draw opacity=1 ][fill={rgb, 255:red, 150; green, 204; blue, 255 }  ,fill opacity=1 ] (137.27,87.4) -- (243.9,87.4) -- (243.9,102.4) -- (137.27,102.4) -- cycle ;
\draw  [color={rgb, 255:red, 81; green, 170; blue, 255 }  ,draw opacity=1 ][fill={rgb, 255:red, 150; green, 204; blue, 255 }  ,fill opacity=1 ] (17.27,87.4) -- (123.9,87.4) -- (123.9,102.4) -- (17.27,102.4) -- cycle ;

\draw (250.83,50.12) node  [font=\footnotesize] [align=left] {\begin{minipage}[lt]{147.56pt}\setlength\topsep{0pt}
\begin{center}
Which data modalities are of interest?
\end{center}

\end{minipage}};
\draw (68.89,78.84) node  [font=\footnotesize] [align=left] {\begin{minipage}[lt]{55.26pt}\setlength\topsep{0pt}
\begin{center}
tabular
\end{center}

\end{minipage}};
\draw (251.11,24) node  [font=\footnotesize] [align=left] {\begin{minipage}[lt]{96.87pt}\setlength\topsep{0pt}
\begin{center}
Granularity
\end{center}

\end{minipage}};
\draw (430.91,79.07) node  [font=\footnotesize] [align=left] {\begin{minipage}[lt]{75.3pt}\setlength\topsep{0pt}
\begin{center}
hierarchical
\end{center}

\end{minipage}};
\draw (189.89,78.84) node  [font=\footnotesize] [align=left] {\begin{minipage}[lt]{55.26pt}\setlength\topsep{0pt}
\begin{center}
image
\end{center}

\end{minipage}};
\draw (310.89,78.84) node  [font=\footnotesize] [align=left] {\begin{minipage}[lt]{95.26pt}\setlength\topsep{0pt}
\begin{center}
image, time series
\end{center}

\end{minipage}};
\draw (310.59,94.9) node  [font=\footnotesize] [align=left] {\begin{minipage}[lt]{72.51pt}\setlength\topsep{0pt}
\begin{center}
Sampling frequency
\end{center}

\end{minipage}};
\draw (429.59,94.9) node  [font=\footnotesize] [align=left] {\begin{minipage}[lt]{72.51pt}\setlength\topsep{0pt}
\begin{center}
Label granuality
\end{center}

\end{minipage}};
\draw (190.59,94.9) node  [font=\footnotesize] [align=left] {\begin{minipage}[lt]{72.51pt}\setlength\topsep{0pt}
\begin{center}
Resolution
\end{center}

\end{minipage}};
\draw (70.59,94.9) node  [font=\footnotesize] [align=left] {\begin{minipage}[lt]{72.51pt}\setlength\topsep{0pt}
\begin{center}
Granularity
\end{center}

\end{minipage}};

\end{tikzpicture}

%% file: decision-trees/variety.tex
\tikzset{every picture/.style={line width=0.75pt}} 

\begin{tikzpicture}[x=0.75pt,y=0.75pt,yscale=-1,xscale=1]

\draw  [color={rgb, 255:red, 116; green, 232; blue, 164 }  ,draw opacity=1 ][fill={rgb, 255:red, 173; green, 243; blue, 203 }  ,fill opacity=1 ] (20,18.6) .. controls (20,16.39) and (21.79,14.6) .. (24,14.6) -- (476,14.6) .. controls (478.21,14.6) and (480,16.39) .. (480,18.6) -- (480,30.6) .. controls (480,32.81) and (478.21,34.6) .. (476,34.6) -- (24,34.6) .. controls (21.79,34.6) and (20,32.81) .. (20,30.6) -- cycle ;
\draw  [color={rgb, 255:red, 116; green, 232; blue, 164 }  ,draw opacity=1 ][fill={rgb, 255:red, 173; green, 243; blue, 203 }  ,fill opacity=1 ] (127,53.27) .. controls (127,50.61) and (129.16,48.45) .. (131.82,48.45) -- (368.88,48.45) .. controls (371.54,48.45) and (373.7,50.61) .. (373.7,53.27) -- (373.7,67.73) .. controls (373.7,70.39) and (371.54,72.55) .. (368.88,72.55) -- (131.82,72.55) .. controls (129.16,72.55) and (127,70.39) .. (127,67.73) -- cycle ;

\draw (250.11,25) node  [font=\footnotesize] [align=left] {\begin{minipage}[lt]{96.87pt}\setlength\topsep{0pt}
\begin{center}
Variety
\end{center}

\end{minipage}};
\draw (249.98,60) node  [font=\footnotesize] [align=left] {\begin{minipage}[lt]{159.31pt}\setlength\topsep{0pt}
\begin{center}
Distribution metrics\\(Target distribution as reference)
\end{center}

\end{minipage}};

\end{tikzpicture}

%% file: decision-trees/target-class-balance.tex
\tikzset{every picture/.style={line width=0.75pt}} 

\begin{tikzpicture}[x=0.75pt,y=0.75pt,yscale=-1,xscale=1]

\draw  [color={rgb, 255:red, 116; green, 232; blue, 164 }  ,draw opacity=1 ][fill={rgb, 255:red, 173; green, 243; blue, 203 }  ,fill opacity=1 ] (19.33,13.6) .. controls (19.33,11.39) and (21.12,9.6) .. (23.33,9.6) -- (475.33,9.6) .. controls (477.54,9.6) and (479.33,11.39) .. (479.33,13.6) -- (479.33,25.6) .. controls (479.33,27.81) and (477.54,29.6) .. (475.33,29.6) -- (23.33,29.6) .. controls (21.12,29.6) and (19.33,27.81) .. (19.33,25.6) -- cycle ;
\draw  [color={rgb, 255:red, 156; green, 158; blue, 255 }  ,draw opacity=1 ][fill={rgb, 255:red, 196; green, 197; blue, 255 }  ,fill opacity=1 ] (137.07,40.37) .. controls (137.07,38.17) and (138.86,36.37) .. (141.07,36.37) -- (358.07,36.37) .. controls (360.28,36.37) and (362.07,38.17) .. (362.07,40.37) -- (362.07,52.37) .. controls (362.07,54.58) and (360.28,56.37) .. (358.07,56.37) -- (141.07,56.37) .. controls (138.86,56.37) and (137.07,54.58) .. (137.07,52.37) -- cycle ;
\draw    (107.67,61.88) -- (369.67,61.88) ;
\draw    (107.67,61.88) -- (107.67,66.88) ;
\draw    (369.67,61.88) -- (369.67,66.88) ;
\draw    (249.53,56.45) -- (249.53,61.45) ;
\draw  [color={rgb, 255:red, 156; green, 158; blue, 255 }  ,draw opacity=1 ][fill={rgb, 255:red, 196; green, 197; blue, 255 }  ,fill opacity=1 ] (18.07,91.37) .. controls (18.07,89.17) and (19.86,87.37) .. (22.07,87.37) -- (202.05,87.37) .. controls (204.26,87.37) and (206.05,89.17) .. (206.05,91.37) -- (206.05,103.37) .. controls (206.05,105.58) and (204.26,107.37) .. (202.05,107.37) -- (22.07,107.37) .. controls (19.86,107.37) and (18.07,105.58) .. (18.07,103.37) -- cycle ;
\draw    (89.47,112.88) -- (204.59,112.88) ;
\draw    (89.47,112.88) -- (89.47,117.88) ;
\draw    (204.59,112.88) -- (204.59,117.88) ;
\draw    (107.46,107.45) -- (107.46,112.45) ;
\draw  [color={rgb, 255:red, 81; green, 170; blue, 255 }  ,draw opacity=1 ][fill={rgb, 255:red, 150; green, 204; blue, 255 }  ,fill opacity=1 ] (16.63,136.4) -- (164,136.4) -- (164,151.4) -- (16.63,151.4) -- cycle ;
\draw    (204.59,133.88) -- (204.59,162.88) ;
\draw  [color={rgb, 255:red, 156; green, 158; blue, 255 }  ,draw opacity=1 ][fill={rgb, 255:red, 196; green, 197; blue, 255 }  ,fill opacity=1 ] (111.07,166.37) .. controls (111.07,164.17) and (112.86,162.37) .. (115.07,162.37) -- (295.05,162.37) .. controls (297.26,162.37) and (299.05,164.17) .. (299.05,166.37) -- (299.05,178.37) .. controls (299.05,180.58) and (297.26,182.37) .. (295.05,182.37) -- (115.07,182.37) .. controls (112.86,182.37) and (111.07,180.58) .. (111.07,178.37) -- cycle ;
\draw    (91.78,187.88) -- (299.91,187.88) ;
\draw    (91.78,187.88) -- (91.78,192.88) ;
\draw    (299.91,187.88) -- (299.91,192.88) ;
\draw    (204.78,182.45) -- (204.78,187.45) ;
\draw  [color={rgb, 255:red, 81; green, 170; blue, 255 }  ,draw opacity=1 ][fill={rgb, 255:red, 150; green, 204; blue, 255 }  ,fill opacity=1 ] (17.63,208.4) -- (165,208.4) -- (165,223.4) -- (17.63,223.4) -- cycle ;
\draw  [color={rgb, 255:red, 81; green, 170; blue, 255 }  ,draw opacity=1 ][fill={rgb, 255:red, 150; green, 204; blue, 255 }  ,fill opacity=1 ] (17.63,228.4) -- (165,228.4) -- (165,253.4) -- (17.63,253.4) -- cycle ;
\draw  [color={rgb, 255:red, 116; green, 232; blue, 164 }  ,draw opacity=1 ][fill={rgb, 255:red, 173; green, 243; blue, 203 }  ,fill opacity=1 ] (258,93.27) .. controls (258,90.61) and (260.16,88.45) .. (262.82,88.45) -- (475.88,88.45) .. controls (478.54,88.45) and (480.7,90.61) .. (480.7,93.27) -- (480.7,107.73) .. controls (480.7,110.39) and (478.54,112.55) .. (475.88,112.55) -- (262.82,112.55) .. controls (260.16,112.55) and (258,110.39) .. (258,107.73) -- cycle ;
\draw  [color={rgb, 255:red, 116; green, 232; blue, 164 }  ,draw opacity=1 ][fill={rgb, 255:red, 173; green, 243; blue, 203 }  ,fill opacity=1 ] (183.3,215.27) .. controls (183.3,212.61) and (185.46,210.45) .. (188.12,210.45) -- (401.18,210.45) .. controls (403.84,210.45) and (406,212.61) .. (406,215.27) -- (406,229.73) .. controls (406,232.39) and (403.84,234.55) .. (401.18,234.55) -- (188.12,234.55) .. controls (185.46,234.55) and (183.3,232.39) .. (183.3,229.73) -- cycle ;

\draw (249.17,46.12) node  [font=\footnotesize] [align=left] {\begin{minipage}[lt]{147.56pt}\setlength\topsep{0pt}
\begin{center}
What is the ML task?
\end{center}

\end{minipage}};
\draw (107.17,74.84) node  [font=\footnotesize] [align=left] {\begin{minipage}[lt]{51pt}\setlength\topsep{0pt}
\begin{center}
classification
\end{center}

\end{minipage}};
\draw (249.44,20) node  [font=\footnotesize] [align=left] {\begin{minipage}[lt]{96.87pt}\setlength\topsep{0pt}
\begin{center}
Target Class Balance
\end{center}

\end{minipage}};
\draw (369.84,75.07) node  [font=\footnotesize] [align=left] {\begin{minipage}[lt]{56.88pt}\setlength\topsep{0pt}
\begin{center}
regression
\end{center}

\end{minipage}};
\draw (111.72,97.12) node  [font=\footnotesize] [align=left] {\begin{minipage}[lt]{123.28pt}\setlength\topsep{0pt}
\begin{center}
Where does the focus lie?
\end{center}

\end{minipage}};
\draw (204.69,126.07) node  [font=\footnotesize] [align=left] {\begin{minipage}[lt]{29.28pt}\setlength\topsep{0pt}
\begin{center}
distribution
\end{center}

\end{minipage}};
\draw (90.31,143.9) node  [font=\footnotesize] [align=left] {\begin{minipage}[lt]{100.21pt}\setlength\topsep{0pt}
\begin{center}
Generalized imbalance ratio
\end{center}

\end{minipage}};
\draw (89.43,126.81) node  [font=\footnotesize] [align=left] {\begin{minipage}[lt]{84.69pt}\setlength\topsep{0pt}
\begin{center}
general estimation
\end{center}

\end{minipage}};
\draw (204.72,172.12) node  [font=\footnotesize] [align=left] {\begin{minipage}[lt]{123.28pt}\setlength\topsep{0pt}
\begin{center}
Classical approach vs distribution agreement?
\end{center}

\end{minipage}};
\draw (92.11,200.59) node  [font=\footnotesize] [align=left] {\begin{minipage}[lt]{65.13pt}\setlength\topsep{0pt}
\begin{center}
classical approach
\end{center}

\end{minipage}};
\draw (300.85,201.11) node  [font=\footnotesize] [align=left] {\begin{minipage}[lt]{107.24pt}\setlength\topsep{0pt}
\begin{center}
distribution agreement
\end{center}

\end{minipage}};
\draw (91.31,215.9) node  [font=\footnotesize] [align=left] {\begin{minipage}[lt]{100.21pt}\setlength\topsep{0pt}
\begin{center}
Imbalance degree
\end{center}

\end{minipage}};
\draw (91.31,240.9) node  [font=\footnotesize] [align=left] {\begin{minipage}[lt]{110.21pt}\setlength\topsep{0pt}
\begin{center}
Likelihood ratio imbalance degree
\end{center}

\end{minipage}};
\draw (369.2,100) node  [font=\footnotesize] [align=left] {\begin{minipage}[lt]{151.64pt}\setlength\topsep{0pt}
\begin{center}
Distribution metrics\\(Target distribution as reference)
\end{center}

\end{minipage}};
\draw (300.5,222) node  [font=\footnotesize] [align=left] {\begin{minipage}[lt]{151.64pt}\setlength\topsep{0pt}
\begin{center}
Distribution metrics\\(Target distribution as reference)
\end{center}

\end{minipage}};

\end{tikzpicture}

%% file: decision-trees/currency.tex
\tikzset{every picture/.style={line width=0.75pt}} 

\begin{tikzpicture}[x=0.75pt,y=0.75pt,yscale=-1,xscale=1]

\draw  [color={rgb, 255:red, 116; green, 232; blue, 164 }  ,draw opacity=1 ][fill={rgb, 255:red, 173; green, 243; blue, 203 }  ,fill opacity=1 ] (15.65,42.6) .. controls (15.65,40.39) and (17.44,38.6) .. (19.65,38.6) -- (471.65,38.6) .. controls (473.86,38.6) and (475.65,40.39) .. (475.65,42.6) -- (475.65,54.6) .. controls (475.65,56.81) and (473.86,58.6) .. (471.65,58.6) -- (19.65,58.6) .. controls (17.44,58.6) and (15.65,56.81) .. (15.65,54.6) -- cycle ;
\draw  [color={rgb, 255:red, 156; green, 158; blue, 255 }  ,draw opacity=1 ][fill={rgb, 255:red, 196; green, 197; blue, 255 }  ,fill opacity=1 ] (133.38,69.37) .. controls (133.38,67.17) and (135.17,65.37) .. (137.38,65.37) -- (354.38,65.37) .. controls (356.59,65.37) and (358.38,67.17) .. (358.38,69.37) -- (358.38,81.37) .. controls (358.38,83.58) and (356.59,85.37) .. (354.38,85.37) -- (137.38,85.37) .. controls (135.17,85.37) and (133.38,83.58) .. (133.38,81.37) -- cycle ;
\draw    (103.98,90.88) -- (383.98,90.88) ;
\draw    (103.98,90.88) -- (103.98,95.88) ;
\draw    (383.98,90.88) -- (383.98,95.88) ;
\draw    (243.85,85.45) -- (243.85,90.45) ;

\draw  [color={rgb, 255:red, 156; green, 158; blue, 255 }  ,draw opacity=1 ][fill={rgb, 255:red, 196; green, 197; blue, 255 }  ,fill opacity=1 ] (15.38,120.37) .. controls (15.38,118.17) and (17.17,116.37) .. (19.38,116.37) -- (207.35,116.37) .. controls (209.56,116.37) and (211.35,118.17) .. (211.35,120.37) -- (211.35,132.37) .. controls (211.35,134.58) and (209.56,136.37) .. (207.35,136.37) -- (19.38,136.37) .. controls (17.17,136.37) and (15.38,134.58) .. (15.38,132.37) -- cycle ;
\draw    (78.78,141.88) -- (210.91,141.88) ;
\draw    (78.78,141.88) -- (78.78,146.88) ;
\draw    (210.91,141.88) -- (210.91,146.88) ;
\draw    (103.78,136.45) -- (103.78,141.45) ;
\draw  [color={rgb, 255:red, 81; green, 170; blue, 255 }  ,draw opacity=1 ][fill={rgb, 255:red, 150; green, 204; blue, 255 }  ,fill opacity=1 ] (14.93,162.4) -- (141.7,162.4) -- (141.7,177.4) -- (14.93,177.4) -- cycle ;
\draw  [color={rgb, 255:red, 81; green, 170; blue, 255 }  ,draw opacity=1 ][fill={rgb, 255:red, 150; green, 204; blue, 255 }  ,fill opacity=1 ] (147.93,162.4) -- (274.7,162.4) -- (274.7,177.4) -- (147.93,177.4) -- cycle ;
\draw  [color={rgb, 255:red, 156; green, 158; blue, 255 }  ,draw opacity=1 ][fill={rgb, 255:red, 196; green, 197; blue, 255 }  ,fill opacity=1 ] (268.26,120.37) .. controls (268.26,118.17) and (270.05,116.37) .. (272.26,116.37) -- (471.23,116.37) .. controls (473.44,116.37) and (475.23,118.17) .. (475.23,120.37) -- (475.23,132.37) .. controls (475.23,134.58) and (473.44,136.37) .. (471.23,136.37) -- (272.26,136.37) .. controls (270.05,136.37) and (268.26,134.58) .. (268.26,132.37) -- cycle ;
\draw  [color={rgb, 255:red, 81; green, 170; blue, 255 }  ,draw opacity=1 ][fill={rgb, 255:red, 150; green, 204; blue, 255 }  ,fill opacity=1 ] (218.93,189.4) -- (345.7,189.4) -- (345.7,204.4) -- (218.93,204.4) -- cycle ;
\draw  [color={rgb, 255:red, 81; green, 170; blue, 255 }  ,draw opacity=1 ][fill={rgb, 255:red, 150; green, 204; blue, 255 }  ,fill opacity=1 ] (351.93,189.4) -- (478.7,189.4) -- (478.7,204.4) -- (351.93,204.4) -- cycle ;
\draw    (283.78,141.88) -- (414.91,141.88) ;
\draw    (283.78,141.88) -- (283.78,146.88) ;
\draw    (414.91,141.88) -- (414.91,146.88) ;
\draw    (383.78,136.45) -- (383.78,141.45) ;
\draw    (282.78,161.88) -- (282.78,188.77) ;
\draw    (414.78,161.88) -- (414.78,188.77) ;

\draw (245.48,75.12) node  [font=\footnotesize] [align=left] {\begin{minipage}[lt]{147.56pt}\setlength\topsep{0pt}
\begin{center}
Does the data have an expiration date?
\end{center}

\end{minipage}};
\draw (103.48,103.84) node  [font=\footnotesize] [align=left] {\begin{minipage}[lt]{51pt}\setlength\topsep{0pt}
\begin{center}
yes
\end{center}

\end{minipage}};
\draw (245.76,49) node  [font=\footnotesize] [align=left] {\begin{minipage}[lt]{96.87pt}\setlength\topsep{0pt}
\begin{center}
Currency
\end{center}

\end{minipage}};
\draw (384.16,104.07) node  [font=\footnotesize] [align=left] {\begin{minipage}[lt]{56.88pt}\setlength\topsep{0pt}
\begin{center}
no
\end{center}

\end{minipage}};
\draw (113.02,126.12) node  [font=\footnotesize] [align=left] {\begin{minipage}[lt]{128.52pt}\setlength\topsep{0pt}
\begin{center}
What kind of decay is expected?
\end{center}

\end{minipage}};
\draw (77.53,154.84) node  [font=\footnotesize] [align=left] {\begin{minipage}[lt]{26.25pt}\setlength\topsep{0pt}
\begin{center}
linear
\end{center}

\end{minipage}};
\draw (211,155.07) node  [font=\footnotesize] [align=left] {\begin{minipage}[lt]{29.28pt}\setlength\topsep{0pt}
\begin{center}
polynomial
\end{center}

\end{minipage}};
\draw (78.32,169.9) node  [font=\footnotesize] [align=left] {\begin{minipage}[lt]{86.2pt}\setlength\topsep{0pt}
\begin{center}
Currency Li et al.
\end{center}

\end{minipage}};
\draw (211.32,169.9) node  [font=\footnotesize] [align=left] {\begin{minipage}[lt]{86.2pt}\setlength\topsep{0pt}
\begin{center}
Currency Ballou et al.
\end{center}

\end{minipage}};
\draw (371.5,125.69) node  [font=\footnotesize] [align=left] {\begin{minipage}[lt]{143pt}\setlength\topsep{0pt}
\begin{center}
Do you have information about the update frequency needed?
\end{center}

\end{minipage}};
\draw (282.32,196.9) node  [font=\footnotesize] [align=left] {\begin{minipage}[lt]{86.2pt}\setlength\topsep{0pt}
\begin{center}
Currency Hinrichs
\end{center}

\end{minipage}};
\draw (415.32,196.9) node  [font=\footnotesize] [align=left] {\begin{minipage}[lt]{86.2pt}\setlength\topsep{0pt}
\begin{center}
Currency Heinrich et al.
\end{center}

\end{minipage}};
\draw (283.53,154.84) node  [font=\footnotesize] [align=left] {\begin{minipage}[lt]{26.25pt}\setlength\topsep{0pt}
\begin{center}
yes
\end{center}

\end{minipage}};
\draw (415,155.07) node  [font=\footnotesize] [align=left] {\begin{minipage}[lt]{29.28pt}\setlength\topsep{0pt}
\begin{center}
no
\end{center}

\end{minipage}};

\end{tikzpicture}

%% file: decision-trees/uniqueness.tex
\tikzset{every picture/.style={line width=0.75pt}} 

\begin{tikzpicture}[x=0.75pt,y=0.75pt,yscale=-1,xscale=1]

\draw  [color={rgb, 255:red, 116; green, 232; blue, 164 }  ,draw opacity=1 ][fill={rgb, 255:red, 173; green, 243; blue, 203 }  ,fill opacity=1 ] (11.67,13.6) .. controls (11.67,11.39) and (13.46,9.6) .. (15.67,9.6) -- (467.67,9.6) .. controls (469.88,9.6) and (471.67,11.39) .. (471.67,13.6) -- (471.67,25.6) .. controls (471.67,27.81) and (469.88,29.6) .. (467.67,29.6) -- (15.67,29.6) .. controls (13.46,29.6) and (11.67,27.81) .. (11.67,25.6) -- cycle ;
\draw  [color={rgb, 255:red, 156; green, 158; blue, 255 }  ,draw opacity=1 ][fill={rgb, 255:red, 196; green, 197; blue, 255 }  ,fill opacity=1 ] (129.4,40.37) .. controls (129.4,38.17) and (131.19,36.37) .. (133.4,36.37) -- (350.4,36.37) .. controls (352.61,36.37) and (354.4,38.17) .. (354.4,40.37) -- (354.4,52.37) .. controls (354.4,54.58) and (352.61,56.37) .. (350.4,56.37) -- (133.4,56.37) .. controls (131.19,56.37) and (129.4,54.58) .. (129.4,52.37) -- cycle ;
\draw    (100,61.88) -- (380,61.88) ;
\draw    (100,61.88) -- (100,66.88) ;
\draw    (380,61.88) -- (380,66.88) ;
\draw    (239.87,56.45) -- (239.87,61.45) ;

\draw  [color={rgb, 255:red, 81; green, 170; blue, 255 }  ,draw opacity=1 ][fill={rgb, 255:red, 150; green, 204; blue, 255 }  ,fill opacity=1 ] (11.98,84) -- (190.82,84) -- (190.82,99) -- (11.98,99) -- cycle ;
\draw  [color={rgb, 255:red, 81; green, 170; blue, 255 }  ,draw opacity=1 ][fill={rgb, 255:red, 150; green, 204; blue, 255 }  ,fill opacity=1 ] (290.98,84) -- (469.82,84) -- (469.82,99) -- (290.98,99) -- cycle ;

\draw (241.5,46.12) node  [font=\footnotesize] [align=left] {\begin{minipage}[lt]{147.56pt}\setlength\topsep{0pt}
\begin{center}
What type of identicality is of concern?
\end{center}

\end{minipage}};
\draw (99.5,74.84) node  [font=\footnotesize] [align=left] {\begin{minipage}[lt]{51pt}\setlength\topsep{0pt}
\begin{center}
fully identical
\end{center}

\end{minipage}};
\draw (241.77,20) node  [font=\footnotesize] [align=left] {\begin{minipage}[lt]{96.87pt}\setlength\topsep{0pt}
\begin{center}
Uniqueness
\end{center}

\end{minipage}};
\draw (380.53,75.43) node  [font=\footnotesize] [align=left] {\begin{minipage}[lt]{80.48pt}\setlength\topsep{0pt}
\begin{center}
logically identical
\end{center}

\end{minipage}};
\draw (101.4,91.5) node  [font=\footnotesize] [align=left] {\begin{minipage}[lt]{121.61pt}\setlength\topsep{0pt}
\begin{center}
Prevalence of duplicates
\end{center}

\end{minipage}};
\draw (380.4,91.5) node  [font=\footnotesize] [align=left] {\begin{minipage}[lt]{121.61pt}\setlength\topsep{0pt}
\begin{center}
Effective sample size
\end{center}

\end{minipage}};

\end{tikzpicture}

%% file: decision-trees/informative-missingness.tex
\tikzset{every picture/.style={line width=0.75pt}} 

\begin{tikzpicture}[x=0.75pt,y=0.75pt,yscale=-1,xscale=1]

\draw  [color={rgb, 255:red, 116; green, 232; blue, 164 }  ,draw opacity=1 ][fill={rgb, 255:red, 173; green, 243; blue, 203 }  ,fill opacity=1 ] (20,17) .. controls (20,14.79) and (21.79,13) .. (24,13) -- (476,13) .. controls (478.21,13) and (480,14.79) .. (480,17) -- (480,29) .. controls (480,31.21) and (478.21,33) .. (476,33) -- (24,33) .. controls (21.79,33) and (20,31.21) .. (20,29) -- cycle ;
\draw  [color={rgb, 255:red, 156; green, 158; blue, 255 }  ,draw opacity=1 ][fill={rgb, 255:red, 196; green, 197; blue, 255 }  ,fill opacity=1 ] (119.73,43.78) .. controls (119.73,41.57) and (121.52,39.78) .. (123.73,39.78) -- (376.5,39.78) .. controls (378.71,39.78) and (380.5,41.57) .. (380.5,43.78) -- (380.5,55.78) .. controls (380.5,57.98) and (378.71,59.78) .. (376.5,59.78) -- (123.73,59.78) .. controls (121.52,59.78) and (119.73,57.98) .. (119.73,55.78) -- cycle ;
\draw    (109.33,65.28) -- (389.33,65.28) ;
\draw    (109.33,65.28) -- (109.33,70.28) ;
\draw    (389.33,65.28) -- (389.33,70.28) ;
\draw    (249.19,65.85) -- (249.19,70.85) ;
\draw    (249.19,59.85) -- (249.19,64.85) ;
\draw  [color={rgb, 255:red, 81; green, 170; blue, 255 }  ,draw opacity=1 ][fill={rgb, 255:red, 150; green, 204; blue, 255 }  ,fill opacity=1 ] (57.4,89.4) -- (306.77,89.4) -- (306.77,104.4) -- (57.4,104.4) -- cycle ;
\draw  [color={rgb, 255:red, 81; green, 170; blue, 255 }  ,draw opacity=1 ][fill={rgb, 255:red, 150; green, 204; blue, 255 }  ,fill opacity=1 ] (325.29,89.4) -- (452.06,89.4) -- (452.06,104.4) -- (325.29,104.4) -- cycle ;

\draw (250.46,49.14) node  [font=\footnotesize] [align=left] {\begin{minipage}[lt]{176.85pt}\setlength\topsep{0pt}
\begin{center}
Which missingness mechanism shall be determined? 
\end{center}

\end{minipage}};
\draw (250,23) node  [font=\footnotesize] [align=left] {\begin{minipage}[lt]{96.87pt}\setlength\topsep{0pt}
\begin{center}
Informative Missingness
\end{center}

\end{minipage}};
\draw (108.58,81.78) node  [font=\footnotesize] [align=left] {\begin{minipage}[lt]{48.62pt}\setlength\topsep{0pt}
\begin{center}
MAR
\end{center}

\end{minipage}};
\draw (248.58,81.78) node  [font=\footnotesize] [align=left] {\begin{minipage}[lt]{48.62pt}\setlength\topsep{0pt}
\begin{center}
MNAR
\end{center}

\end{minipage}};
\draw (389.58,80.78) node  [font=\footnotesize] [align=left] {\begin{minipage}[lt]{48.62pt}\setlength\topsep{0pt}
\begin{center}
MCAR
\end{center}

\end{minipage}};
\draw (182.09,96.9) node  [font=\footnotesize] [align=left] {\begin{minipage}[lt]{169.58pt}\setlength\topsep{0pt}
\begin{center}
Likelihood model for informative dropout
\end{center}

\end{minipage}};
\draw (388.67,96.9) node  [font=\footnotesize] [align=left] {\begin{minipage}[lt]{86.2pt}\setlength\topsep{0pt}
\begin{center}
Little's test
\end{center}

\end{minipage}};

\end{tikzpicture}

%% file: decision-trees/feature-importance.tex
\tikzset{every picture/.style={line width=0.75pt}} 

\begin{tikzpicture}[x=0.75pt,y=0.75pt,yscale=-1,xscale=1]

\draw  [color={rgb, 255:red, 116; green, 232; blue, 164 }  ,draw opacity=1 ][fill={rgb, 255:red, 173; green, 243; blue, 203 }  ,fill opacity=1 ] (20,17.6) .. controls (20,15.39) and (21.79,13.6) .. (24,13.6) -- (476,13.6) .. controls (478.21,13.6) and (480,15.39) .. (480,17.6) -- (480,29.6) .. controls (480,31.81) and (478.21,33.6) .. (476,33.6) -- (24,33.6) .. controls (21.79,33.6) and (20,31.81) .. (20,29.6) -- cycle ;
\draw  [color={rgb, 255:red, 116; green, 232; blue, 164 }  ,draw opacity=1 ][fill={rgb, 255:red, 173; green, 243; blue, 203 }  ,fill opacity=1 ] (137.73,44.37) .. controls (137.73,42.17) and (139.52,40.37) .. (141.73,40.37) -- (358.73,40.37) .. controls (360.94,40.37) and (362.73,42.17) .. (362.73,44.37) -- (362.73,56.37) .. controls (362.73,58.58) and (360.94,60.37) .. (358.73,60.37) -- (141.73,60.37) .. controls (139.52,60.37) and (137.73,58.58) .. (137.73,56.37) -- cycle ;

\draw (249.83,50.12) node  [font=\footnotesize] [align=left] {\begin{minipage}[lt]{147.56pt}\setlength\topsep{0pt}
\begin{center}
Correlation coefficients
\end{center}

\end{minipage}};

\draw (250.11,24) node  [font=\footnotesize] [align=left] {\begin{minipage}[lt]{96.87pt}\setlength\topsep{0pt}
\begin{center}
Feature Importance
\end{center}

\end{minipage}};

\end{tikzpicture}

%% file: decision-trees/distribution-metrics.tex
\tikzset{every picture/.style={line width=0.75pt}} 

\begin{tikzpicture}[x=0.75pt,y=0.75pt,yscale=-1,xscale=1]

\draw  [color={rgb, 255:red, 116; green, 232; blue, 164 }  ,draw opacity=1 ][fill={rgb, 255:red, 173; green, 243; blue, 203 }  ,fill opacity=1 ] (20.85,14) .. controls (20.85,11.79) and (22.64,10) .. (24.85,10) -- (476.85,10) .. controls (479.06,10) and (480.85,11.79) .. (480.85,14) -- (480.85,26) .. controls (480.85,28.21) and (479.06,30) .. (476.85,30) -- (24.85,30) .. controls (22.64,30) and (20.85,28.21) .. (20.85,26) -- cycle ;
\draw  [color={rgb, 255:red, 156; green, 158; blue, 255 }  ,draw opacity=1 ][fill={rgb, 255:red, 196; green, 197; blue, 255 }  ,fill opacity=1 ] (137.58,40.78) .. controls (137.58,38.57) and (139.37,36.78) .. (141.58,36.78) -- (358.58,36.78) .. controls (360.79,36.78) and (362.58,38.57) .. (362.58,40.78) -- (362.58,52.78) .. controls (362.58,54.98) and (360.79,56.78) .. (358.58,56.78) -- (141.58,56.78) .. controls (139.37,56.78) and (137.58,54.98) .. (137.58,52.78) -- cycle ;
\draw    (110.18,62.28) -- (390.18,62.28) ;
\draw    (110.18,62.28) -- (110.18,67.28) ;
\draw    (390.18,62.28) -- (390.18,67.28) ;
\draw    (250.04,56.85) -- (250.04,61.85) ;
\draw  [color={rgb, 255:red, 156; green, 158; blue, 255 }  ,draw opacity=1 ][fill={rgb, 255:red, 196; green, 197; blue, 255 }  ,fill opacity=1 ] (21,91.78) .. controls (21,89.57) and (22.79,87.78) .. (25,87.78) -- (206.3,87.78) .. controls (208.51,87.78) and (210.3,89.57) .. (210.3,91.78) -- (210.3,103.78) .. controls (210.3,105.98) and (208.51,107.78) .. (206.3,107.78) -- (25,107.78) .. controls (22.79,107.78) and (21,105.98) .. (21,103.78) -- cycle ;
\draw  [color={rgb, 255:red, 156; green, 158; blue, 255 }  ,draw opacity=1 ][fill={rgb, 255:red, 196; green, 197; blue, 255 }  ,fill opacity=1 ] (290.3,91.78) .. controls (290.3,89.57) and (292.09,87.78) .. (294.3,87.78) -- (475.76,87.78) .. controls (477.97,87.78) and (479.76,89.57) .. (479.76,91.78) -- (479.76,103.78) .. controls (479.76,105.98) and (477.97,107.78) .. (475.76,107.78) -- (294.3,107.78) .. controls (292.09,107.78) and (290.3,105.98) .. (290.3,103.78) -- cycle ;
\draw    (74.1,113.28) -- (193.62,113.28) ;
\draw    (74.1,113.28) -- (74.1,118.28) ;
\draw    (193.62,113.28) -- (193.62,118.28) ;
\draw    (110.04,107.85) -- (110.04,112.85) ;
\draw  [color={rgb, 255:red, 81; green, 170; blue, 255 }  ,draw opacity=1 ][fill={rgb, 255:red, 150; green, 204; blue, 255 }  ,fill opacity=1 ] (19.05,133) -- (130.03,133) -- (130.03,148) -- (19.05,148) -- cycle ;
\draw  [color={rgb, 255:red, 81; green, 170; blue, 255 }  ,draw opacity=1 ][fill={rgb, 255:red, 150; green, 204; blue, 255 }  ,fill opacity=1 ] (139.05,133) -- (250.03,133) -- (250.03,148) -- (139.05,148) -- cycle ;
\draw  [color={rgb, 255:red, 81; green, 170; blue, 255 }  ,draw opacity=1 ][fill={rgb, 255:red, 150; green, 204; blue, 255 }  ,fill opacity=1 ] (139.05,153) -- (250.03,153) -- (250.03,168) -- (139.05,168) -- cycle ;
\draw  [color={rgb, 255:red, 81; green, 170; blue, 255 }  ,draw opacity=1 ][fill={rgb, 255:red, 150; green, 204; blue, 255 }  ,fill opacity=1 ] (139.05,173) -- (250.03,173) -- (250.03,195) -- (139.05,195) -- cycle ;
\draw    (311.21,113.28) -- (469.73,113.28) ;
\draw    (311.04,113.28) -- (311.04,118.28) ;
\draw    (469.36,113.28) -- (469.36,118.28) ;
\draw    (391.04,107.85) -- (391.04,112.85) ;
\draw    (391.04,113.85) -- (391.04,118.85) ;
\draw    (311.04,152.28) -- (311.04,237.28) ;
\draw    (89.53,237.28) -- (311.04,237.28) ;
\draw    (89.1,237.28) -- (89.1,242.28) ;
\draw  [color={rgb, 255:red, 81; green, 170; blue, 255 }  ,draw opacity=1 ][fill={rgb, 255:red, 150; green, 204; blue, 255 }  ,fill opacity=1 ] (20.05,243) -- (158.97,243) -- (158.97,258) -- (20.05,258) -- cycle ;
\draw  [color={rgb, 255:red, 81; green, 170; blue, 255 }  ,draw opacity=1 ][fill={rgb, 255:red, 150; green, 204; blue, 255 }  ,fill opacity=1 ] (20.05,263) -- (158.99,263) -- (158.99,278) -- (20.05,278) -- cycle ;
\draw  [color={rgb, 255:red, 81; green, 170; blue, 255 }  ,draw opacity=1 ][fill={rgb, 255:red, 150; green, 204; blue, 255 }  ,fill opacity=1 ] (20.05,283) -- (158.99,283) -- (158.99,298) -- (20.05,298) -- cycle ;
\draw  [color={rgb, 255:red, 81; green, 170; blue, 255 }  ,draw opacity=1 ][fill={rgb, 255:red, 150; green, 204; blue, 255 }  ,fill opacity=1 ] (19.05,303) -- (158.99,303) -- (158.99,318) -- (19.05,318) -- cycle ;
\draw  [color={rgb, 255:red, 156; green, 158; blue, 255 }  ,draw opacity=1 ][fill={rgb, 255:red, 196; green, 197; blue, 255 }  ,fill opacity=1 ] (179.36,275.77) .. controls (179.36,273.57) and (181.15,271.77) .. (183.36,271.77) -- (407.3,271.77) .. controls (409.51,271.77) and (411.3,273.57) .. (411.3,275.77) -- (411.3,287.77) .. controls (411.3,289.98) and (409.51,291.77) .. (407.3,291.77) -- (183.36,291.77) .. controls (181.15,291.77) and (179.36,289.98) .. (179.36,287.77) -- cycle ;
\draw    (221.1,297.28) -- (363.62,297.28) ;
\draw    (221.1,297.28) -- (221.1,302.28) ;
\draw    (363.62,297.28) -- (363.62,302.28) ;
\draw    (294.04,291.85) -- (294.04,296.85) ;
\draw  [color={rgb, 255:red, 81; green, 170; blue, 255 }  ,draw opacity=1 ][fill={rgb, 255:red, 150; green, 204; blue, 255 }  ,fill opacity=1 ] (145.05,327) -- (284.99,327) -- (284.99,342) -- (145.05,342) -- cycle ;
\draw  [color={rgb, 255:red, 81; green, 170; blue, 255 }  ,draw opacity=1 ][fill={rgb, 255:red, 150; green, 204; blue, 255 }  ,fill opacity=1 ] (145.05,347) -- (284.99,347) -- (284.99,362) -- (145.05,362) -- cycle ;
\draw  [color={rgb, 255:red, 81; green, 170; blue, 255 }  ,draw opacity=1 ][fill={rgb, 255:red, 150; green, 204; blue, 255 }  ,fill opacity=1 ] (294.06,327) -- (452.26,327) -- (452.26,342) -- (294.06,342) -- cycle ;
\draw  [color={rgb, 255:red, 81; green, 170; blue, 255 }  ,draw opacity=1 ][fill={rgb, 255:red, 150; green, 204; blue, 255 }  ,fill opacity=1 ] (294.06,347) -- (452.26,347) -- (452.26,362) -- (294.06,362) -- cycle ;
\draw  [color={rgb, 255:red, 81; green, 170; blue, 255 }  ,draw opacity=1 ][fill={rgb, 255:red, 150; green, 204; blue, 255 }  ,fill opacity=1 ] (294.06,367) -- (452.26,367) -- (452.26,382) -- (294.06,382) -- cycle ;
\draw    (390.04,151.28) -- (390,255) ;
\draw    (293.49,255) -- (390,255) ;
\draw    (294.04,254.85) -- (294.04,271.85) ;
\draw    (470.04,152.28) -- (470,399) ;
\draw  [color={rgb, 255:red, 156; green, 158; blue, 255 }  ,draw opacity=1 ][fill={rgb, 255:red, 196; green, 197; blue, 255 }  ,fill opacity=1 ] (268,416.78) .. controls (268,414.57) and (269.79,412.78) .. (272,412.78) -- (441.76,412.78) .. controls (443.97,412.78) and (445.76,414.57) .. (445.76,416.78) -- (445.76,428.78) .. controls (445.76,430.98) and (443.97,432.78) .. (441.76,432.78) -- (272,432.78) .. controls (269.79,432.78) and (268,430.98) .. (268,428.78) -- cycle ;
\draw    (190.47,438.28) -- (386.68,438.28) ;
\draw    (190.47,438.28) -- (190.47,443.28) ;
\draw    (386.68,438.28) -- (386.68,443.28) ;
\draw    (295.55,432.85) -- (295.55,437.85) ;
\draw  [color={rgb, 255:red, 81; green, 170; blue, 255 }  ,draw opacity=1 ][fill={rgb, 255:red, 150; green, 204; blue, 255 }  ,fill opacity=1 ] (136.05,461) -- (247.03,461) -- (247.03,476) -- (136.05,476) -- cycle ;
\draw  [color={rgb, 255:red, 156; green, 158; blue, 255 }  ,draw opacity=1 ][fill={rgb, 255:red, 196; green, 197; blue, 255 }  ,fill opacity=1 ] (286.3,464.78) .. controls (286.3,462.57) and (288.09,460.78) .. (290.3,460.78) -- (475,460.78) .. controls (477.21,460.78) and (479,462.57) .. (479,464.78) -- (479,476.78) .. controls (479,478.98) and (477.21,480.78) .. (475,480.78) -- (290.3,480.78) .. controls (288.09,480.78) and (286.3,478.98) .. (286.3,476.78) -- cycle ;
\draw    (298.04,486.95) -- (422.23,486.95) ;
\draw    (298.04,486.95) -- (298.04,491.95) ;
\draw    (422.23,486.95) -- (422.23,491.95) ;
\draw    (386.35,481.52) -- (386.35,486.52) ;
\draw  [color={rgb, 255:red, 81; green, 170; blue, 255 }  ,draw opacity=1 ][fill={rgb, 255:red, 150; green, 204; blue, 255 }  ,fill opacity=1 ] (368.05,508) -- (479.03,508) -- (479.03,535) -- (368.05,535) -- cycle ;
\draw  [color={rgb, 255:red, 81; green, 170; blue, 255 }  ,draw opacity=1 ][fill={rgb, 255:red, 150; green, 204; blue, 255 }  ,fill opacity=1 ] (231.7,508) -- (357.89,508) -- (357.89,523) -- (231.7,523) -- cycle ;
\draw  [color={rgb, 255:red, 81; green, 170; blue, 255 }  ,draw opacity=1 ][fill={rgb, 255:red, 150; green, 204; blue, 255 }  ,fill opacity=1 ] (231.67,528) -- (357.86,528) -- (357.86,543) -- (231.67,543) -- cycle ;
\draw  [color={rgb, 255:red, 81; green, 170; blue, 255 }  ,draw opacity=1 ][fill={rgb, 255:red, 150; green, 204; blue, 255 }  ,fill opacity=1 ] (231.67,548) -- (357.86,548) -- (357.86,563) -- (231.67,563) -- cycle ;
\draw    (350.79,399) -- (470,399) ;
\draw    (350.62,398.28) -- (350.62,412.28) ;

\draw (250.38,46.84) node  [font=\footnotesize] [align=left] {\begin{minipage}[lt]{152.59pt}\setlength\topsep{0pt}
\begin{center}
What aspect is examined?
\end{center}

\end{minipage}};
\draw (250.85,20) node  [font=\footnotesize] [align=left] {\begin{minipage}[lt]{96.87pt}\setlength\topsep{0pt}
\begin{center}
Distribution metrics
\end{center}

\end{minipage}};
\draw (110.49,77.75) node  [font=\footnotesize] [align=left] {\begin{minipage}[lt]{220.78pt}\setlength\topsep{0pt}
\begin{center}
characteristics of sample distribution
\end{center}

\end{minipage}};
\draw (390.32,77.57) node  [font=\footnotesize] [align=left] {\begin{minipage}[lt]{172.75pt}\setlength\topsep{0pt}
\begin{center}
comparison against other distribution
\end{center}

\end{minipage}};
\draw (115.9,97.84) node  [font=\footnotesize] [align=left] {\begin{minipage}[lt]{128.38pt}\setlength\topsep{0pt}
\begin{center}
What is the data type?
\end{center}

\end{minipage}};
\draw (385.03,97.78) node  [font=\footnotesize] [align=left] {\begin{minipage}[lt]{128.49pt}\setlength\topsep{0pt}
\begin{center}
Which mathematical approach?
\end{center}

\end{minipage}};
\draw (75.17,127.89) node  [font=\footnotesize] [align=left] {\begin{minipage}[lt]{38.62pt}\setlength\topsep{0pt}
\begin{center}
categorical
\end{center}

\end{minipage}};
\draw (193.86,128.01) node  [font=\footnotesize] [align=left] {\begin{minipage}[lt]{50.57pt}\setlength\topsep{0pt}
\begin{center}
continuous
\end{center}

\end{minipage}};
\draw (74.54,140.5) node  [font=\footnotesize] [align=left] {\begin{minipage}[lt]{75.47pt}\setlength\topsep{0pt}
\begin{center}
Hill numbers
\end{center}

\end{minipage}};
\draw (194.54,140.5) node  [font=\footnotesize] [align=left] {\begin{minipage}[lt]{75.47pt}\setlength\topsep{0pt}
\begin{center}
Range
\end{center}

\end{minipage}};
\draw (194.54,160.5) node  [font=\footnotesize] [align=left] {\begin{minipage}[lt]{75.47pt}\setlength\topsep{0pt}
\begin{center}
Interquartile range
\end{center}

\end{minipage}};
\draw (194.54,184) node  [font=\footnotesize] [align=left] {\begin{minipage}[lt]{75.47pt}\setlength\topsep{0pt}
\begin{center}
Mean and standard deviation
\end{center}

\end{minipage}};

\draw (310.42,136.5) node  [font=\footnotesize] [align=left] {\begin{minipage}[lt]{63.42pt}\setlength\topsep{0pt}
\begin{center}
divergence between two distributions
\end{center}

\end{minipage}};
\draw (469.68,132.68) node  [font=\footnotesize] [align=left] {\begin{minipage}[lt]{68.12pt}\setlength\topsep{0pt}
\begin{center}
statistical \\tests
\end{center}

\end{minipage}};
\draw (390.85,132.18) node  [font=\footnotesize] [align=left] {\begin{minipage}[lt]{68.12pt}\setlength\topsep{0pt}
\begin{center}
distance between two distributions
\end{center}

\end{minipage}};
\draw (89.51,250.5) node  [font=\footnotesize] [align=left] {\begin{minipage}[lt]{100.46pt}\setlength\topsep{0pt}
\begin{center}
Kullback-Leibler divergence
\end{center}

\end{minipage}};
\draw (89.52,270.5) node  [font=\footnotesize] [align=left] {\begin{minipage}[lt]{94.48pt}\setlength\topsep{0pt}
\begin{center}
Jensen Shannon divergence
\end{center}

\end{minipage}};
\draw (89.52,290.5) node  [font=\footnotesize] [align=left] {\begin{minipage}[lt]{94.48pt}\setlength\topsep{0pt}
\begin{center}
Population stability index
\end{center}

\end{minipage}};
\draw (89.02,310.5) node  [font=\footnotesize] [align=left] {\begin{minipage}[lt]{95.16pt}\setlength\topsep{0pt}
\begin{center}
Cohen's d
\end{center}

\end{minipage}};
\draw (294.13,281.77) node  [font=\footnotesize] [align=left] {\begin{minipage}[lt]{157.3pt}\setlength\topsep{0pt}
\begin{center}
Which data modalities exist in your data?
\end{center}

\end{minipage}};
\draw (221.14,312.17) node  [font=\footnotesize] [align=left] {\begin{minipage}[lt]{79.1pt}\setlength\topsep{0pt}
\begin{center}
images \\(inception network)
\end{center}

\end{minipage}};
\draw (363.99,312.37) node  [font=\footnotesize] [align=left] {\begin{minipage}[lt]{72.5pt}\setlength\topsep{0pt}
\begin{center}
images, tabular,\\time series
\end{center}

\end{minipage}};
\draw (215.02,334.5) node  [font=\footnotesize] [align=left] {\begin{minipage}[lt]{95.16pt}\setlength\topsep{0pt}
\begin{center}
Kernel inception distance
\end{center}

\end{minipage}};
\draw (215.02,354.5) node  [font=\footnotesize] [align=left] {\begin{minipage}[lt]{95.16pt}\setlength\topsep{0pt}
\begin{center}
Frechet inception distance
\end{center}

\end{minipage}};
\draw (373.16,334.5) node  [font=\footnotesize] [align=left] {\begin{minipage}[lt]{107.58pt}\setlength\topsep{0pt}
\begin{center}
Wasserstein distance
\end{center}

\end{minipage}};
\draw (373.16,354.5) node  [font=\footnotesize] [align=left] {\begin{minipage}[lt]{107.58pt}\setlength\topsep{0pt}
\begin{center}
Maximum mean discrepancy
\end{center}

\end{minipage}};
\draw (373.16,374.5) node  [font=\footnotesize] [align=left] {\begin{minipage}[lt]{107.58pt}\setlength\topsep{0pt}
\begin{center}
Energy distance
\end{center}

\end{minipage}};
\draw (357.12,422.84) node  [font=\footnotesize] [align=left] {\begin{minipage}[lt]{120.56pt}\setlength\topsep{0pt}
\begin{center}
What is the data type?
\end{center}

\end{minipage}};
\draw (190.58,452.89) node  [font=\footnotesize] [align=left] {\begin{minipage}[lt]{64.48pt}\setlength\topsep{0pt}
\begin{center}
categorical
\end{center}

\end{minipage}};
\draw (387.08,453.01) node  [font=\footnotesize] [align=left] {\begin{minipage}[lt]{84.43pt}\setlength\topsep{0pt}
\begin{center}
continuous
\end{center}

\end{minipage}};
\draw (191.54,468.5) node  [font=\footnotesize] [align=left] {\begin{minipage}[lt]{75.47pt}\setlength\topsep{0pt}
\begin{center}
$\displaystyle \chi ^{2}$ test
\end{center}

\end{minipage}};
\draw (382.91,470.84) node  [font=\footnotesize] [align=left] {\begin{minipage}[lt]{130.69pt}\setlength\topsep{0pt}
\begin{center}
How many distribution groups exist?
\end{center}

\end{minipage}};
\draw (293.6,499.25) node  [font=\footnotesize] [align=left] {\begin{minipage}[lt]{8.98pt}\setlength\topsep{0pt}
\begin{center}
two
\end{center}

\end{minipage}};
\draw (422.9,499.31) node  [font=\footnotesize] [align=left] {\begin{minipage}[lt]{27.74pt}\setlength\topsep{0pt}
\begin{center}
multiple
\end{center}

\end{minipage}};
\draw (423.54,521.5) node  [font=\footnotesize] [align=left] {\begin{minipage}[lt]{75.47pt}\setlength\topsep{0pt}
\begin{center}
k-sample Anderson-Darling test
\end{center}

\end{minipage}};
\draw (423.51,535.5) node  [font=\footnotesize] [align=left] {\begin{minipage}[lt]{75.47pt}\setlength\topsep{0pt}
\begin{center}

\end{center}

\end{minipage}};
\draw (294.8,515.5) node  [font=\footnotesize] [align=left] {\begin{minipage}[lt]{85.81pt}\setlength\topsep{0pt}
\begin{center}
Epps-Singelton test
\end{center}

\end{minipage}};
\draw (295.51,535.56) node  [font=\footnotesize] [align=left] {\begin{minipage}[lt]{135.19pt}\setlength\topsep{0pt}
\begin{center}
Mann-Whitney U-rank test
\end{center}

\end{minipage}};
\draw (294.94,555.56) node  [font=\footnotesize] [align=left] {\begin{minipage}[lt]{104.13pt}\setlength\topsep{0pt}
\begin{center}
Kolmogorov-Smirnov test
\end{center}

\end{minipage}};

\end{tikzpicture}

%% file: decision-trees/correlation-coefficients.tex
\tikzset{every picture/.style={line width=0.75pt}} 

\begin{tikzpicture}[x=0.75pt,y=0.75pt,yscale=-1,xscale=1]

\draw  [color={rgb, 255:red, 116; green, 232; blue, 164 }  ,draw opacity=1 ][fill={rgb, 255:red, 173; green, 243; blue, 203 }  ,fill opacity=1 ] (10.67,11.6) .. controls (10.67,9.39) and (12.46,7.6) .. (14.67,7.6) -- (466.67,7.6) .. controls (468.88,7.6) and (470.67,9.39) .. (470.67,11.6) -- (470.67,23.6) .. controls (470.67,25.81) and (468.88,27.6) .. (466.67,27.6) -- (14.67,27.6) .. controls (12.46,27.6) and (10.67,25.81) .. (10.67,23.6) -- cycle ;
\draw  [color={rgb, 255:red, 156; green, 158; blue, 255 }  ,draw opacity=1 ][fill={rgb, 255:red, 196; green, 197; blue, 255 }  ,fill opacity=1 ] (113.4,38.37) .. controls (113.4,36.17) and (115.19,34.37) .. (117.4,34.37) -- (360.7,34.37) .. controls (362.91,34.37) and (364.7,36.17) .. (364.7,38.37) -- (364.7,50.37) .. controls (364.7,52.58) and (362.91,54.37) .. (360.7,54.37) -- (117.4,54.37) .. controls (115.19,54.37) and (113.4,52.58) .. (113.4,50.37) -- cycle ;
\draw    (100,179.88) -- (380,179.88) ;
\draw    (100,179.88) -- (100,184.88) ;
\draw    (380,179.88) -- (380,184.88) ;
\draw    (239.35,174.45) -- (239.35,179.45) ;
\draw    (100,59.88) -- (380,59.88) ;
\draw    (100,59.88) -- (100,64.88) ;
\draw    (380,59.88) -- (380,64.88) ;
\draw    (239.86,60.45) -- (239.86,65.45) ;
\draw    (239.86,54.45) -- (239.86,59.45) ;
\draw  [color={rgb, 255:red, 81; green, 170; blue, 255 }  ,draw opacity=1 ][fill={rgb, 255:red, 150; green, 204; blue, 255 }  ,fill opacity=1 ] (8.98,84) -- (187.82,84) -- (187.82,99) -- (8.98,99) -- cycle ;
\draw  [color={rgb, 255:red, 81; green, 170; blue, 255 }  ,draw opacity=1 ][fill={rgb, 255:red, 150; green, 204; blue, 255 }  ,fill opacity=1 ] (8.98,104) -- (187.82,104) -- (187.82,119) -- (8.98,119) -- cycle ;
\draw  [color={rgb, 255:red, 81; green, 170; blue, 255 }  ,draw opacity=1 ][fill={rgb, 255:red, 150; green, 204; blue, 255 }  ,fill opacity=1 ] (8.98,124) -- (187.82,124) -- (187.82,139) -- (8.98,139) -- cycle ;
\draw  [color={rgb, 255:red, 81; green, 170; blue, 255 }  ,draw opacity=1 ][fill={rgb, 255:red, 150; green, 204; blue, 255 }  ,fill opacity=1 ] (291.98,84) -- (470.82,84) -- (470.82,99) -- (291.98,99) -- cycle ;
\draw    (239.86,83.3) -- (239.86,154.2) ;
\draw  [color={rgb, 255:red, 156; green, 158; blue, 255 }  ,draw opacity=1 ][fill={rgb, 255:red, 196; green, 197; blue, 255 }  ,fill opacity=1 ] (114.11,158.19) .. controls (114.11,155.96) and (115.92,154.15) .. (118.15,154.15) -- (360.65,154.15) .. controls (362.89,154.15) and (364.7,155.96) .. (364.7,158.19) -- (364.7,170.33) .. controls (364.7,172.56) and (362.89,174.38) .. (360.65,174.38) -- (118.15,174.38) .. controls (115.92,174.38) and (114.11,172.56) .. (114.11,170.33) -- cycle ;
\draw  [color={rgb, 255:red, 81; green, 170; blue, 255 }  ,draw opacity=1 ][fill={rgb, 255:red, 150; green, 204; blue, 255 }  ,fill opacity=1 ] (8.98,200) -- (187.82,200) -- (187.82,215) -- (8.98,215) -- cycle ;
\draw  [color={rgb, 255:red, 81; green, 170; blue, 255 }  ,draw opacity=1 ][fill={rgb, 255:red, 150; green, 204; blue, 255 }  ,fill opacity=1 ] (8.98,220) -- (187.82,220) -- (187.82,235) -- (8.98,235) -- cycle ;
\draw  [color={rgb, 255:red, 81; green, 170; blue, 255 }  ,draw opacity=1 ][fill={rgb, 255:red, 150; green, 204; blue, 255 }  ,fill opacity=1 ] (292.98,200) -- (471.82,200) -- (471.82,215) -- (292.98,215) -- cycle ;

\draw (239.39,44.44) node  [font=\footnotesize] [align=left] {\begin{minipage}[lt]{170.43pt}\setlength\topsep{0pt}
\begin{center}
What is the data type?
\end{center}

\end{minipage}};
\draw (99,192.18) node  [font=\footnotesize] [align=left] {\begin{minipage}[lt]{20.4pt}\setlength\topsep{0pt}
\begin{center}
two
\end{center}

\end{minipage}};
\draw (379.7,191.92) node  [font=\footnotesize] [align=left] {\begin{minipage}[lt]{20.4pt}\setlength\topsep{0pt}
\begin{center}
multiple
\end{center}

\end{minipage}};
\draw (240.67,17.6) node  [font=\footnotesize] [align=left] {\begin{minipage}[lt]{96.87pt}\setlength\topsep{0pt}
\begin{center}
Correlation coefficients
\end{center}

\end{minipage}};
\draw (99.25,76.38) node  [font=\footnotesize] [align=left] {\begin{minipage}[lt]{48.62pt}\setlength\topsep{0pt}
\begin{center}
ordinal
\end{center}

\end{minipage}};
\draw (240.25,76.38) node  [font=\footnotesize] [align=left] {\begin{minipage}[lt]{48.62pt}\setlength\topsep{0pt}
\begin{center}
numerical
\end{center}

\end{minipage}};
\draw (380.25,75.38) node  [font=\footnotesize] [align=left] {\begin{minipage}[lt]{48.62pt}\setlength\topsep{0pt}
\begin{center}
categorical
\end{center}

\end{minipage}};
\draw (101.5,91.5) node  [font=\footnotesize] [align=left] {\begin{minipage}[lt]{130.77pt}\setlength\topsep{0pt}
\begin{center}
Kendall's rank correlation coefficient
\end{center}

\end{minipage}};
\draw (101.5,111.5) node  [font=\footnotesize] [align=left] {\begin{minipage}[lt]{135.77pt}\setlength\topsep{0pt}
\begin{center}
Spearman's rank corr. coefficient
\end{center}

\end{minipage}};
\draw (101.5,131.5) node  [font=\footnotesize] [align=left] {\begin{minipage}[lt]{125.77pt}\setlength\topsep{0pt}
\begin{center}
Goodman-Kruskal's gamma
\end{center}

\end{minipage}};
\draw (379.14,91.5) node  [font=\footnotesize] [align=left] {\begin{minipage}[lt]{125.77pt}\setlength\topsep{0pt}
\begin{center}
Cramer's V
\end{center}

\end{minipage}};
\draw (239.4,164.26) node  [font=\footnotesize] [align=left] {\begin{minipage}[lt]{170.41pt}\setlength\topsep{0pt}
\begin{center}
How many repeated measurements/labels exist?
\end{center}

\end{minipage}};
\draw (101.46,207.5) node  [font=\footnotesize] [align=left] {\begin{minipage}[lt]{125.77pt}\setlength\topsep{0pt}
\begin{center}
Concordance correlation coefficient
\end{center}

\end{minipage}};
\draw (101.46,227.5) node  [font=\footnotesize] [align=left] {\begin{minipage}[lt]{125.77pt}\setlength\topsep{0pt}
\begin{center}
Pearson's correlation coefficient
\end{center}

\end{minipage}};
\draw (382.14,207.5) node  [font=\footnotesize] [align=left] {\begin{minipage}[lt]{125.77pt}\setlength\topsep{0pt}
\begin{center}
Intraclass correlation coefficient
\end{center}

\end{minipage}};

\end{tikzpicture}
 